\documentclass[final]{cvpr}
\usepackage{times}
\usepackage{epsfig}
\usepackage{graphicx}
\usepackage{amsmath}
\usepackage{amssymb}
\usepackage{mathtools}
\usepackage{resizegather}
\usepackage[export]{adjustbox}
\usepackage{bbm}
\usepackage{multirow}
\usepackage{enumitem}
\usepackage{pifont}
\newcommand{\cmark}{\ding{51}}
\newcommand{\xmark}{\ding{55}}
\usepackage{comment}
\usepackage{hyperref}
\hypersetup{pagebackref=true,breaklinks=true,colorlinks,bookmarks=false}

\def\cvprPaperID{*****}
\def\confYear{CVPR 2021}
\newcommand{\bx}{\boldsymbol{x}}
\newcommand{\bX}{\boldsymbol{X}}

\newcommand{\by}{\boldsymbol{y}}
\newcommand{\bY}{\textbf{Y}}
\newcommand{\bz}{\boldsymbol{z}}

\makeatletter
\DeclareRobustCommand\onedot{\futurelet\@let@token\@onedot}
\def\@onedot{\ifx\@let@token.\else.\null\fi\xspace}
\def\eg{\emph{e.g}\onedot} 
\def\ie{\emph{i.e}\onedot}

\def\etal{\emph{et al}\onedot}
\makeatother

\newcommand*{\our}{DeCo\@\xspace}

\begin{document}
\title{Denoise and Contrast for Category Agnostic Shape Completion}
\author{Antonio Alliegro$^{1}$
\hspace{0.8cm}
Diego Valsesia$^{1}$
\hspace{0.8cm}
Giulia Fracastoro$^{1}$\vspace{1mm}\\
Enrico Magli$^{1}$
\hspace{0.8cm}
Tatiana Tommasi$^{1,2}$ \vspace{4mm}\\
$^{1}$Politecnico di Torino, Italy
\hspace{0.5cm}
$^{2}$Italian Institute of Technology\vspace{2mm}\\
{\tt\small \{name.surname\}@polito.it}\\
}

\maketitle

\begin{abstract}
In this paper, we present a deep learning model that exploits the power of 
self-supervision to perform 3D point cloud completion, estimating the missing part and a context region around it. Local and global information are encoded in a combined embedding.
A denoising pretext task provides the network with the needed local cues, decoupled from the high-level semantics and naturally shared over multiple classes. On the other hand, contrastive learning maximizes the agreement between variants of the same shape with different missing portions, thus producing a  representation which captures the global appearance of the shape. 
The combined embedding inherits category-agnostic properties from the chosen pretext tasks. Differently from existing approaches, this allows to better generalize the completion properties to new categories unseen at training time. 
Moreover, while decoding the obtained joint representation, we better blend the reconstructed missing part with the partial shape by paying attention to its known surrounding region and reconstructing this frame as auxiliary objective. 
Our extensive experiments and detailed ablation on the ShapeNet dataset show the effectiveness of each part of the method with new state of the art results. Our quantitative and qualitative analysis confirms how our approach is able to work on novel categories without relying neither on classification and shape symmetry priors, nor on adversarial training procedures. 
\end{abstract}
\section{Introduction}
\begin{figure}[t]
\includegraphics[width=0.48\textwidth]{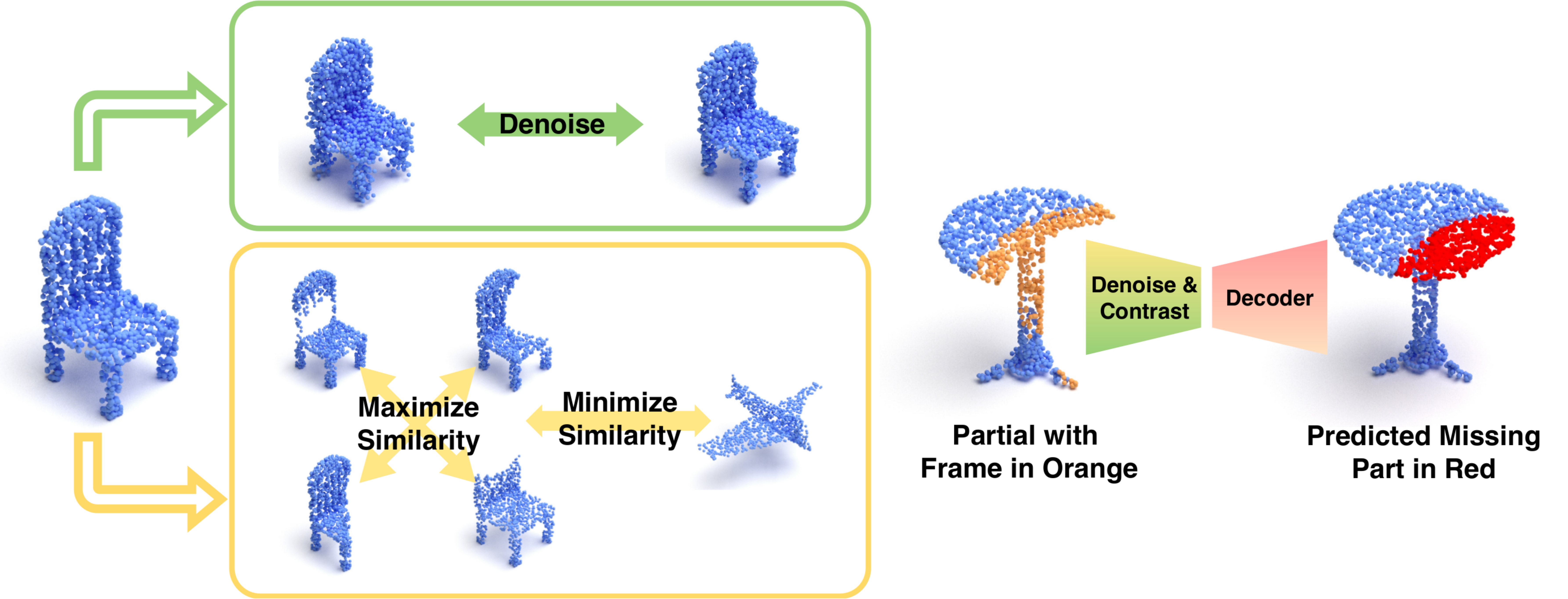}
\caption{Our \our encodes local and global information from the training data via denoising and contrastive learning. The learned embedding is finally decoded  to estimate the missing part of the input shape and a frame, \ie a context region around the hole. Thanks to the class-agnostic nature of the self-supervised pretext tasks, our model is effective for point-cloud completion on novel object categories.}
\label{fig:teaser}
\vspace{-4mm}
\end{figure}

Cameras that scan and render objects in 3D are becoming more and more available as standard feature in many smartphones, drones, robots and cars. Most of these 3D sensing technologies are low-cost stereo cameras as well as depth and laser scanners that output point clouds which are often incomplete due to occlusions, transparency, light reflections or limitations in resolution and viewing angle. The missing regions corrupt the object shape preventing its direct use in tasks like robotic manipulation \cite{Varley_2017_IROS_robotgrasp}, scene understanding \cite{hou_cvpr_2019_sceneunderstanding}, autonomous driving \cite{Chen_2020_IEEEspm_autonomousdrive} and augmented reality \cite{Liu_2020_CVR_augmentedreal}. To overcome those issues, point cloud completion aims at estimating the complete geometry of the missing regions from partial observations. 

There have been several efforts to tackle the completion problem including volumetric representations and related distance fields or mesh models. The most recent literature focuses on the efficient solution of 
directly inferring new points: 
a widely used pipeline consists in encoding the partial input into a latent representation which is then decoded to produce the whole shape.
However, this strategy leads to an overly difficult setting, where the method attempts at reconstructing the entire point cloud rather than simply filling the missing part. As a consequence, the learned model captures the global geometry more than local properties of each sample, resulting in reconstructions that resemble a generic average object rather than the specific input instance. 
Na\"ive design choices of the encoder also contribute to this effect by squashing all the structural information of the point cloud into a single latent global feature with a significant information loss on the details of local regions. 

Among the techniques proposed to improve local and global feature fusion, some try to improve the encoder by describing the point cloud as a collection of surface elements with expansion constraints \cite{Liu_2020_AAAI_morphing}, others model the 3D skeleton of the object \cite{nie_2020_NIPS_skeletonbridged} or propose new pooling operations \cite{Wang_2020_ECCV_softpoolnet}. Better decoders have been also developed by revisiting and accumulating low level features as local descriptors \cite{Huang_2020_CVPR_pfnet,Wen_2020_CVPR_skipattention, Zhang_2020_ECCV_detail}, adopting a pyramid strategy to recover the missing geometry at multiple resolution levels \cite{Huang_2020_CVPR_pfnet}, or including skip and cascaded connections to share information with the encoder \cite{Wen_2020_CVPR_skipattention, Wang_2020_CVPR_cascaded}.  Other approaches exploit local refinements by point upsampling \cite{Liu_2020_AAAI_morphing, yuan_2018_3dv_pcn}, or via adversarial training of patch discriminators \cite{Wang_2020_CVPR_cascaded, nie_2020_NIPS_skeletonbridged}.  
Most of these techniques have never been challenged neither with point clouds corrupted with more than one hole, nor with the reconstruction of object categories unseen at training time. As a matter of fact, in some cases, per-class shape priors are adopted as supervised oracle initialization for the missing points \cite{Wang_2020_CVPR_cascaded}. In order to overcome this closed-set scenario, we propose a novel point cloud completion method that exploits the power of two self-supervised pretext tasks and inherits their category-agnostic properties with a clear generalization effect. Specifically, the main contributions of this work are summarized as follows:
\begin{itemize}[leftmargin=*]
    \vspace{-1mm}\item We propose \our (see Figure \ref{fig:teaser}), a model for point cloud completion, that combines local information from Denoising \cite{Pistilli_2020_ECCV_denoise} and global information from Contrastive learning \cite{chen_2020_icml_simclr}. In this way we shed new light on local and global cues which are otherwise reduced just to features at different network depths.
    \vspace{-1mm}\item The self-supervised learned embedding is finally decoded  to estimate the missing part of the input shape and a frame, \ie a context region around the hole. This solution avoids the risks of  genus-wise distortions \cite{Yang_CVPR_2018_foldingnet-genuswise}, and allows to better blend the predicted missing part to the incomplete input.
    \vspace{-1mm}\item \our's architecture is designed by exploiting graph convolutions. To the best of our knowledge, the graph logic is used here on point cloud completion for the first time.
    \vspace{-1mm}\item We present extensive experiments on ShapeNet \cite{Chang_shapenet_2015} with point clouds corrupted by single and multiple holes as well as testing on novel categories. Our quantitative and qualitative results show the effectiveness of \our and set the new state of the art.
\end{itemize}

\section{Related Work}
Shape completion has a long tradition in the computer graphics and vision fields and has recently attracted the attention of the deep learning community. Early works engineered effective descriptors by leveraging geometric cues \cite{DavisMGL_2002_geometry,Kazhdan_2013_TOG_geometry} and symmetric priors \cite{ThrunW_2005_ICCV_symmetry,Mitra_2006_TOG_symmetry}, while data-driven methods were mainly based on retrieval procedures from large 3D shape databases \cite{li_2015_datadriven,Sung_2015_TOG_datadriven}. The most recent learning-based approaches learn a mapping between the partial and corresponding completed shape by exploiting voxel-grids, meshes or point-clouds. 
Voxel-based approaches exploit 3D convolution networks which lead to large computation and memory cost \cite{dai_2017_cvpr_voxel, Han_2017_iccv_voxel, Stutz_2018_IJCV_voxel}: this forces a reduction in the resolution of input data and limits the processing of fine-grained shapes.  
In \cite{groueix_2018_CVPR_mesh, wang_2018_ECCV_pixel2mesh} reference 
meshes are progressively deformed to match the target. However, this strategy does not generalize across topologies. 

Point-cloud representations are much more flexible because new points can be easily added during the learning procedure. The pioneering Point Completion Network (PCN, \cite{yuan_2018_3dv_pcn}) was based on an encoder-decoder architecture to reconstruct dense and complete point sets. TopNet \cite{Tchapmi_2019_cvpr_topnet} includes a tree-structured decoder to improve the point-cloud generation. Differently, Sarmad \etal \cite{Sarmad_2019_CVPR_rlgan} proposed a GAN-based solution where reinforcement learning is used to better control the adversarial loss function. 
Several works have also revisited both the global feature encoding process and the following local refinement.
The method proposed in  \cite{Wen_2020_CVPR_skipattention} combines a skip-attention mechanism to avoid information loss about structure details in local regions: the local geometric information is kept when encoding the original incomplete point cloud, and also used at different resolutions in the decoder.
In CRN \cite{Wang_2020_CVPR_cascaded}, the feature encoder and coarse reconstructor produce a rough complete object shape, which is then updated with points of higher resolution through subsampling from the partial input. Moreover, a feature contraction-expansion unit refines the point position gradually and is further guided by a patch based discriminator, trained to force every local region to have the same pattern as real complete point-clouds.
In MSN \cite{Liu_2020_AAAI_morphing} the refinement procedure is still based on subsampling. The input point cloud and the coarse-grained prediction are recombined to obtain an evenly distributed point cloud, and then a residual model is exploited to enable the generation of fine-grained structures.
A different solution based on extracting the 3D skeleton from the partial scan was presented in \cite{nie_2020_NIPS_skeletonbridged}. The proposed model learns the displacement from the skeletal points to the global surface space. To further preserve fidelity on observable regions, the method also includes local refinement through an adversarial patch discriminator.
The approach in \cite{Zhang_2020_ECCV_detail} tackles the completion problem by processing in a distinct way the partial known shape and its missing chunk. It uses local features to represent the known part and keep the original details, while global features are exploited for the missing part to describe the latent underlying surface. Multi-level features are extracted via a hierarchical learning architecture with gradually increasing grouping radius, inspired by \cite{yu_2018_CVPR_upsampling}. 

Very recently PF-Net \cite{Huang_2020_CVPR_pfnet} has shown how to generate exclusively the missing part with good completion performance. Multi-scale features are learned from the partial shape to get both local and global information. Then, the missing part is produced hierarchically with primary and secondary points from layers of different depth. Furthermore, an adversarial loss is included to match the distribution of predicted and real missing regions.

As in this last reference, \our combines local and global information and focuses mainly on the missing part of the shape. However, we jointly leverage the denoising task to gather local cues and contrastive learning for overall global features. Thus, we extract point features at various scales in a different way with respect to just exploiting the network activations at several depths. Moreover, our solution of involving the context of the missing region as auxiliary reconstruction objective defines a new intermediate framework between the alternatives of reconstructing the entire shape or only the missing part. In this way \our ensures a smooth blending of the generated points with the partial input: it takes advantage of the structure around the missing part, while avoiding deformations on the known points.

\section{Method}
\begin{figure*}[t]
\includegraphics[width=\textwidth]{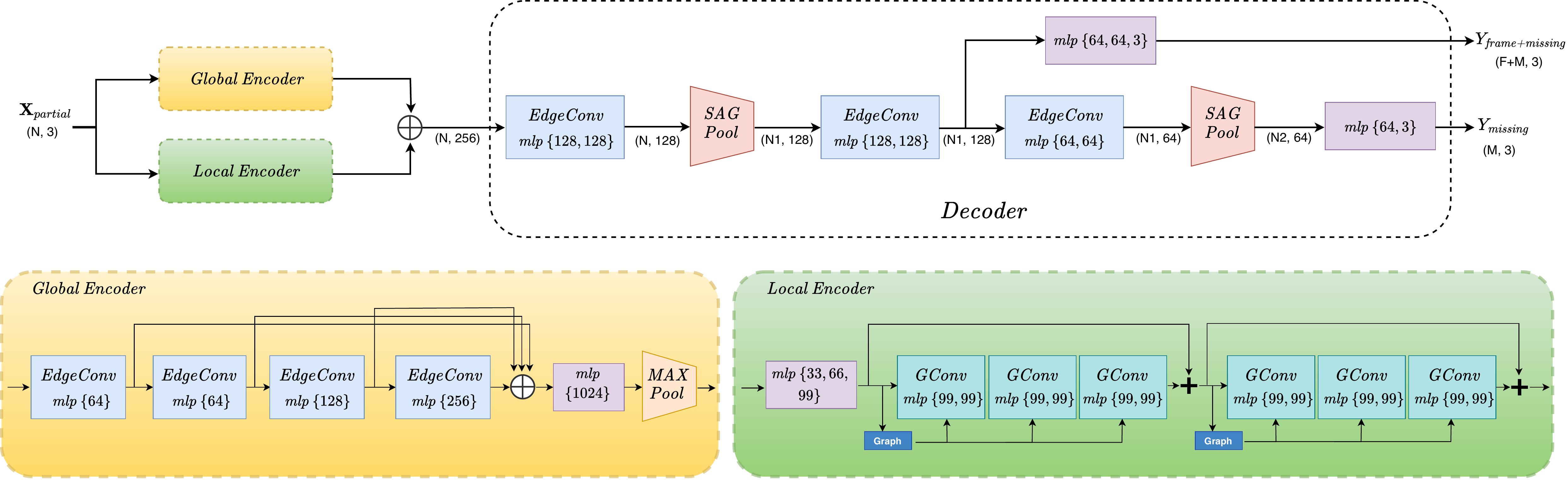}
\caption{DeCo point cloud completion. Global and local encoders extract semantic and geometric information, respectively, from the partial point cloud by pretraining with contrastive and denoising pretext tasks. The decoder converts this information into the points of the missing part. EdgeConv \cite{wang_2019_dgcnn} and GConv \cite{Pistilli_2020_ECCV_denoise} are graph convolutional layers, SAG Pool \cite{Lee_2019_icml_sagpool} is a graph pooling method. $\bigoplus$ denotes concatenation, $\textbf{+}$ denotes summation. Refer to Sec. \ref{sec:impl-details} for all the implementation details.}\vspace{-2mm}
\label{fig:architecture_scheme}
\end{figure*}

In the following we will indicate with $\bX_p$ the known partial shape, which is an $N\times3$ unordered point cloud, and with $\bX_m$ the corresponding missing part of dimension $M\times3$, with $M \leq N$. They are respectively the input and output ground truth of our \our model. 
The missing chunk is defined by starting from a random viewpoint and sorting the points in the cloud on the basis of the distance from the observer, finally dropping the closest $M$ set.
We will also use $\bX$ to refer to the original complete shape of dimension $(N+M)\times3$. To specify each point in the respective clouds we adopt lower-case letters, \eg, $\bx\in \bX$. Moreover, we indicate with $\bY$ the generated shape which is composed of $\bY_m$ and $\bY_p$: for the latter it holds $\bY_p=\bX_p$ since we keep the original partial input while seeking an estimate of the missing part.
We train our model starting from a set of $\{\bX\}_{k=1}^K$ complete point clouds which are used both for the pretext tasks that warm-up the encoders, and for the following downstream completion task. 

An overview of the proposed \our is shown in Figure \ref{fig:architecture_scheme}. We will delve into the details of its main components in the next sections. At high level, there are two parallel encoders, implemented as graph convolutional neural network. The local encoder, pre-trained with a denoising task, processes the partial shape to extract a feature vector per input point. The global encoder, pre-trained with contrastive feature learning, produces a single feature vector for the whole point cloud. The two representations are then combined and processed by a graph convolutional decoder, using pooling layers to  gradually reduce the number points and match the cardinality of the missing part.

\subsection{Local Information by Denoising}
Denoising is a highly localized task, mostly relying on low-level geometric cues that are decoupled from the global, high-level semantics and naturally shared over multiple classes.  These characteristics perfectly fit with our need of a locality prior for the category-agnostic completion model. 
We coded the task following \cite{Pistilli_2020_ECCV_denoise}, which exploits graph convolutional layers in a fully convolutional network.  
The architecture of the local encoder is shown in the bottom-right part of Figure  \ref{fig:architecture_scheme}. It is composed of residual blocks that perform graph-based operations to transform the features associated to each point. Specifically, the graph convolution aggregates features belonging to a neighborhood of limited size to maintain locality, while dynamically updating the graph via nearest neighbors in the feature space. 
With respect to the widely known Dynamic Graph CNN (DGCNN, \cite{wang_2019_dgcnn}), the solution in \cite{Pistilli_2020_ECCV_denoise} uses a lightweight Edge-Conditioned Convolution (ECC, \cite{Simonovsky_2017_cvpr_ecc}) layer, well suited for the denoising task. Besides introducing a more general definition of graph convolution, it also addresses the vanishing gradient and over-parametrization issues of the original ECC. 
Finally, a single graph convolutional layer projects the features back to the 3D space. We drop this last layer after pre-training to retain the high-dimensional feature space in the full \our achitecture.

The denoising network is trained by perturbing the input point cloud with additive white Gaussian noise and minimizing the Mean Squared Error (MSE) between the denoised point cloud $\tilde{\bX}$ and its noiseless ground truth $\bX$:
\begin{equation}
    \mathcal{L}_{MSE} = \frac{1}{N+M} \sum_{
    \mathclap{\substack{\tilde{\bx} \in \tilde{\bX} \\ \bx \in \bX}}} 
    \|\tilde{\bx}-\bx\|^2~, 
\end{equation}
with the total loss obtained by averaging the contributions over all the $K$ training samples.

\subsection{Global Information by Contrasting}
Contrasting positive from negative sample pairs is a common practice for representation and metric learning \cite{Weinberger_jmlr_2009_metric,song_CVPR16_metric}. The goal is to learn an embedding where similar examples (positive pairs) are mapped close to each other, and dissimilar examples (negative pairs) are mapped far apart. Recently, there has been a shift in the pairs definition, moving from an assignment based on the original sample class label to instance identity \cite{Dosovitskiy_nips_2014_instanceContrast,wu_2018_cvpr_instance}.
Indeed, it has been shown that treating each sample as a class and exploiting data augmentation to create surrogate data pairs allows to get discriminative features from unlabeled data \cite{chen_2020_icml_simclr, He_2020_CVPR_moco}. Inspired by this literature, we code the global point cloud shape information via contrastive learning and we show the corresponding architecture in the bottom-left part of Figure  \ref{fig:architecture_scheme}. Specifically, given a randomly sampled mini-batch of point clouds, each one is augmented four times using a combination of rotation (y-axis), random scaling and jittering. All the transformed versions are also randomly cropped. These new variants $\{\bX_{4k}, \bX_{4k-1}, \bX_{4k-2}, \bX_{4k-3}\}$ are provided as input to four stacked EdgeConv blocks \cite{wang_2019_dgcnn}. The obtained convolved representations at different depths are concatenated and further processed by an MLP layer. The global shape feature vector is finally obtained by performing max pooling. A further MLP head (two hidden layer inter spaced by ReLU) is used to project the global shape feature vector to a lower dimensional space (128).
This yields a representation per sample variant $\mathcal{P}(k)=\{{\bz}_{4k}, {\bz}_{4k-1}, {\bz}_{4k-2}, {\bz}_{4k-3}\}$ that enters a generalized version of the the Normalized Temperature-scaled cross entropy loss (NT-Xent):
 \begin{equation}
    \mathcal{L}_{NT-Xent} = \frac{-1}{|\mathcal{P}(i)|} \sum_{j\in\mathcal{P}(i)}\log \frac{\text{exp}(\text{sim}(\bz_i,\bz_j)/\tau)}{\sum_{k=1}^{4K}\mathbbm{1}_{[k\neq i]}\text{exp}(\text{sim}(\bz_i,\bz_k)/\tau)}~. 
\end{equation}
Here $\text{sim}(\bz_i,\bz_j)$ is the cosine similarity between two feature vectors, and the final loss is computed across all positive pairs within a quadruplet, while considering, as negatives all the remaining transformed samples of the mini-batch. Following standard practice \cite{chen_2020_icml_simclr}, we include the MLP projection head only during the global encoder pre-training, while it is removed in the final \our architecture.

This task is well suited for our global encoder: it promotes a robust semantic embedding by learning instance representations that are close to each other regardless of which portion of the point cloud is missing, and thus capturing a global understanding of it. Critically, it does not require supervision in the form of class labels.

\subsection{Framing and Reconstructing the Missing Part}
The information collected by the local and global encoders are finally combined to guide the generation of the missing shape part. The two obtained feature embeddings are aggregated and fed as input to the following decoder network. Our decoder is composed by three EdgeConv layers \cite{wang_2019_dgcnn} and two Self-Attention Graph Pooling layers (SAG Pool, \cite{Lee_2019_icml_sagpool}), whose purpose is to reduce the number of points down to the number of points of the missing part. As can be noticed from the top part of Figure \ref{fig:architecture_scheme} the decoder has two outputs at different levels. The final head is defined by an MLP that generates $\bY_m$. The intermediate head positioned between the two central EdgeConv layers steers the feature space to correctly represent the region around the missing part as well as the missing part itself. 

Specifically, starting from the sorted points used for the definition of the missing chunk $\bX_{m}$, we extend our attention to the following set of $F$ point in the same list to define the \emph{frame + missing region} $\bX_{fm}$ of dimension $(F+M) \times 3$.
We regularize training by constraining the decoder to generate an estimate of the missing part $\bY_{m}$ consistent with the ground truth $\bX_m$, and to reconstruct correctly the frame and the missing part as $\bY_{fm}$ from the intermediate head. For both objectives, the training procedure minimizes the Chamfer Distance (CD) loss: 
\begin{gather}
    {\mathcal{L}_{CD} = \frac{1}{2M}\Big\{\sum_{\bx\in\bX_m} \min_{\by \in \bY_m}\|\bx-\by\|_2^2 + \sum_{\by\in\bY_m} \min_{\bx \in \bX_m}\|\by-\bx\|_2^2 \Big\}} \nonumber \\
   + \frac{1}{2(M+F)} \Big\{\sum_{\bx\in\bX_{fm}} \min_{\by \in \bY_{fm}}\|\bx-\by\|_2^2 + \sum_{\by\in\bY_{fm}} \min_{\bx \in \bX_{fm}}\|\by-\bx\|_2^2 \Big\}~. 
\end{gather}
During testing we do not have control on the exact nature of the missing shape part and the output of the intermediate head is neglected.

\subsection{Implementation Details}
\label{sec:impl-details}
We designed the \emph{local encoder} architecture on the basis of a graph constructed by searching dynamically for the $k$-nearest neighbor ($k=8$) of each point in terms of Euclidean distance between their feature vectors. For pre-training we used shapes corrupted by Gaussian noise with average set to $0$ and standard deviation equal to 0.02. The \emph{global encoder} has four stacked EdgeConv layers, each with $k=24$ nearest neighbors. For pre-training we used shapes with random crops of $25\%$ ($512$ points out of point clouds of $N=2048$).
We set the temperature scaling parameter in the NT-Xent loss $\tau=0.5$.
The \emph{decoder} alternates EdgeConv layers with $k=16$ and two attention-based pooling layers, respectively based on graphs with $k=16$ and $k=6$ neighbors. The intermediate feature dimensions indicated in Figure  \ref{fig:architecture_scheme} are $N_1=1280$ and $N_2=512$ with $M=F=512$.  

We trained one single network over all 13 known object categories for 240 epochs with a batch size of 30. 
We used Adam \cite{adam} with initial learning rate set to 0.001, halved every 25 epochs for the (pre-trained) encoder and every 40 epochs for the decoder. We implemented all the network using  PyTorch and train it on two NVIDIA Titan RTX GPUs (PyTorch DataParallel) with CUDA 10.0.
The model is finally tested on a single GPU with batch size 64.
Ablation experiments ran on hpc cluster with NVIDIA V100 GPUs.

The code of \our is available at 
\url{https://github.com/antoalli/Deco}.

\section{Experiments}

\subsection{Dataset, Baselines and Evaluation Metric}
To evaluate the proposed \our  we follow the experimental protocol of \cite{Huang_2020_CVPR_pfnet}, selecting 13 object classes from the benchmark dataset Shapenet-Part \cite{Yi_ACM_2016_shapenetpart}. The total number of shapes is 14473 with 11705 used for training and 2768 for testing. All the input point cloud data is centered at the origin and their coordinate are normalized to $[-1,1]$. The ground truth is created by sampling 2048 points uniformly on each shape.
For the novel categories we selected 12 objects classes from Shapenet-Core \cite{Chang_shapenet_2015}, getting a total of 7873 shapes. In the chosen set, six classes are semantically related to the seen ones (bicycle-motorbike, basket-bag, helmet-cap, bowl-mug, rifle-pistol, vessel-airplane), while the remaining six (piano, bookshelf, bottle, clock, microwave, telephone) were chosen randomly.

We compare against several recent completion methods: PCN \cite{yuan_2018_3dv_pcn}, MSN \cite{Liu_2020_AAAI_morphing}, CRN \cite{Wang_2020_CVPR_cascaded}, as well as two variants of PF-Net \cite{Huang_2020_CVPR_pfnet} with and without (vanilla version) its adversarial discriminator. For all these baselines we ran the code provided by the authors to get both the quantitative and qualitative results. To keep a fair comparison on 2048 points, for CRN we consider a single iteration through its refinement sub-network.
We quantitatively assess the performance of all the methods by using the Chamfer Distance (CD) on the reconstructed missing part. More precisely, we follow the evaluation strategy already validated by PF-Net: for the methods predicting the overall shape we report the CD computed only on the $M$ closest points to the crop centroid. We present the results per class and the overall average CD on all the test shapes.

\begin{table}[t!]
    \hspace{-2mm}
    \resizebox{0.49\textwidth}{!}{
    \begin{tabular}{l@{~~}|@{~~}c@{~~~~}c@{~~~~}c@{~~~~}c@{~~~~}c@{~~~~}c}
\hline
\multirow{2}{*}{\textbf{Category}} & {\textbf{PCN}}   & {\textbf{MSN}}   &  {\textbf{CRN}}   & {\textbf{PF-Net}}  & {\textbf{PF-Net}}   &   \multirow{2}{*}{\textbf{\our}} \\
 & \cite{yuan_2018_3dv_pcn} &  \cite{Liu_2020_AAAI_morphing} &   \cite{Wang_2020_CVPR_cascaded} &  \textbf{vanilla} \cite{Huang_2020_CVPR_pfnet} &  \cite{Huang_2020_CVPR_pfnet} &    \\
\hline
Airplane & 31.515 & 15.907 & 39.334 & 11.015 & 10.805 & \textbf{10.003} \\
Bag & 37.825 & 59.185 & 33.593 & 40.000 & 38.485 & \textbf{28.508} \\
Cap & 66.275 & 40.276 & 53.146 & 49.945 & 50.450 & \textbf{36.436}\\
Car & 24.320 & 24.176 & 39.537 & 21.925 & \textbf{21.640} & 22.963\\
Chair & 31.265 & 20.751 & 28.688 & 19.130 & 19.490 & \textbf{16.428}\\
Lamp & 93.745 & 41.094 & 30.207 & 41.555 & 42.910 & \textbf{24.150} \\
Laptop & 22.460 & 11.718 & 26.393 & 11.520 & \textbf{11.220} & 12.706 \\
Motorbike & 34.420 & 21.276 & 41.292 & 20.525 & 19.905 & \textbf{19.136}\\
Mug & 35.905 & 57.007 & 41.153 & 32.800 & \textbf{31.880} & 34.239 \\
Pistol & 29.490 & 14.560 & 26.845 & 11.395 & \textbf{10.885} & 12.266\\
Skateboard & 23.815 & 14.146 & 34.358 & 12.275 & 12.365 & \textbf{9.861} \\
Table & 24.775 & 22.103 & 23.953 & 20.560 & 20.845 & \textbf{17.120}\\
Guitar & 10.540 & 6.959 & 15.224 & 4.350 & \textbf{4.425} & 4.482\\
\hline
Overall & 34.095 & 22.410 & 29.044 & 20.209 & 20.445 & \textbf{16.517} \\
\hline
\end{tabular}
}
\caption{\emph{Known Categories - Quantitative}. Chamfer Distance on the missing region of point clouds scaled by $10^4$. The lower, the better.}
\label{table:single-hole-quantitative}
\vspace{-2mm}
\end{table}

\begin{figure*}[t]
\centering
\begin{tabular}{@{~}c@{~~}c@{~~}c@{~~}c@{~~}c@{~~}c@{~~} c@{~~}c@{~~}c@{~}}
    Input & PCN & MSN & CRN &  PF-Net vanilla & PF-Net & \our & GT \\
    \includegraphics[width=0.1\textwidth]{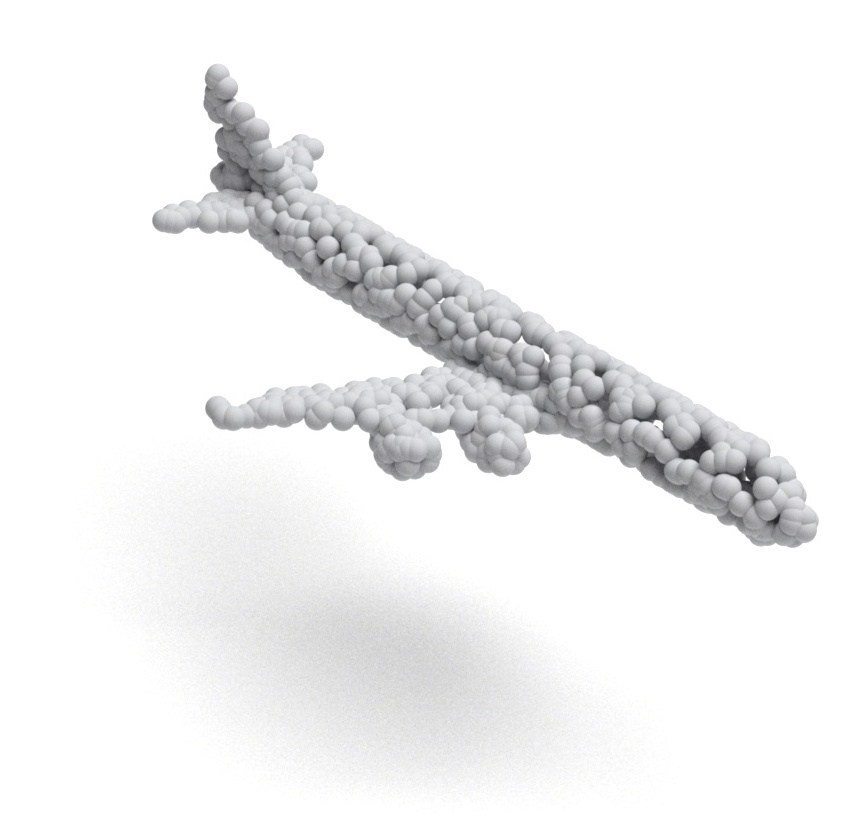} & 
    \includegraphics[width=0.1\textwidth]{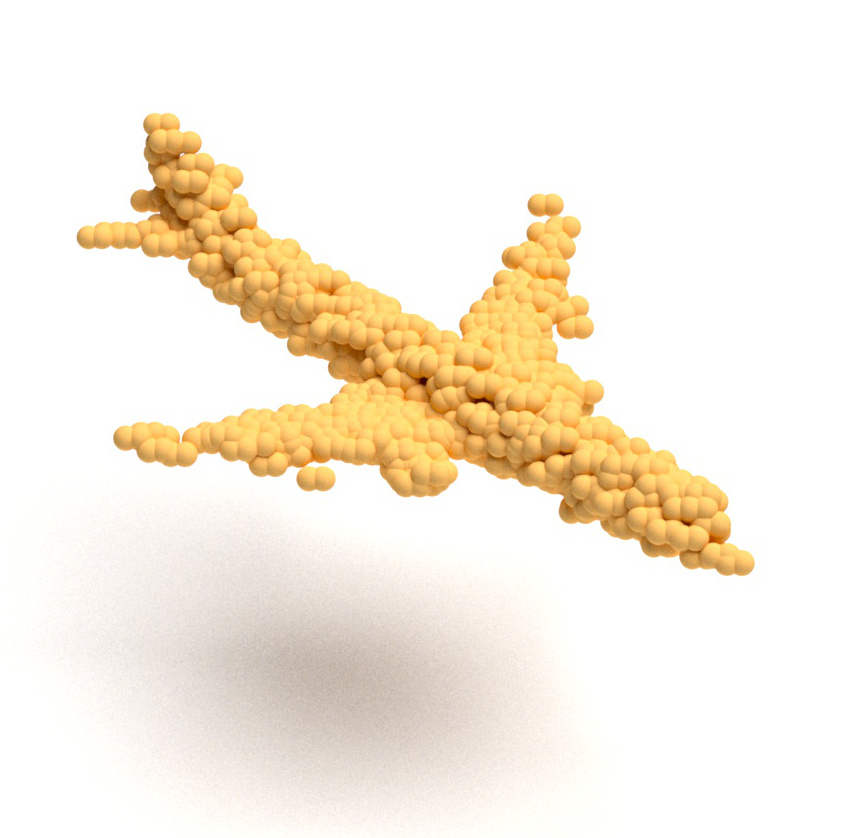} 
    &\includegraphics[width=0.1\textwidth]{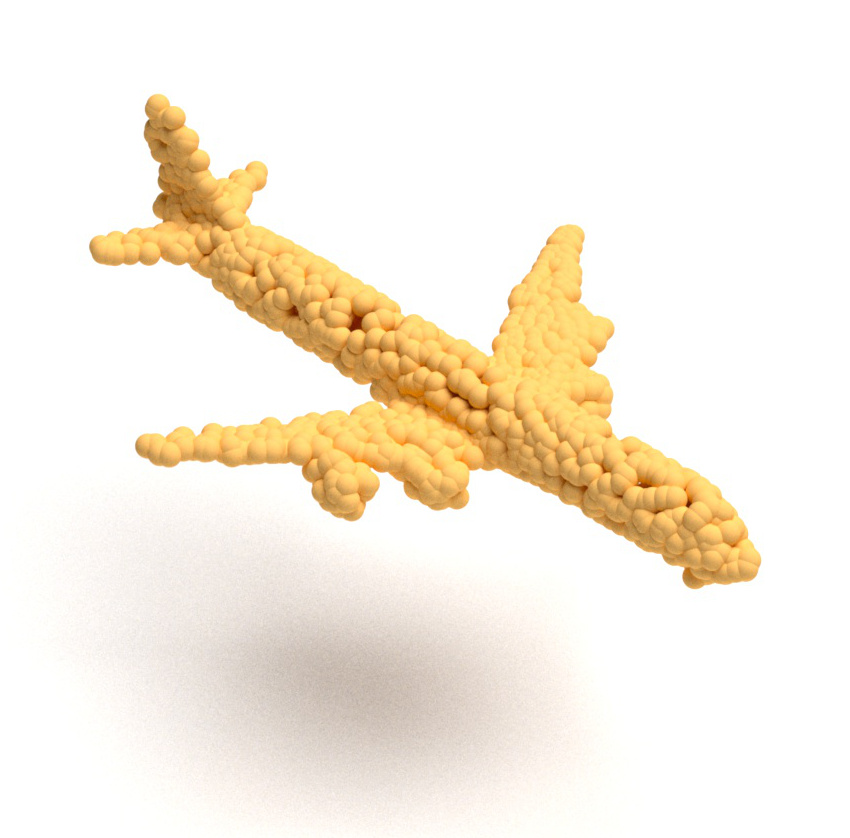}
    & \includegraphics[width=0.1\textwidth]{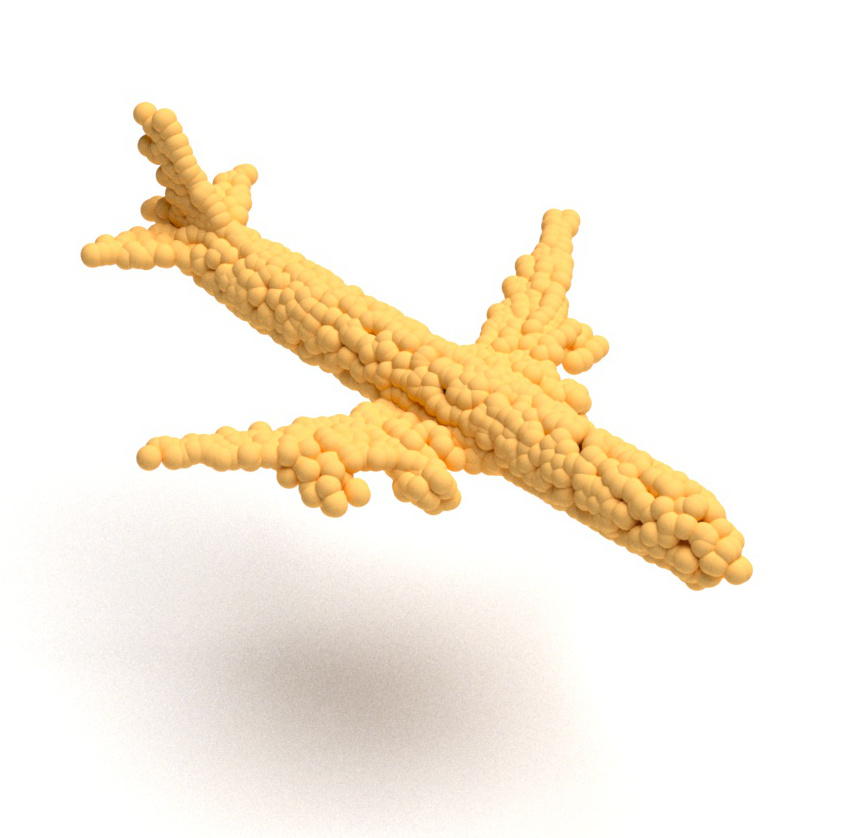} 
    & \includegraphics[width=0.1\textwidth]{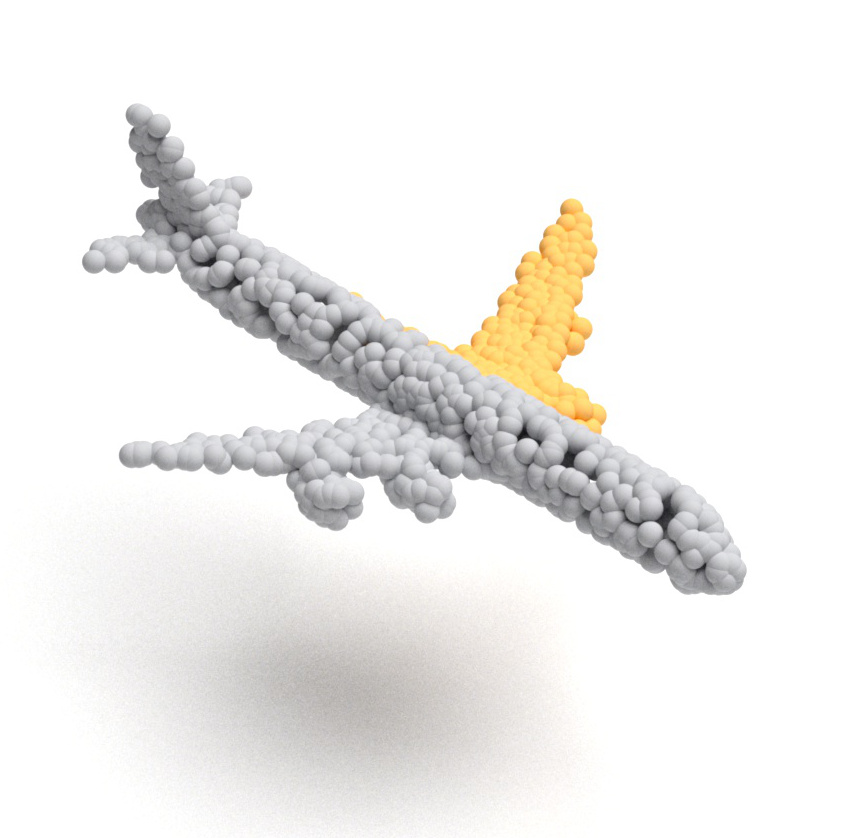} &
    \includegraphics[width=0.1\textwidth]{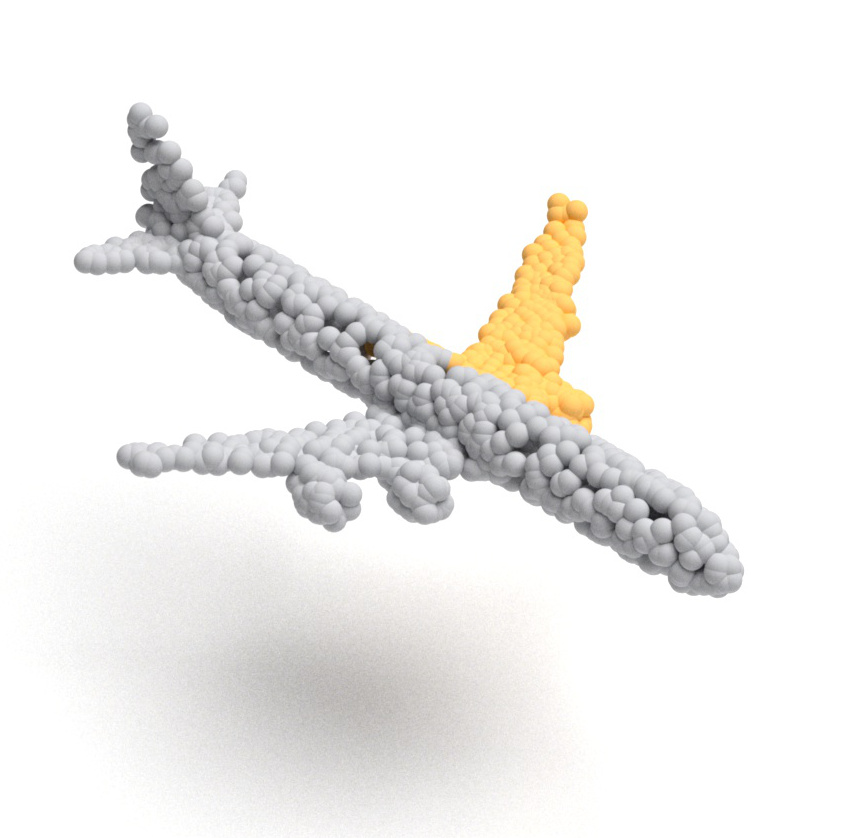} &
    \includegraphics[width=0.1\textwidth]{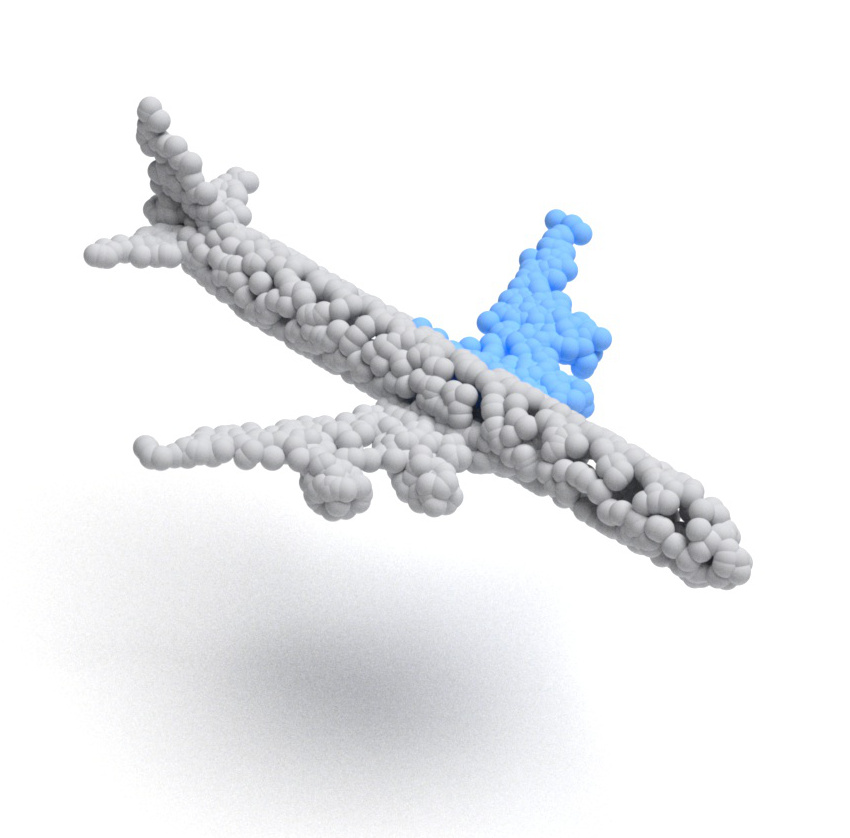} &
    \includegraphics[width=0.1\textwidth]{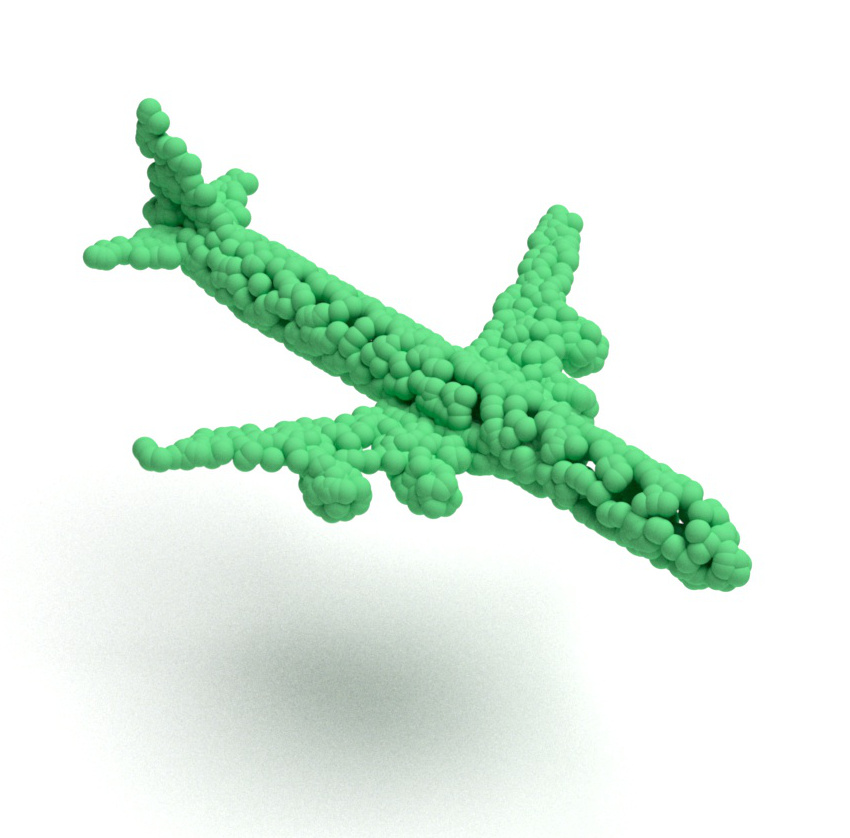}\\
    \includegraphics[width=0.1\textwidth]{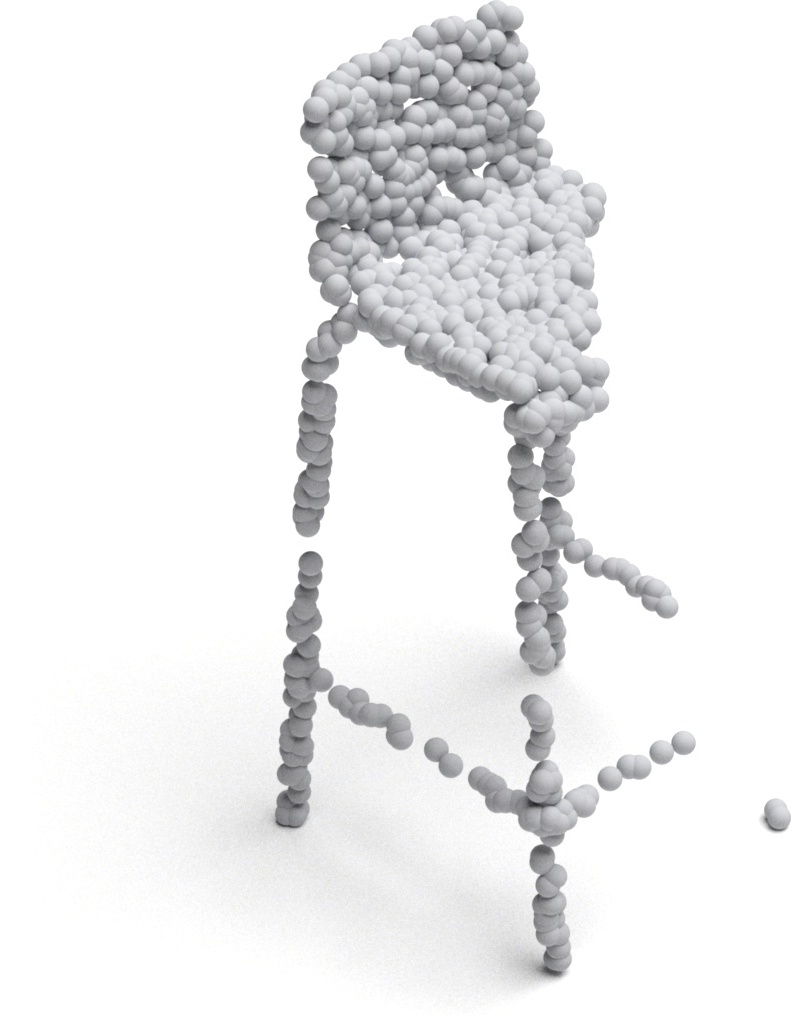} & 
    \includegraphics[width=0.1\textwidth]{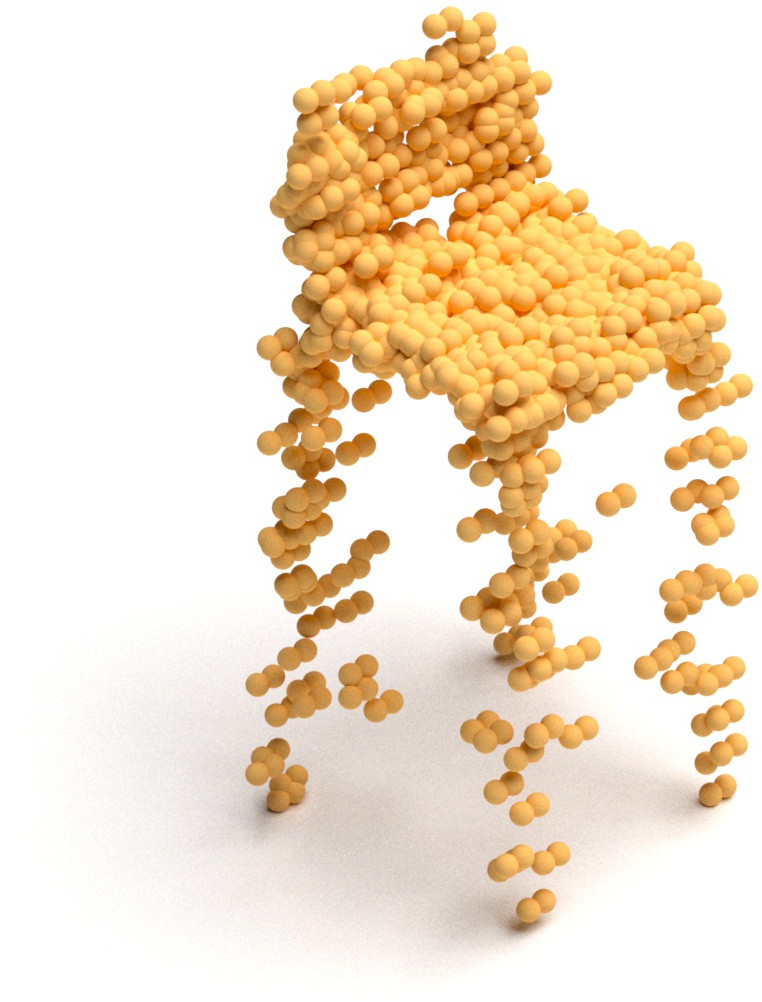} 
    & \includegraphics[width=0.1\textwidth]{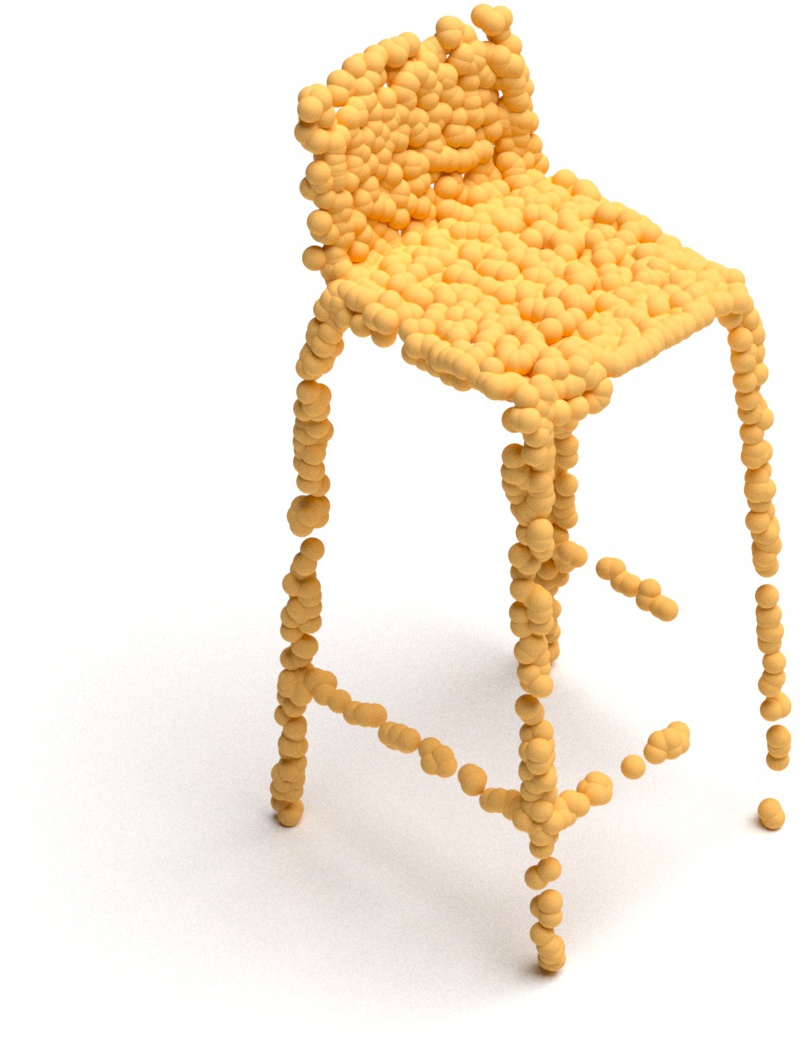}
    & \includegraphics[width=0.1\textwidth]{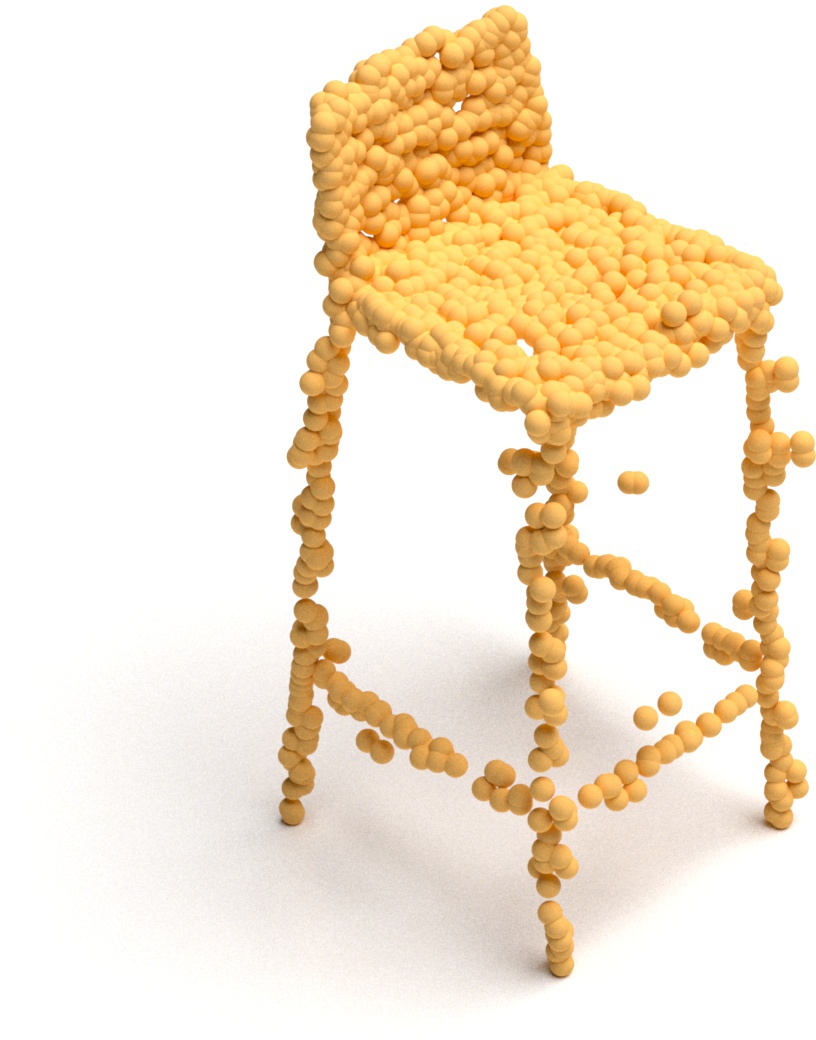} 
    & \includegraphics[width=0.1\textwidth]{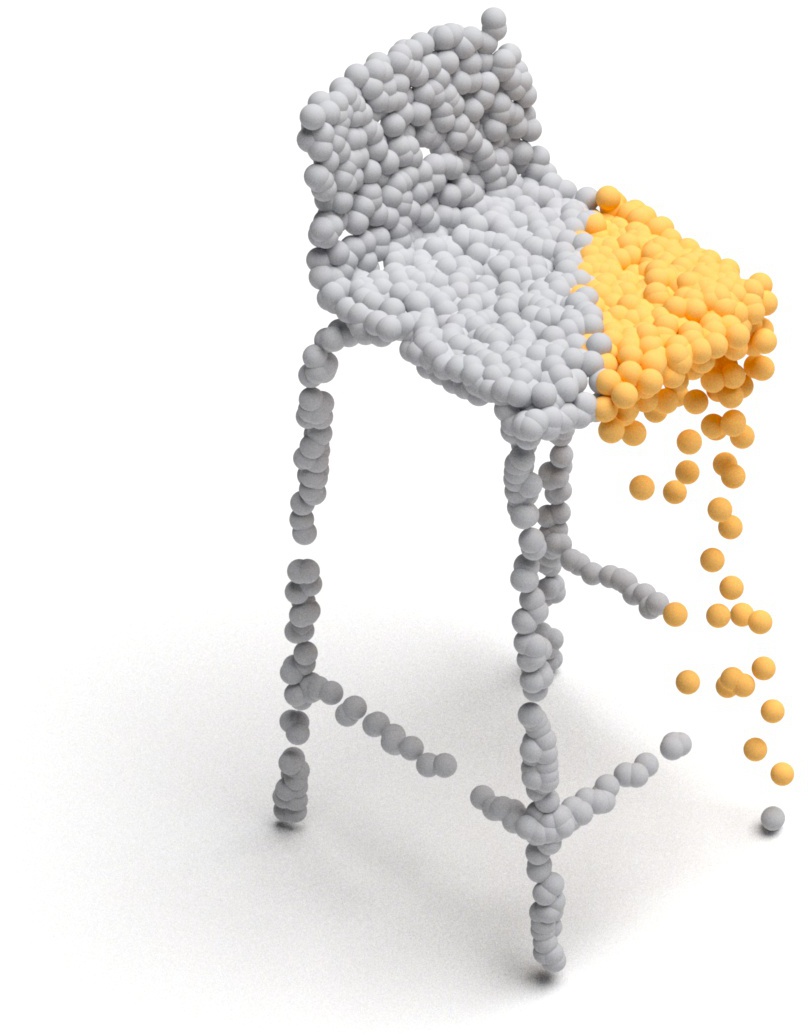} &
    \includegraphics[width=0.1\textwidth]{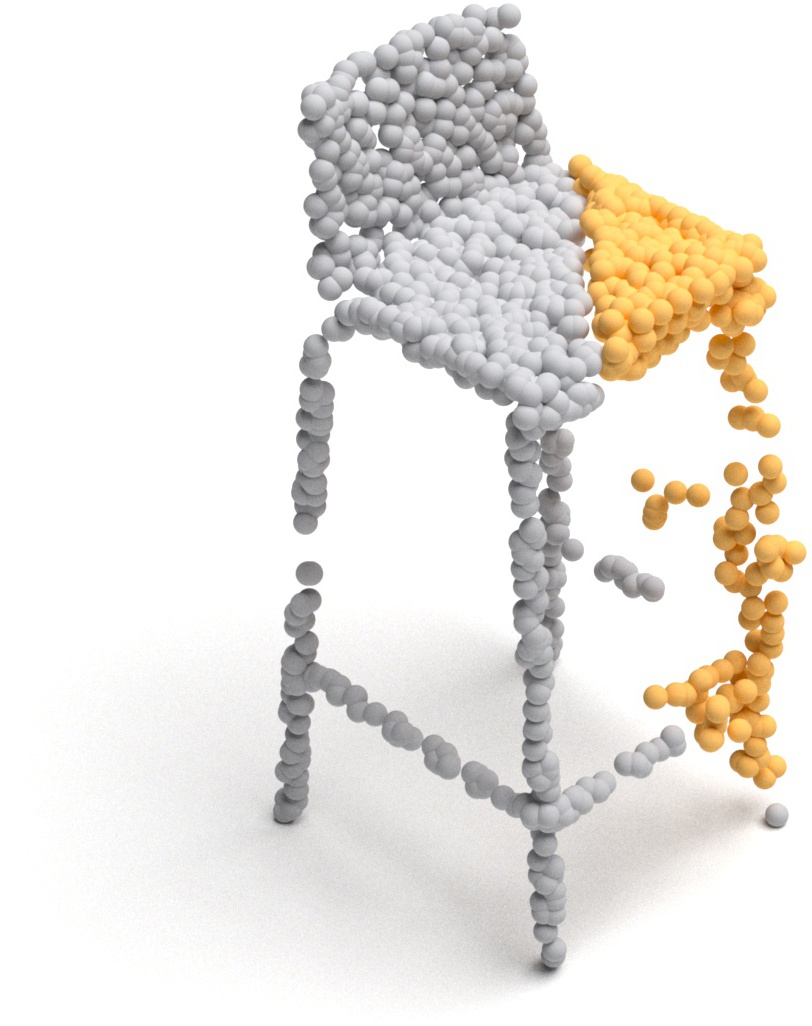} &
    \includegraphics[width=0.1\textwidth]{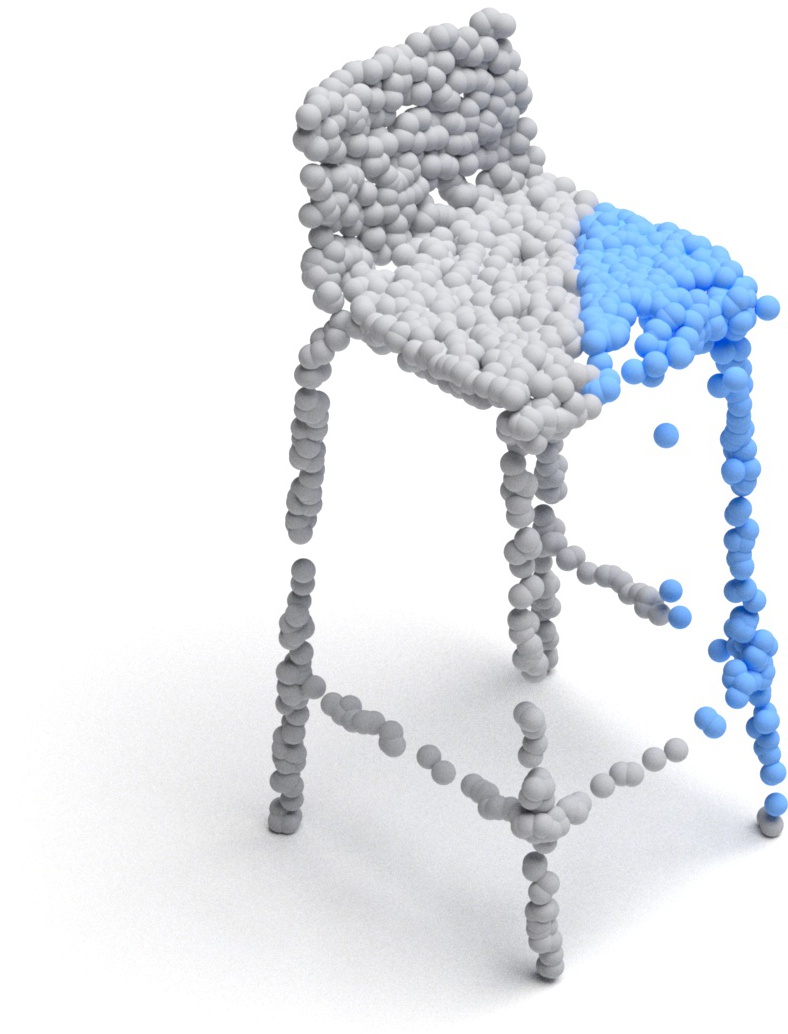} &
    \includegraphics[width=0.1\textwidth]{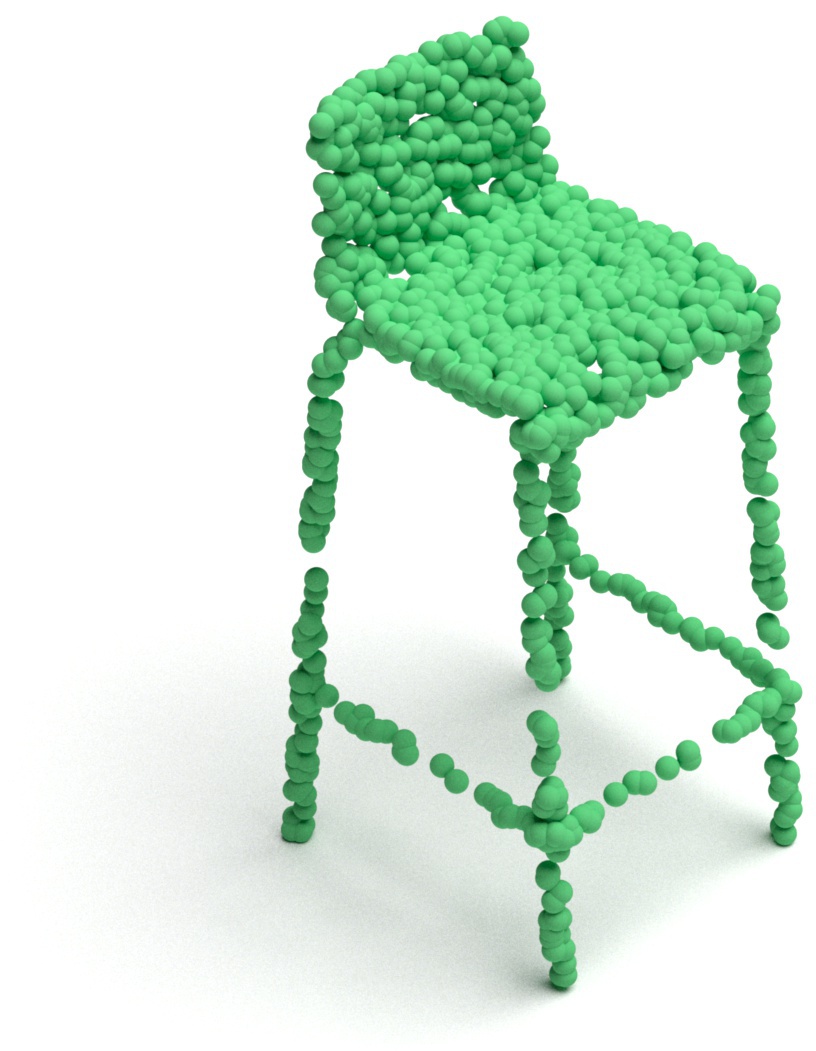}\\
    \includegraphics[width=0.1\textwidth]{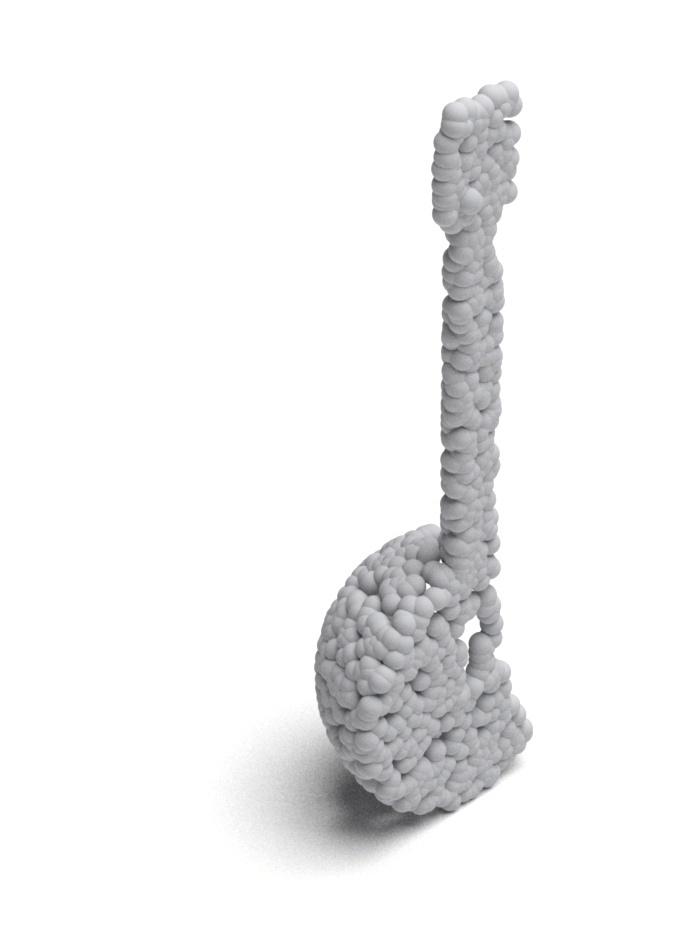} & 
    \includegraphics[width=0.1\textwidth]{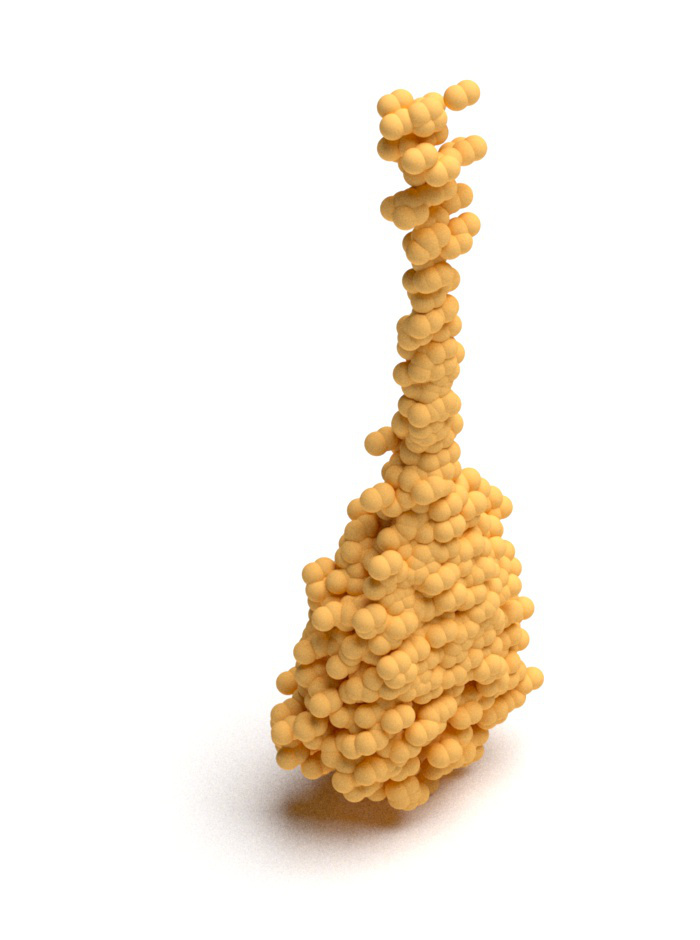}
    &\includegraphics[width=0.1\textwidth]{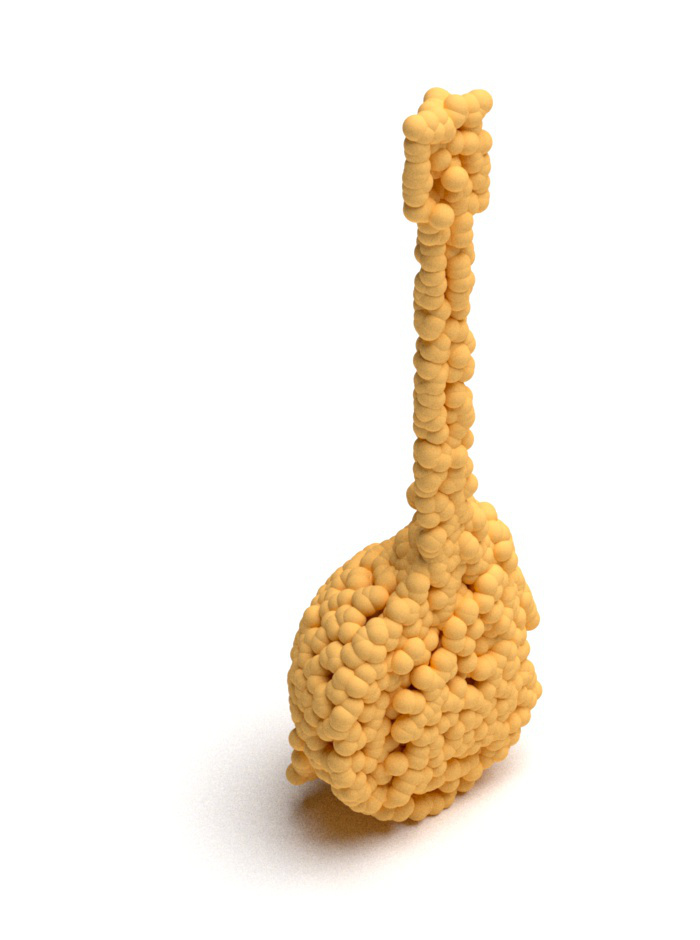}
    & \includegraphics[width=0.1\textwidth]{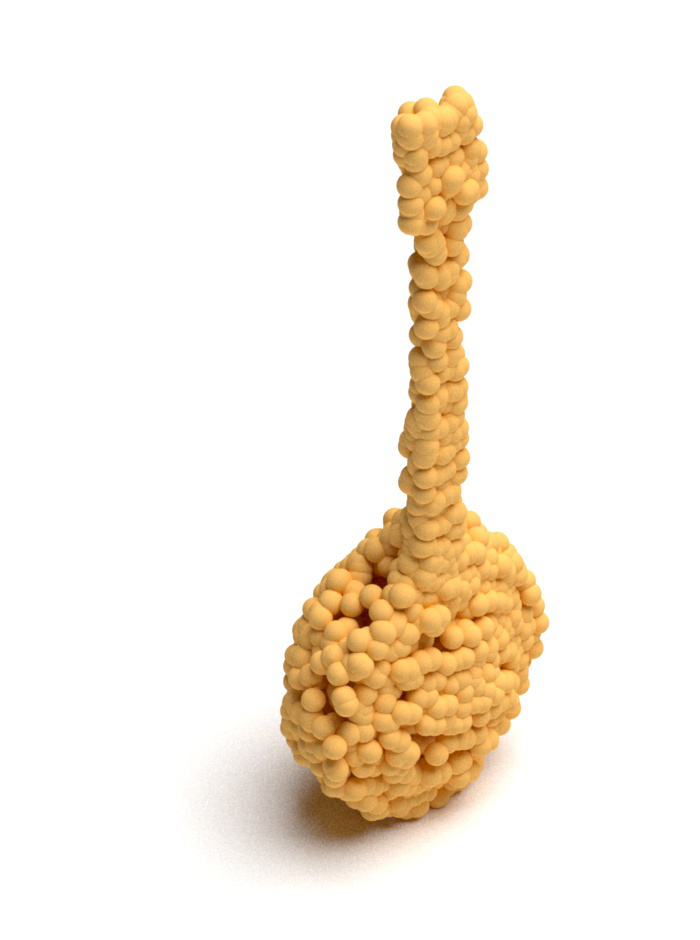} 
    & \includegraphics[width=0.1\textwidth]{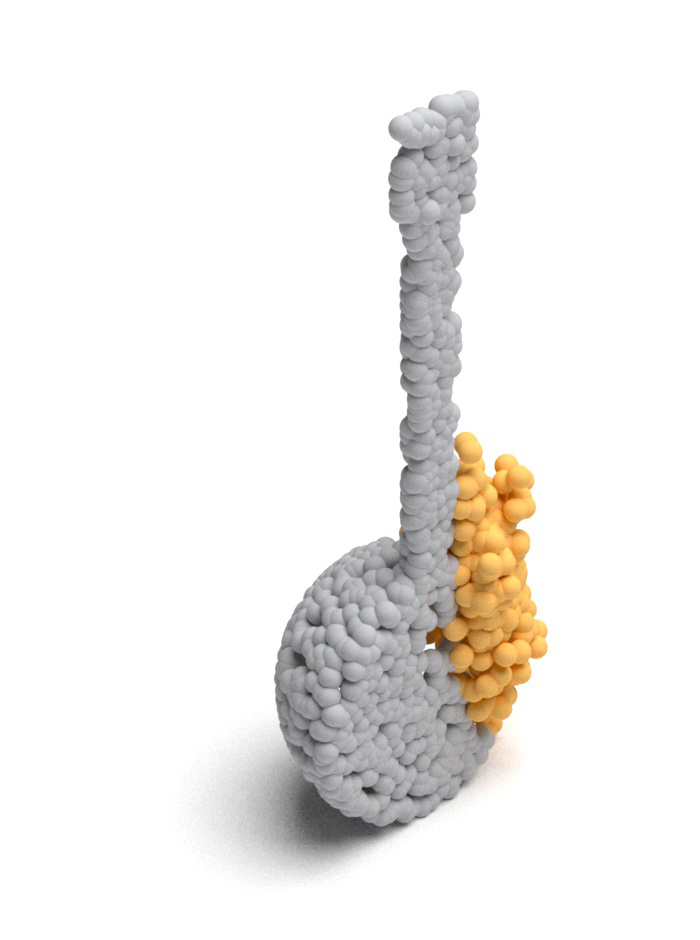} &
    \includegraphics[width=0.1\textwidth]{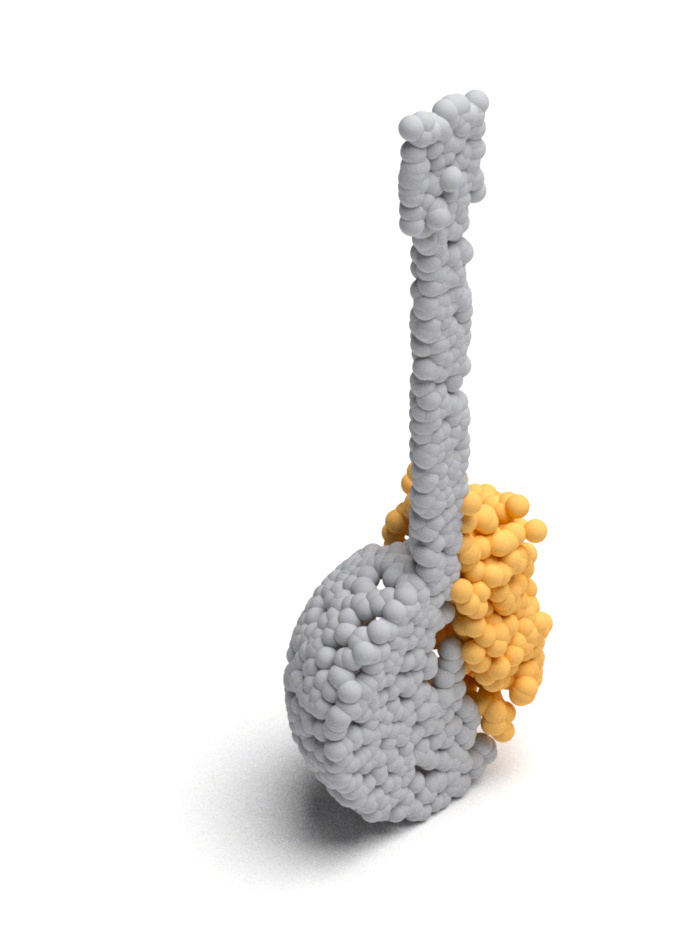} &
    \includegraphics[width=0.1\textwidth]{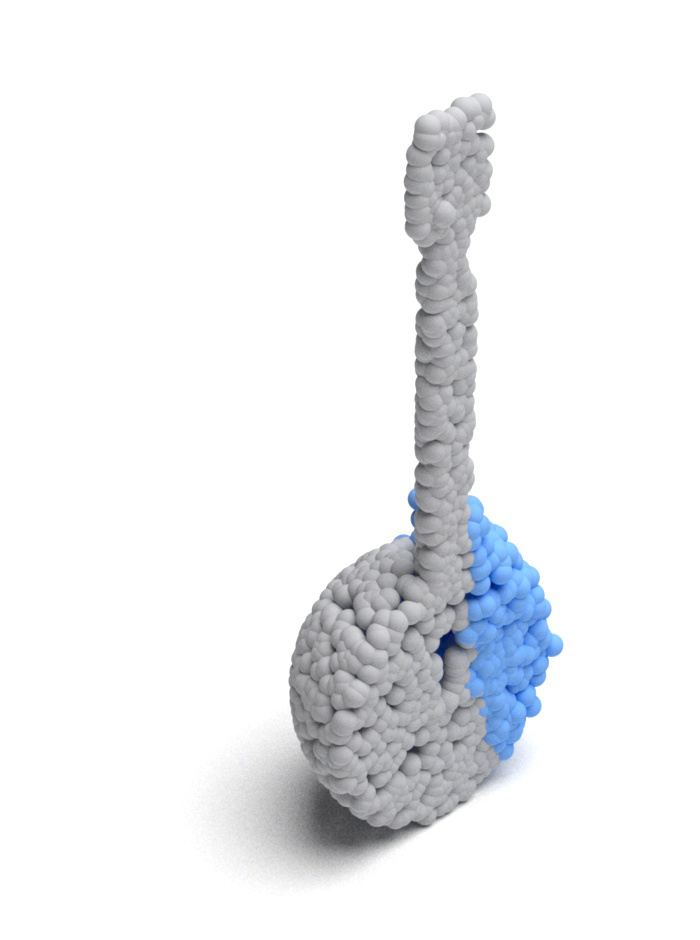} &
    \includegraphics[width=0.1\textwidth]{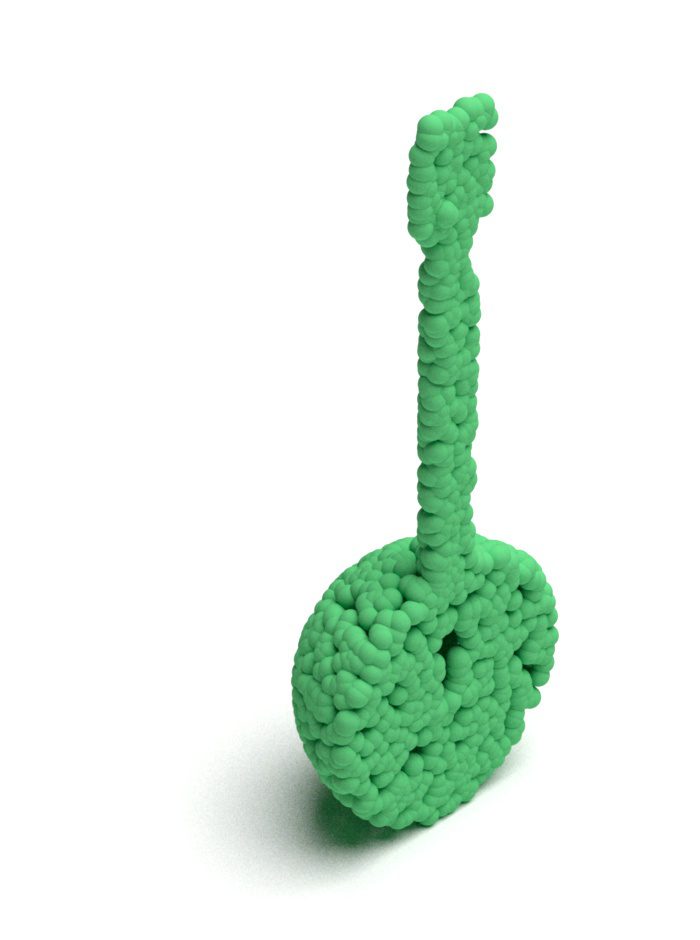}\\
    \includegraphics[width=0.1\textwidth]{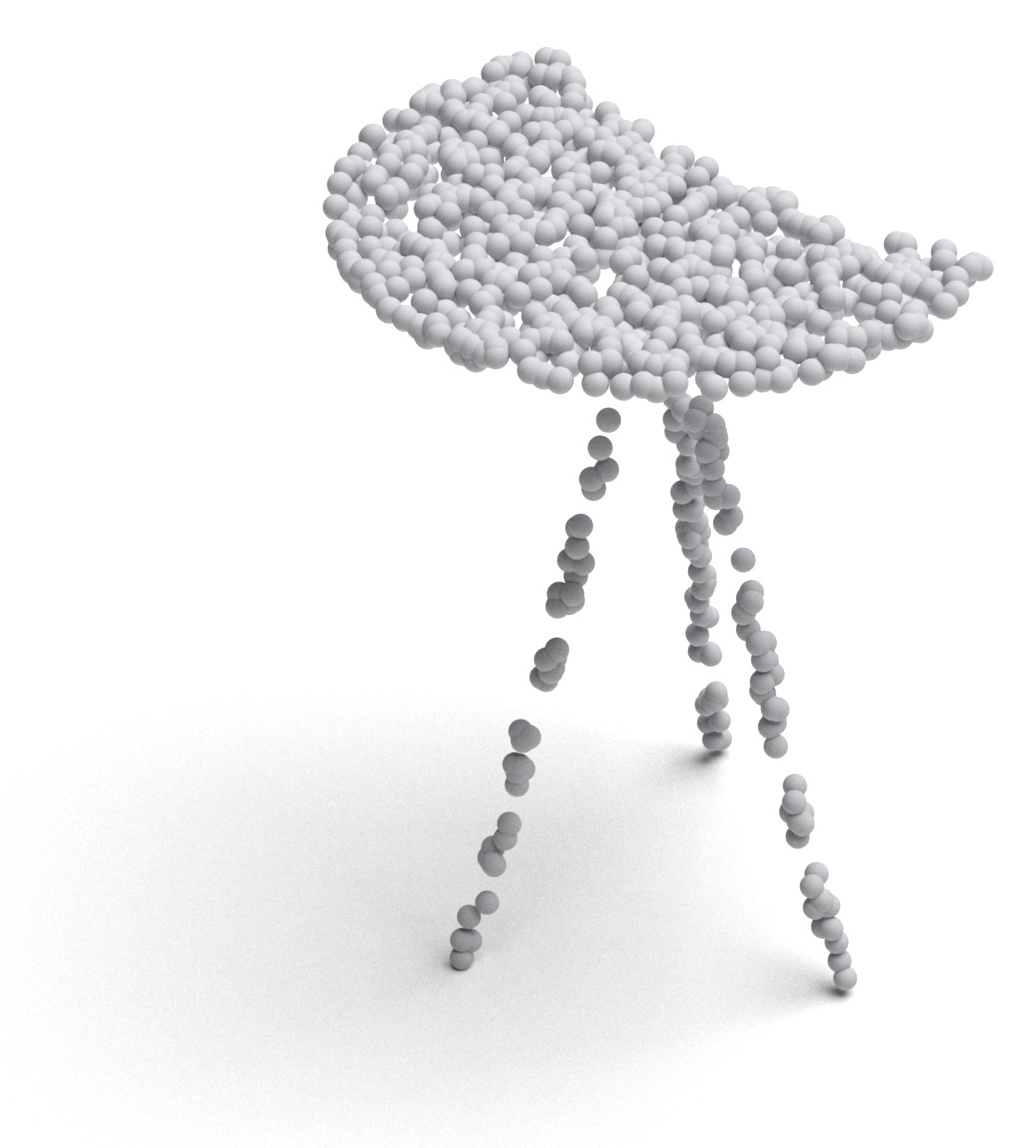} & 
    \includegraphics[width=0.1\textwidth]{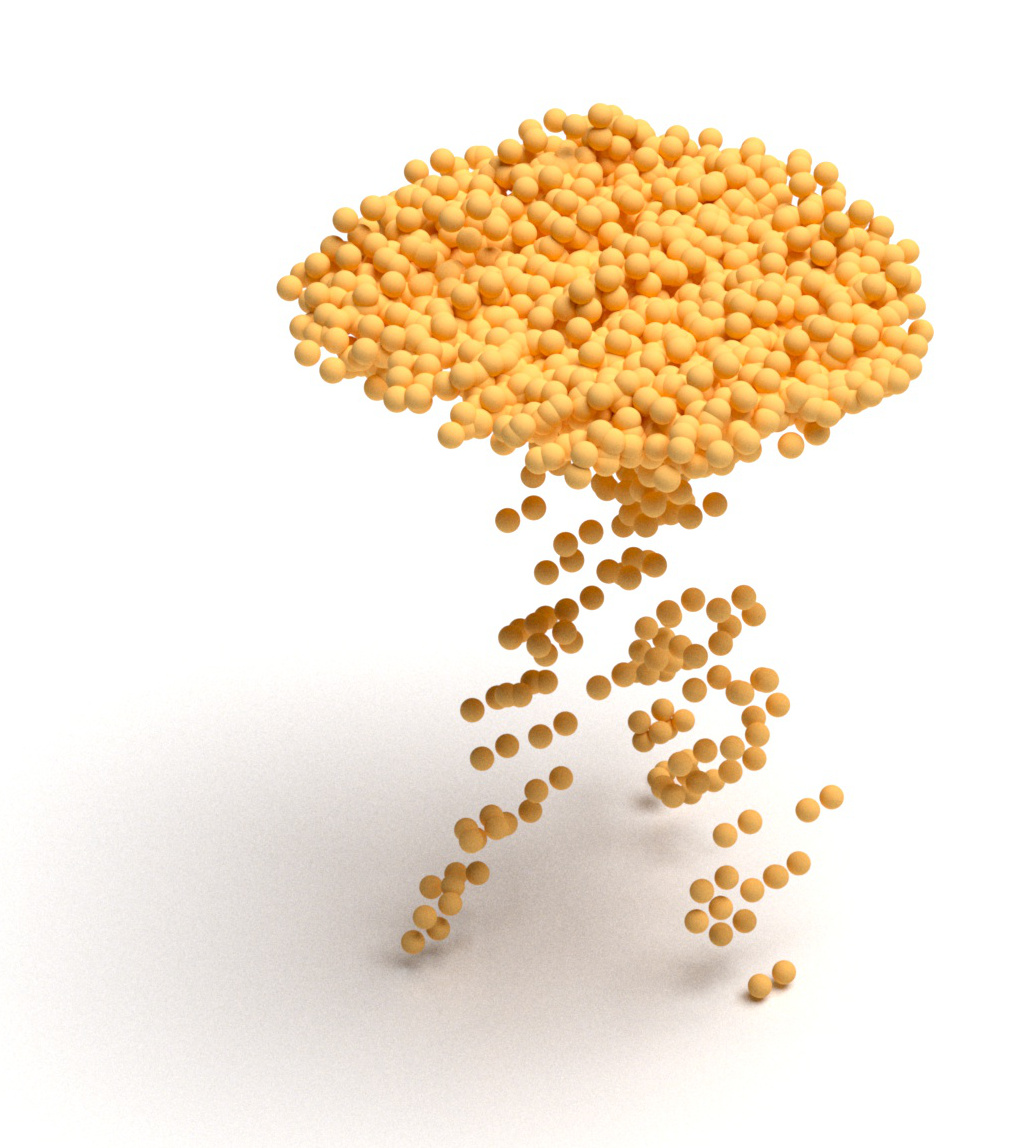}
    &\includegraphics[width=0.1\textwidth]{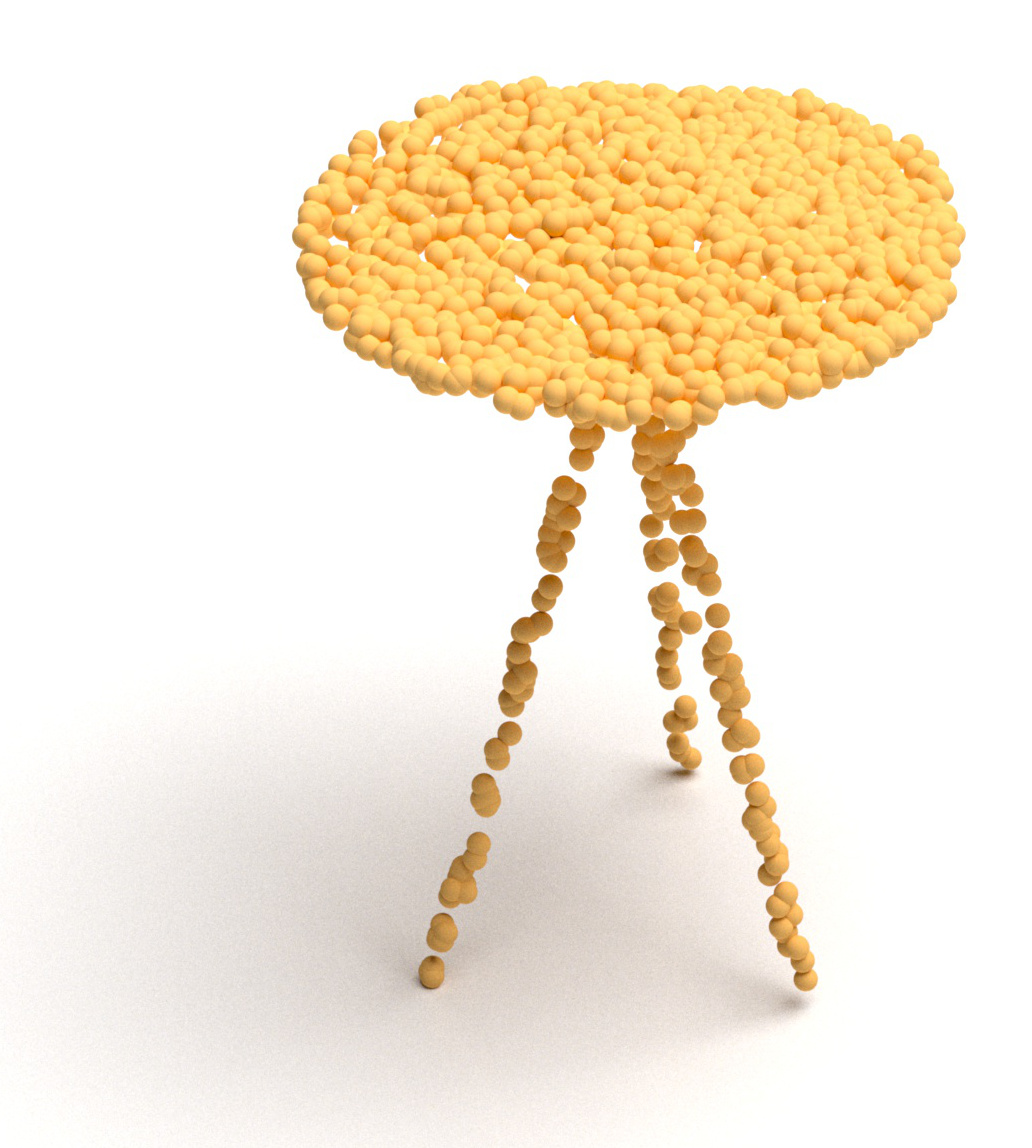}
    &\includegraphics[width=0.1\textwidth]{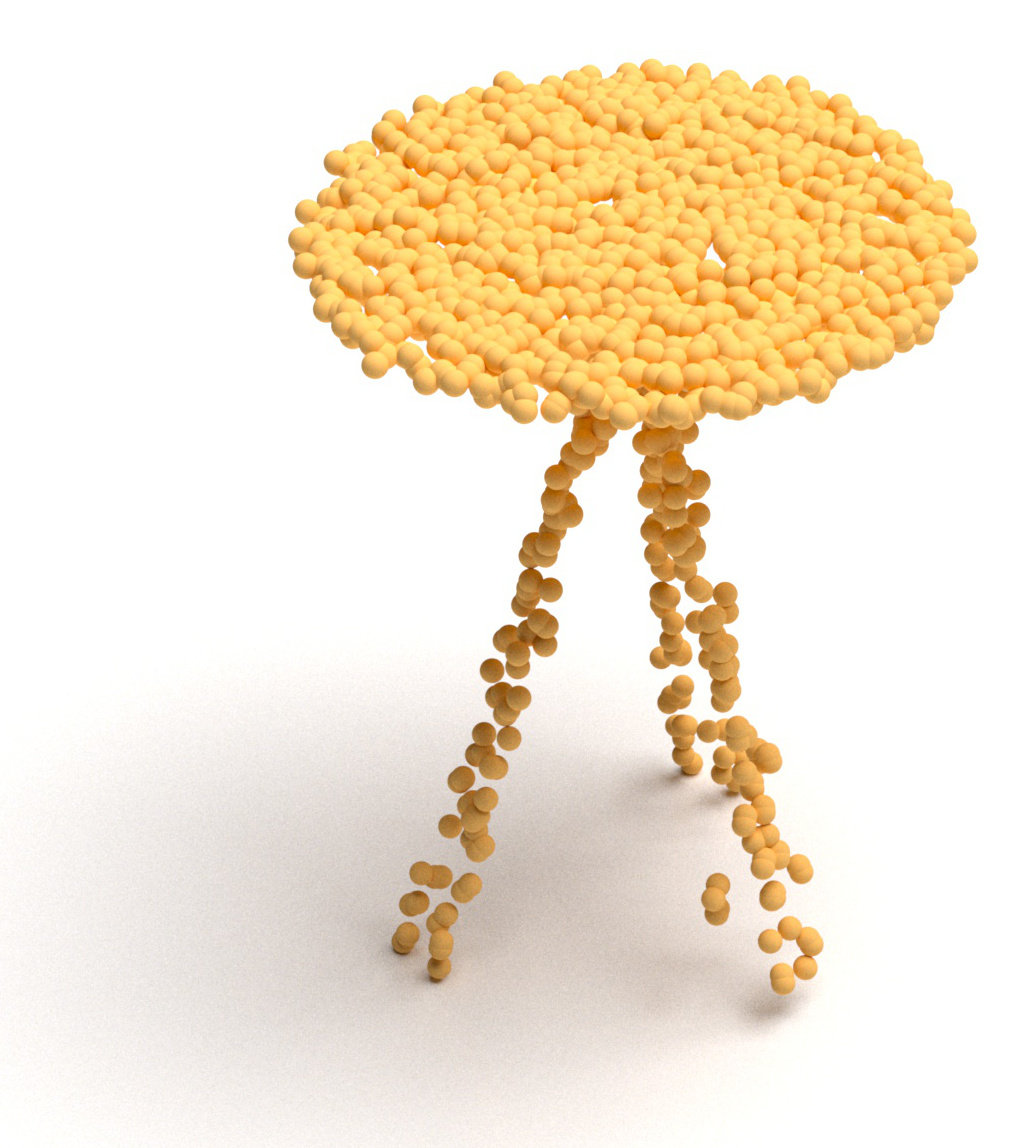} 
    &\includegraphics[width=0.1\textwidth]{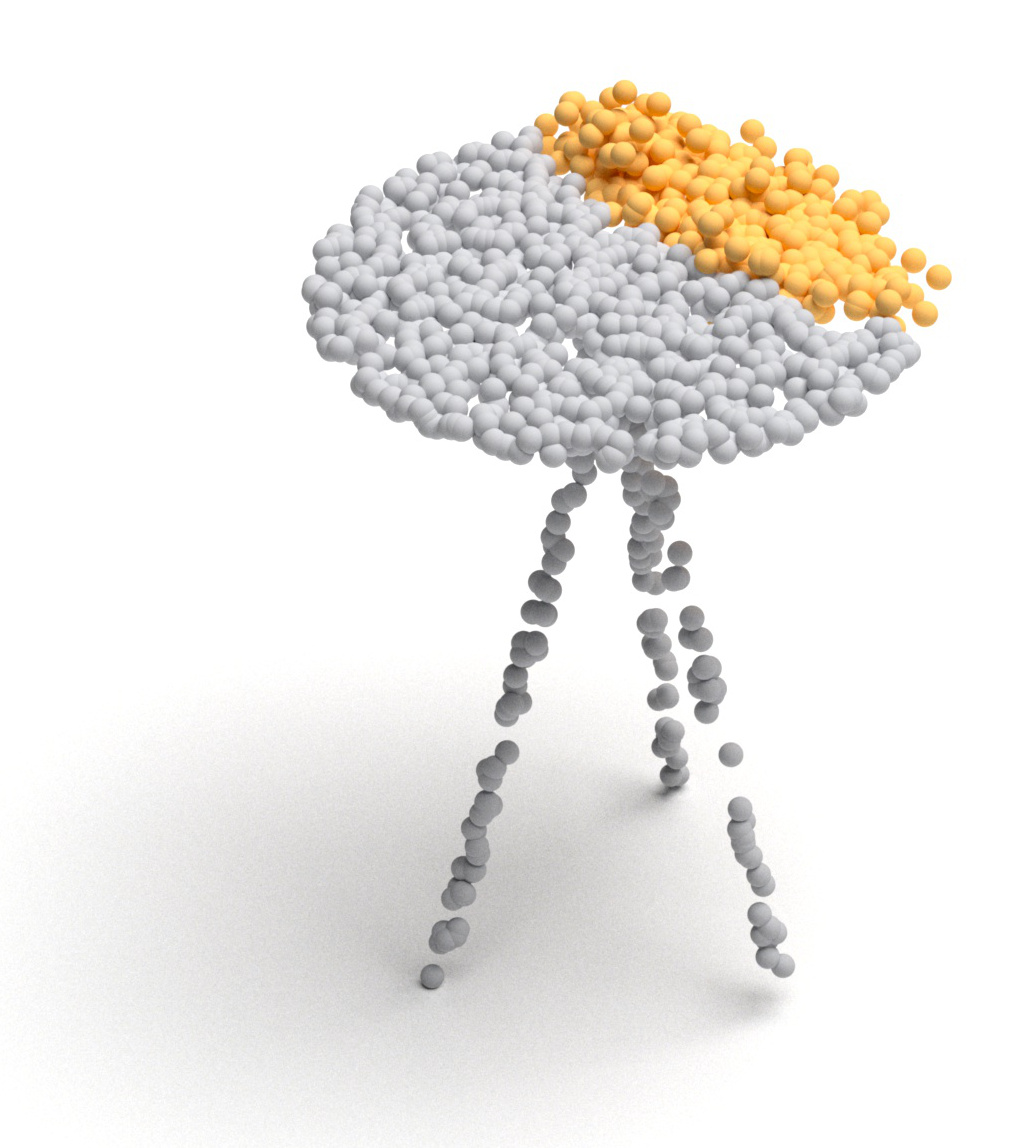} &
    \includegraphics[width=0.1\textwidth]{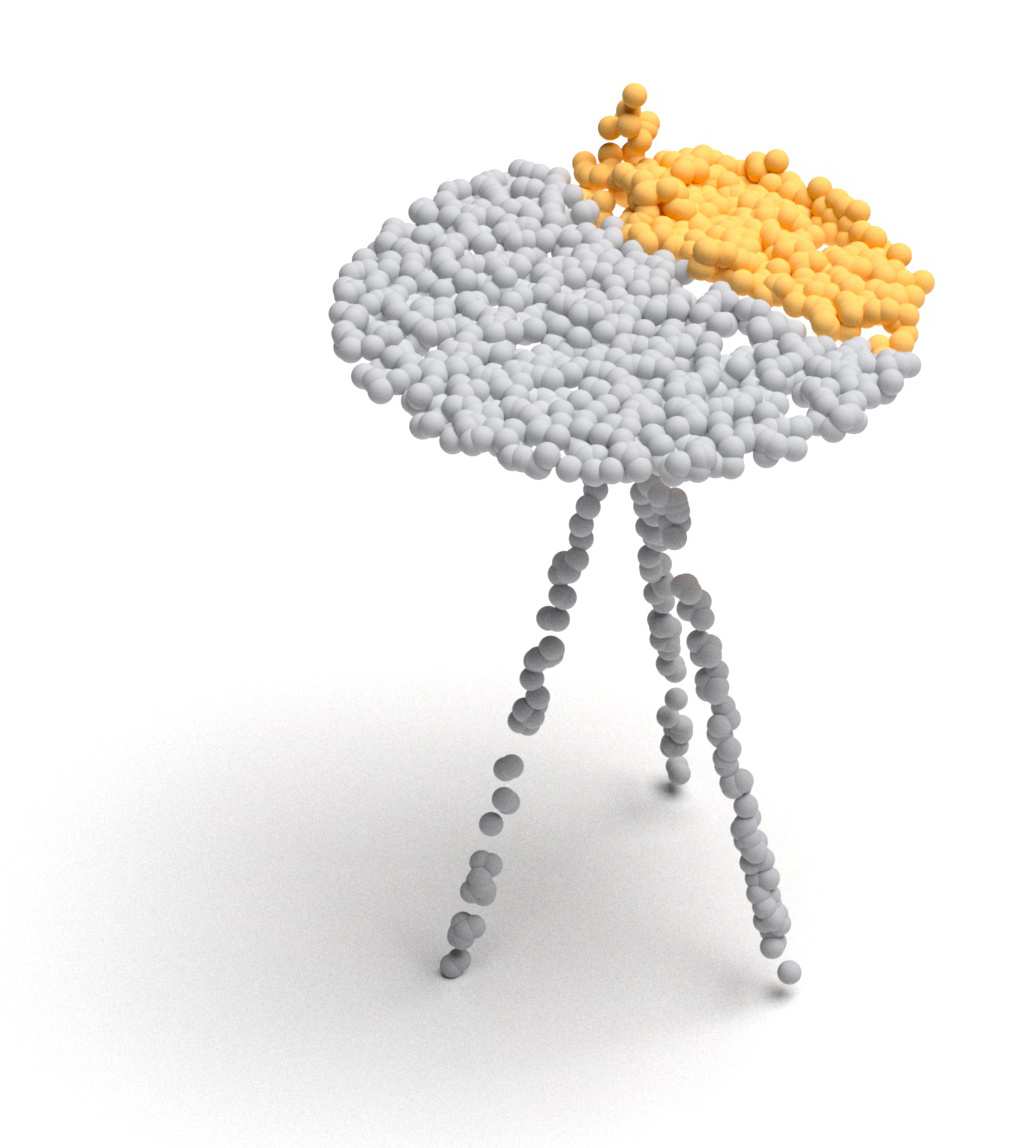} &
    \includegraphics[width=0.1\textwidth]{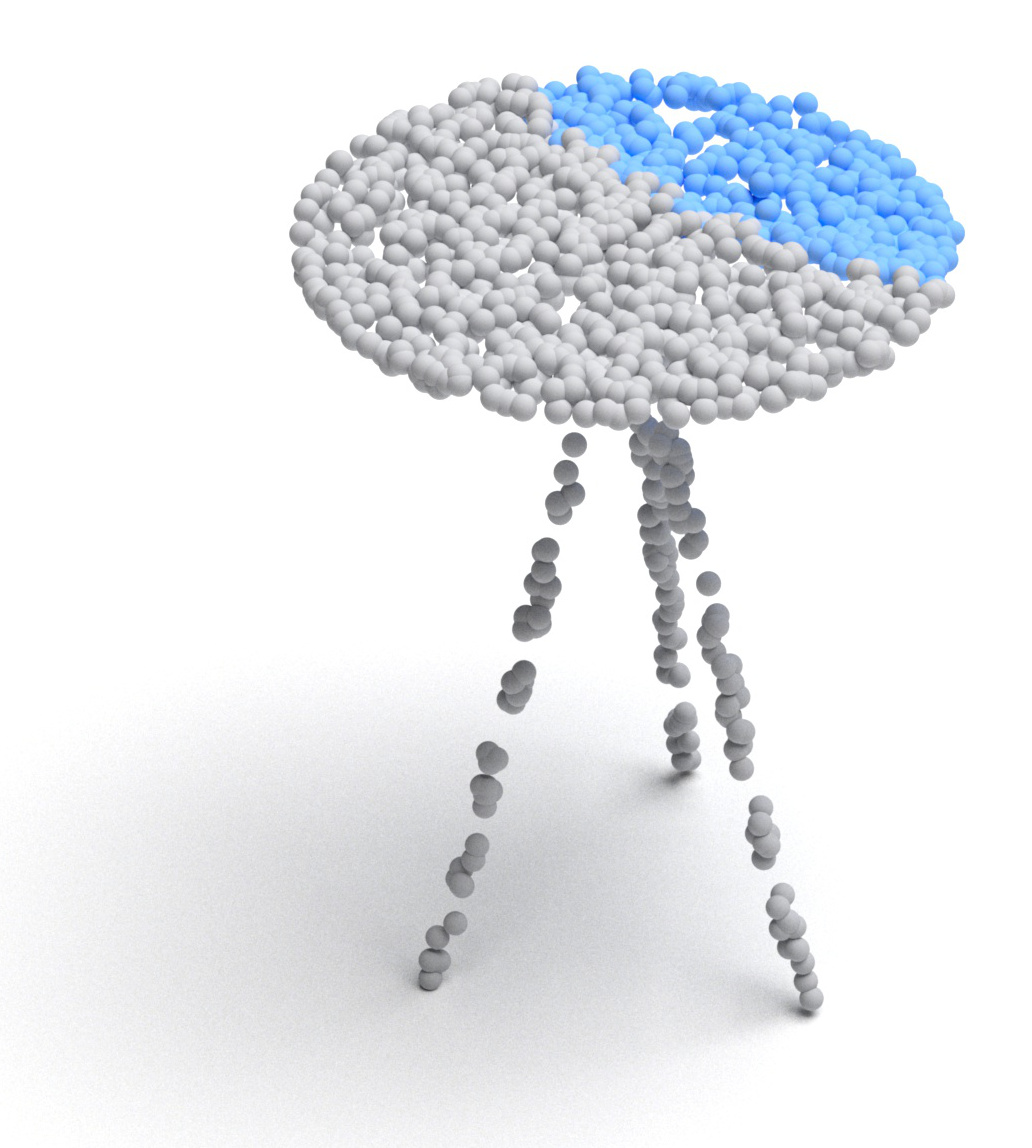} &
    \includegraphics[width=0.1\textwidth]{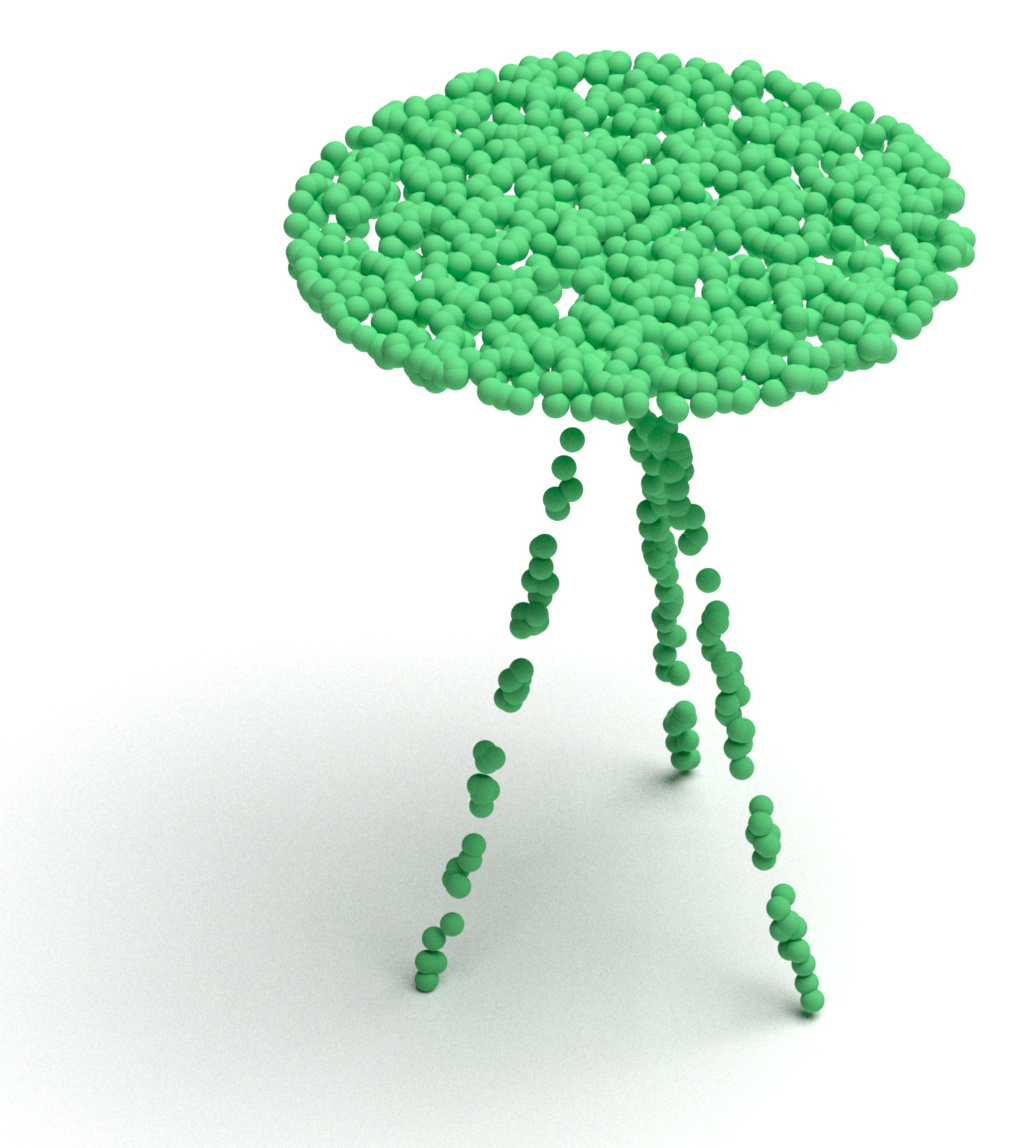}\\
    \includegraphics[width=0.1\textwidth]{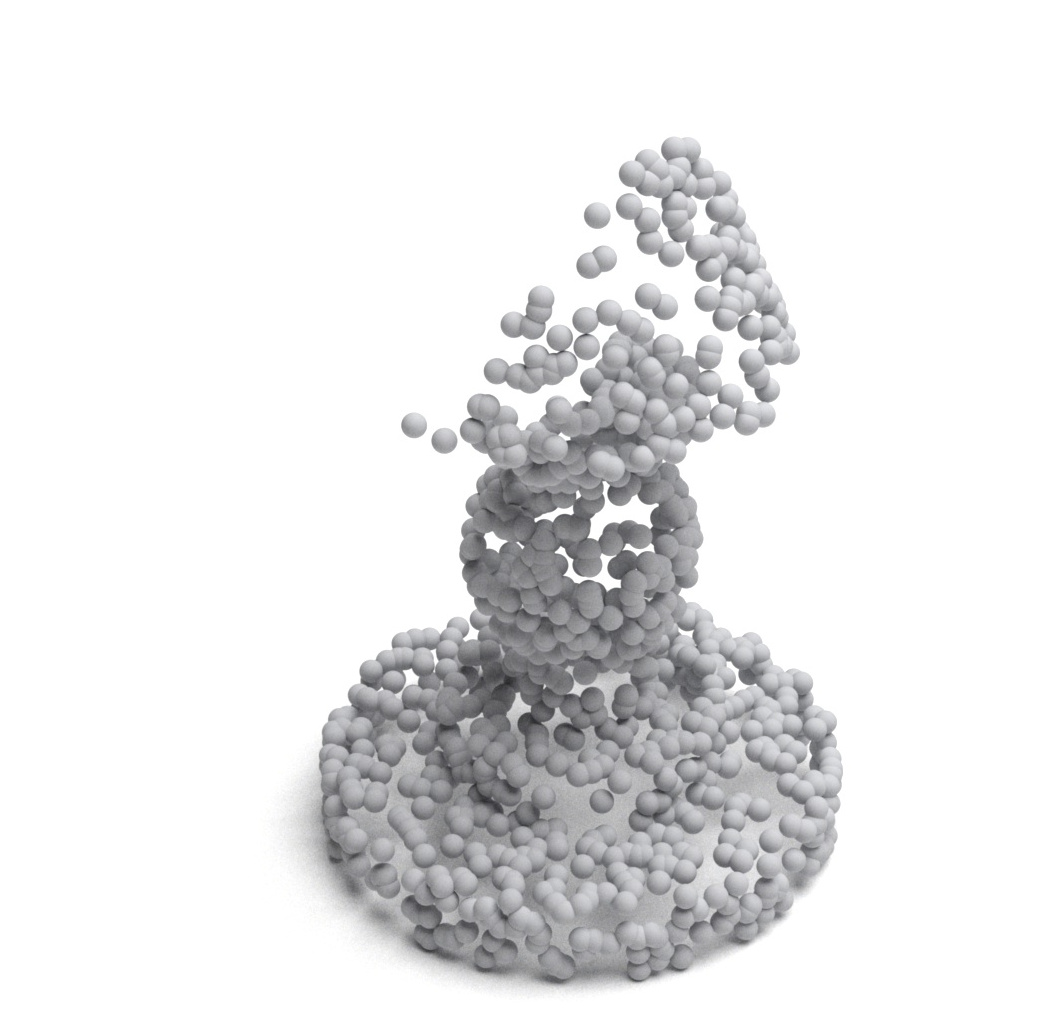} & 
    \includegraphics[width=0.1\textwidth]{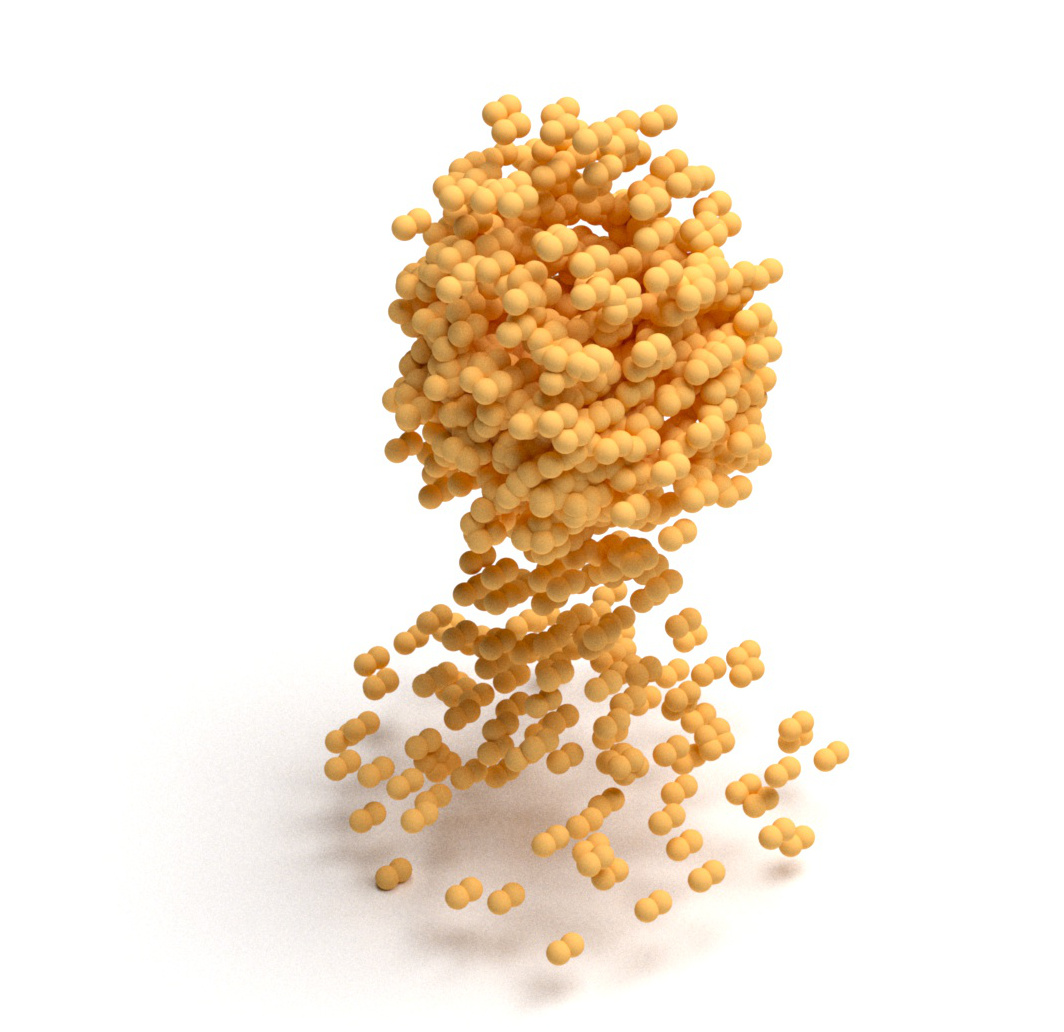}
    &\includegraphics[width=0.1\textwidth]{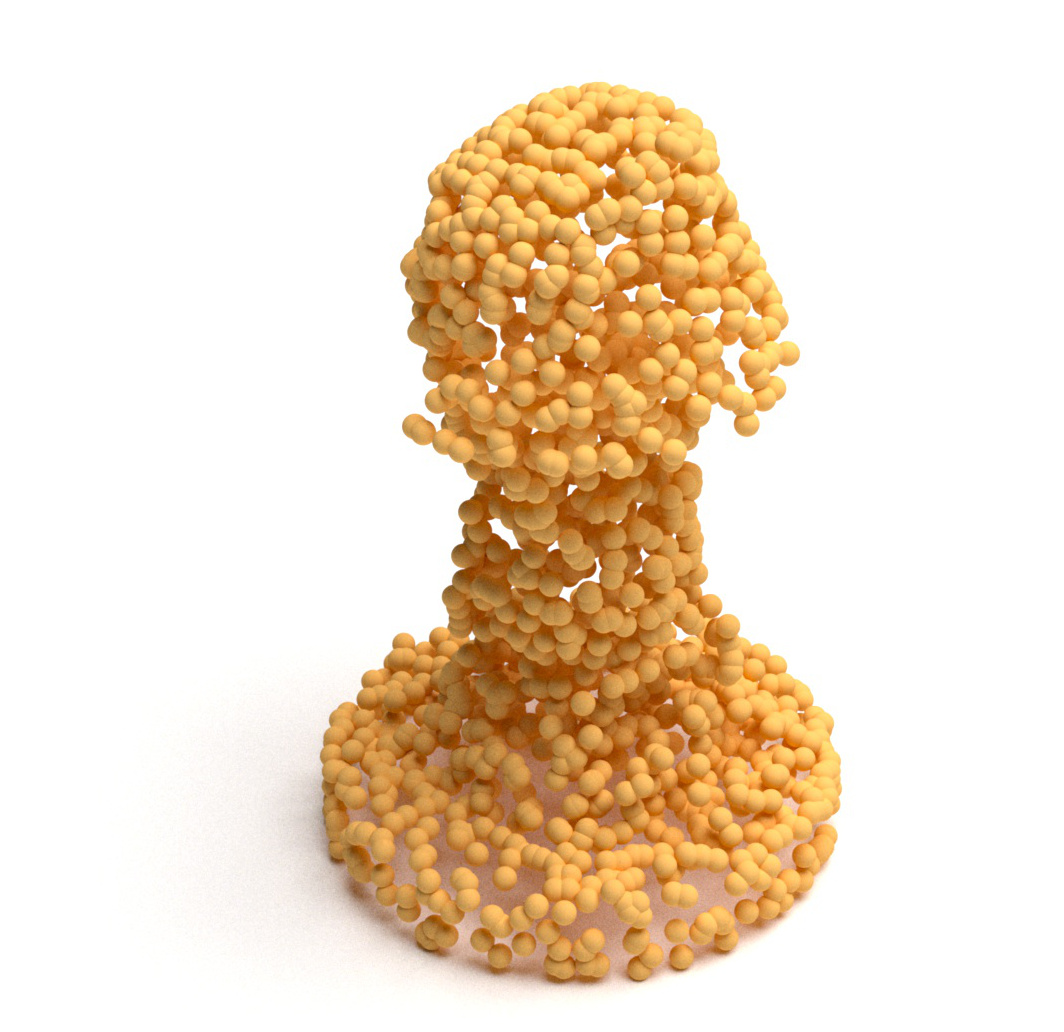}
    &\includegraphics[width=0.1\textwidth]{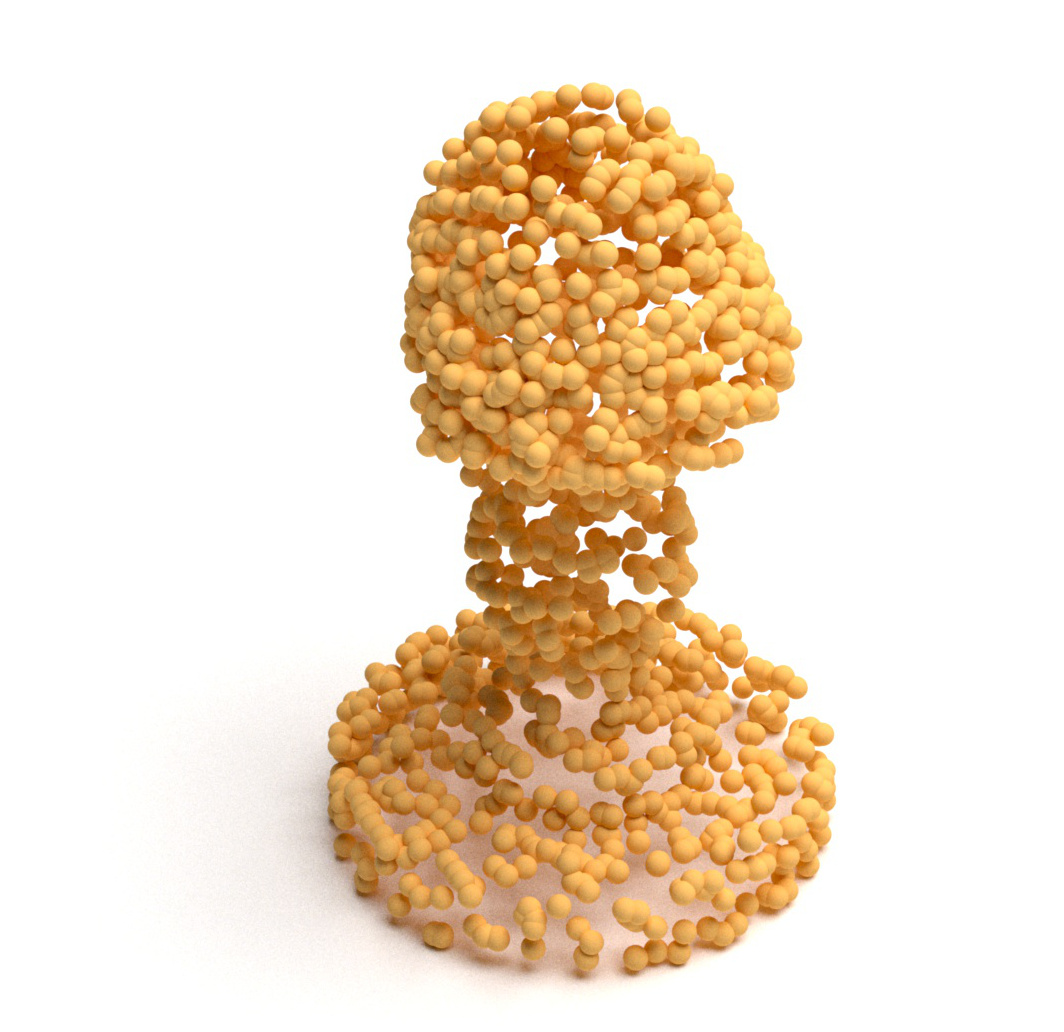} 
    &\includegraphics[width=0.1\textwidth]{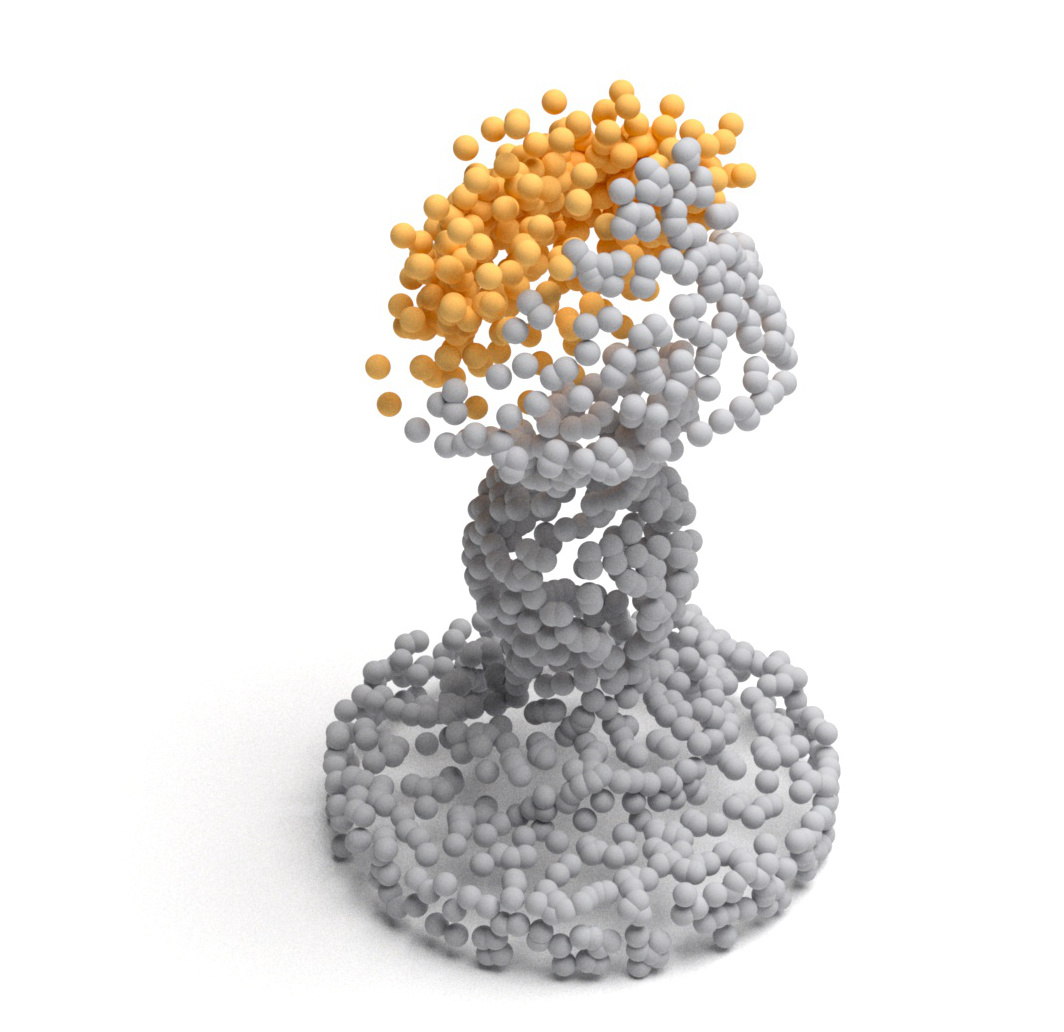} &
    \includegraphics[width=0.1\textwidth]{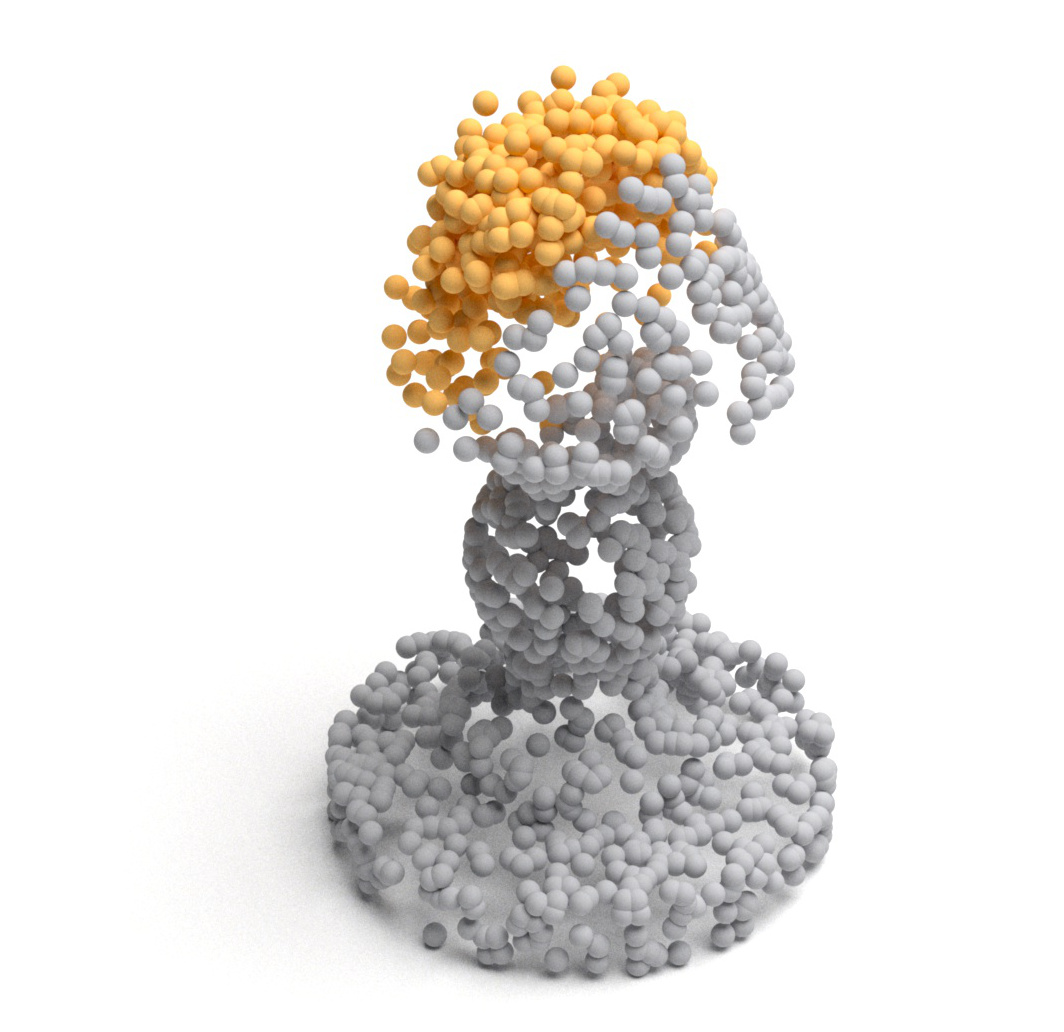} &
    \includegraphics[width=0.1\textwidth]{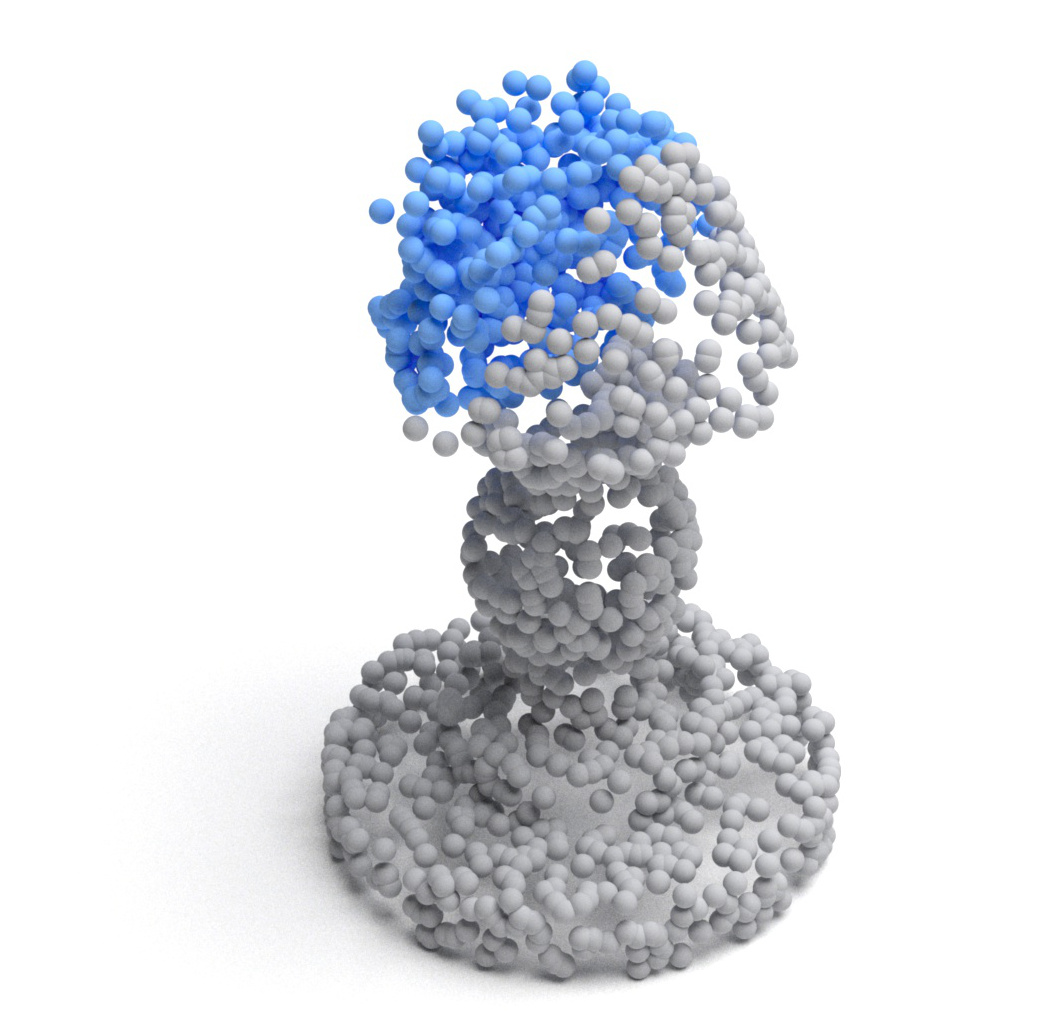}&
    \includegraphics[width=0.1\textwidth]{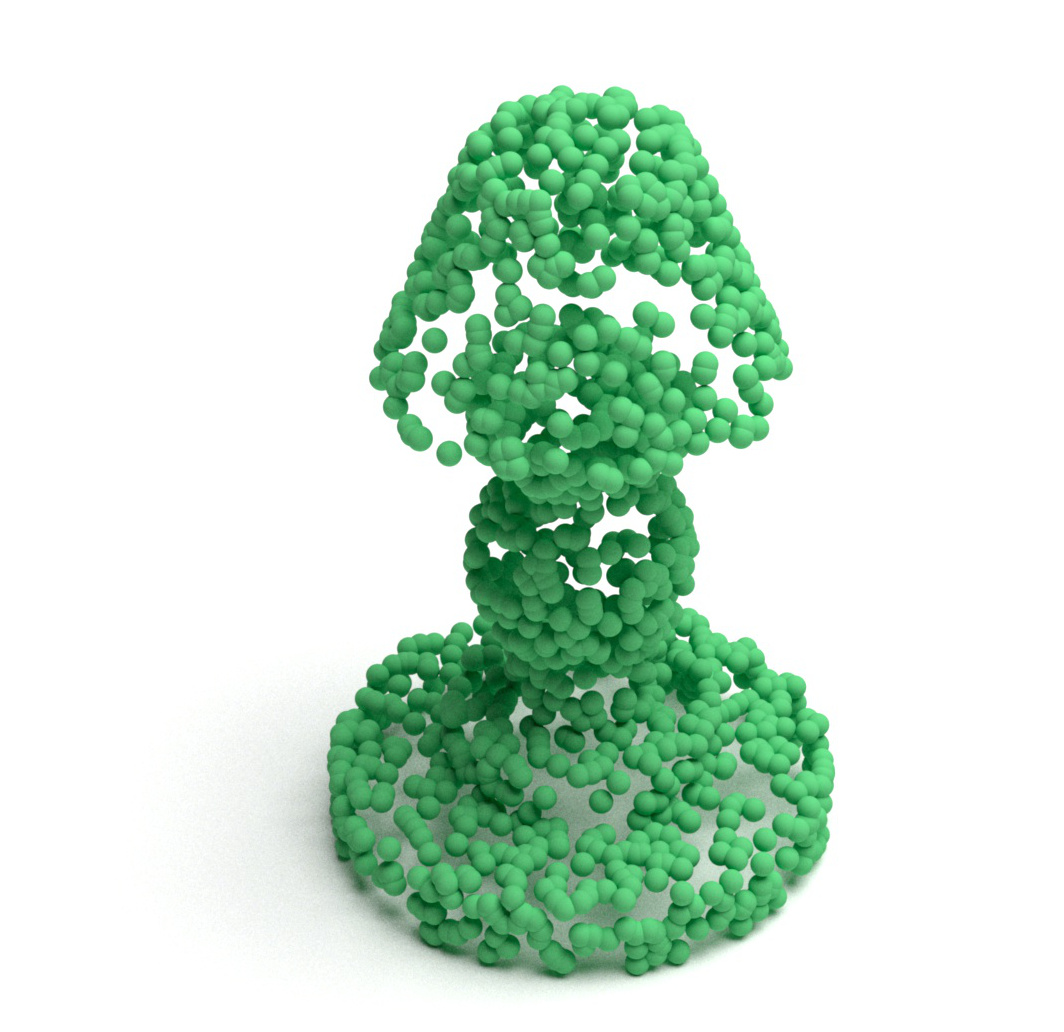}\\ 
    \end{tabular}
\caption{\emph{Known Categories - Qualitative}. The first four rows (airplane, chair, guitar, table) show how \our generates the missing shape part with more details and a less noisy appearance than its competitors. The last row (lamp) shows a general failure case for all the approaches. 
For PFNet and \our we visualize the predicted missing part (resp. yellow and blue points) w.r.t. the partial input (grey).
}
\vspace{-2mm}
\label{table:single-hole-qualitative}
\end{figure*}

\begin{table}[t!]
    \centering
    \begin{adjustbox}{width=0.36\textwidth}
    \begin{tabular}{@{~~}l@{~~}|c@{~~}c@{~~}|c@{~~}}
\hline
\multirow{2}{*}{\textbf{Method}} & \multicolumn{2}{c|}{\textbf{Single Hole}} & \multirow{2}{*}{\textbf{Two Holes}} \\
& {25\%} & {50\%} & \\
 \hline
 \textbf{PF-Net vanilla} & 20.209 &  20.950 & 25.140\\
 \textbf{PF-Net} & 20.445 & 19.325 & 33.632 \\
 \textbf{\our} & \textbf{16.517} & \textbf{17.554} & \textbf{24.430}\\
 \hline
\end{tabular}
\end{adjustbox}
\caption{\emph{Known Categories - Robustness Test} Overall average Chamfer Distance scaled by $10^4$. The results confirm the advantage of \our against its best competitor PF-Net.}
\label{table:robustness}
\vspace{-2mm}
\end{table}

\subsection{Known Categories}
\noindent\textbf{Quantitative Analysis} Table \ref{table:single-hole-quantitative} presents the  completion results that indicate the superiority of \our with respect to all the considered baseline approaches. More in details, \our outperforms its best competitor PF-Net on eight out of thirteen categories, with a large margin on lamp, cap, and bag.  

\noindent\textbf{Qualitative Analysis} Figure \ref{table:single-hole-qualitative} shows the point cloud reconstructed by the different baseline methods and by \our. On the airplane point cloud, most of the baselines lack one or both the wing engines. On the chair point cloud, \our is the only method to reconstruct the missing leg without any significant noise. The guitar highlights the clear advantage of generating only the missing part, rather than reconstructing the whole shape, while also showing how \our is much more precise than the two PF-Net variants. For the table, all the baselines present artifacts either on the horizontal surface or on the legs. Finally, the last row shows a failure case: none of the methods is able to generate a precise reconstruction for the missing part of the lamp. In general, it is evident that the results of PCN are too noisy while those of MSN are often discontinuous or incomplete. 

\noindent\textbf{Robustness Test}
To analyze the robustness of \our we ran two sets of experiments over all the known classes by considering a single larger hole or two separate smaller holes in the point clouds. In the first case, we changed $M$ passing from $512$ to $1024$, thus extending the missing part from $25\%$ to $50\%$ of the original shape. In the second case, we randomly chose two viewpoints, each used to define the origin of a $12.5\%$ ($M_1=M_2=256$ points) hole. We focus on the comparison with the best performing baseline, \ie PF-Net, and the results in Table \ref{table:robustness} confirm that \our outperforms it in all the settings, thus showing a stronger robustness. It is interesting to note that, when dealing with two holes, the adversarial discriminator of PF-Net is detrimental: by monitoring the training, our intuition is that the PF-Net reconstructed output remains too different from the ground truth to properly trigger the beneficial effect of the adversarial game. 

\begin{table}[t!]
    \centering
    \resizebox{0.49\textwidth}{!}{
    \begin{tabular}{@{~~}c@{~~}|@{~~}c@{~~~}c@{~~~}c@{~~~}c@{~~~}c@{~~~}}
\hline
\multirow{2}{*}{\textbf{Categories}} & \textbf{MSN}  &  \textbf{CRN}  & \textbf{PF-Net}  & \textbf{PF-Net} & \multirow{2}{*}{\textbf{\our}}\\
& \cite{Liu_2020_AAAI_morphing} &  \cite{Wang_2020_CVPR_cascaded} & \textbf{vanilla} \cite{Huang_2020_CVPR_pfnet} & \cite{Huang_2020_CVPR_pfnet} & \\
\hline
\multicolumn{6}{c}{\textbf{Similar}}\\
\hline
Bicycle & 47.423 & 64.275 & 49.779 & 47.186 & 39.684 \\
Basket & 48.100 & 50.692 & 58.866 & 57.066 & \textbf{34.613} \\
Helmet & 71.161 & 57.851 & 63.742 & 69.849 & \textbf{47.412} \\
Bowl & 52.002 & 63.357 & 97.316 & 78.793 & \textbf{35.209}\\
Rifle & 34.712 & 47.239 & 25.438 & 28.684 & \textbf{12.004} \\
Vessel & 30.948 & 41.418 & 27.122 & 31.114 & \textbf{18.836} \\
\hline
Overall & 35.544 & 46.166 & 31.232 & 33.844 & \textbf{17.680} \\
\hline
\multicolumn{6}{c}{\textbf{Dissimilar}}\\
\hline
Piano & 62.969 & 61.643 & 62.131 & 62.994 & \textbf{49.429} \\
Bookshelf & 48.397 & 44.738 & 58.920 & 55.123 & \textbf{34.681} \\
Bottle & 29.580 & 20.134 & 25.543 & 24.578 & \textbf{20.002} \\
Clock & 57.222 & 38.132 & 50.964 & 48.373 & \textbf{32.826} \\
Microwave & 53.354 & 56.259 & 61.702 & 56.152 & \textbf{41.877} \\
Telephone & 38.032 & 25.554 & 38.085 & 32.063 & \textbf{20.106} \\
\hline
Overall & 45.049 & 34.625 & 45.014 & 41.449 & \textbf{28.403}\\
\hline
\end{tabular}
}
\caption{\emph{Novel Categories - Quantitative}. Chamfer distance on the missing region of point clouds scaled by $10^4$. The lower, the better. }
\label{table:novelcat-quantitative}
\vspace{-3mm}
\end{table}

\begin{figure*}[ht!]
\centering
    \begin{adjustbox}{width=0.90\textwidth}
    \begin{tabular}{@{~}c@{~~}c@{~~}c@{~~}c@{~~}c@{~~} c@{~~}c@{~~}c@{~}}
    Input & MSN & CRN &  PF-Net vanilla & PF-Net & \our & GT \\
    \includegraphics[width=0.14\textwidth]{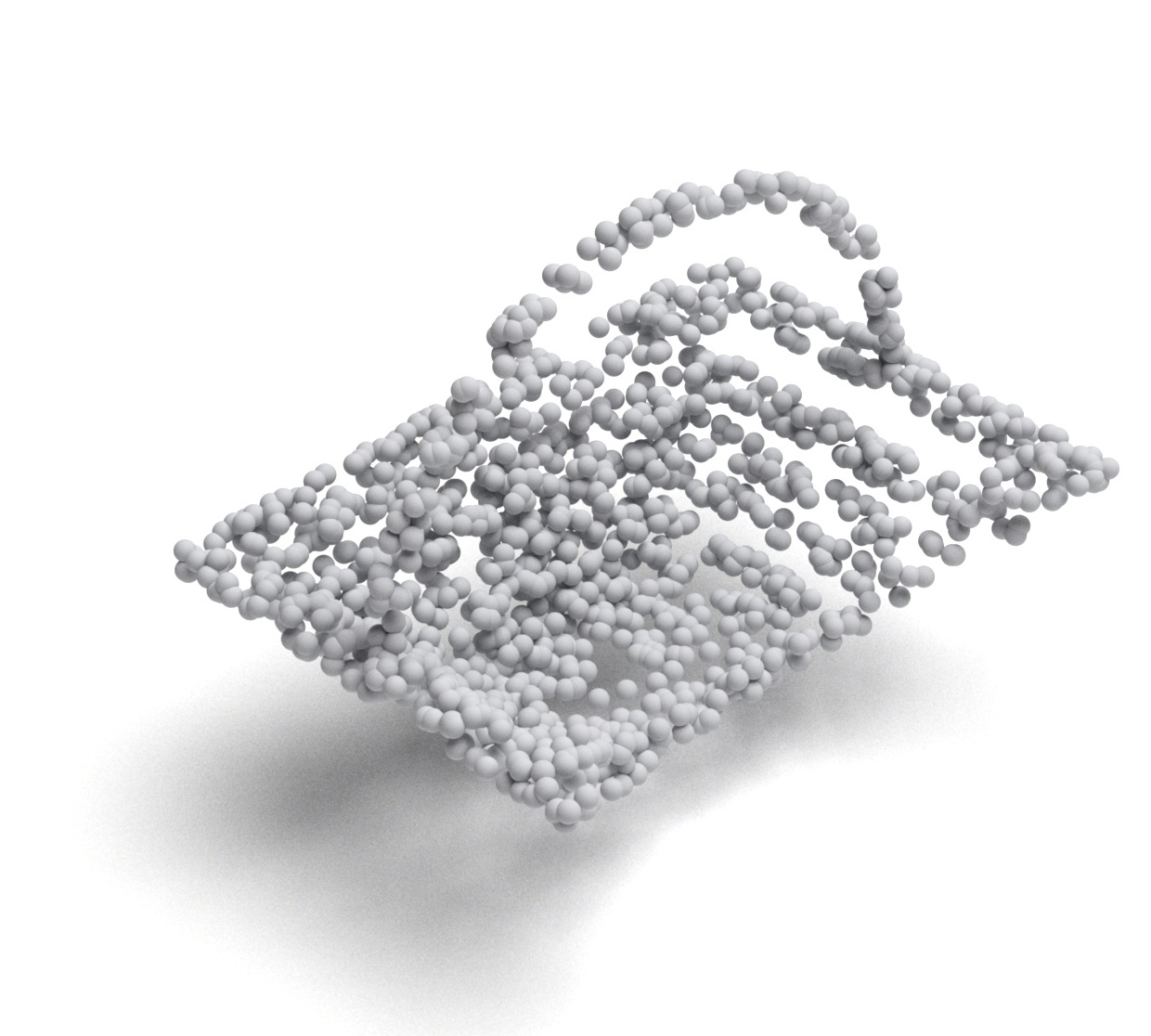}
    &\includegraphics[width=0.14\textwidth]{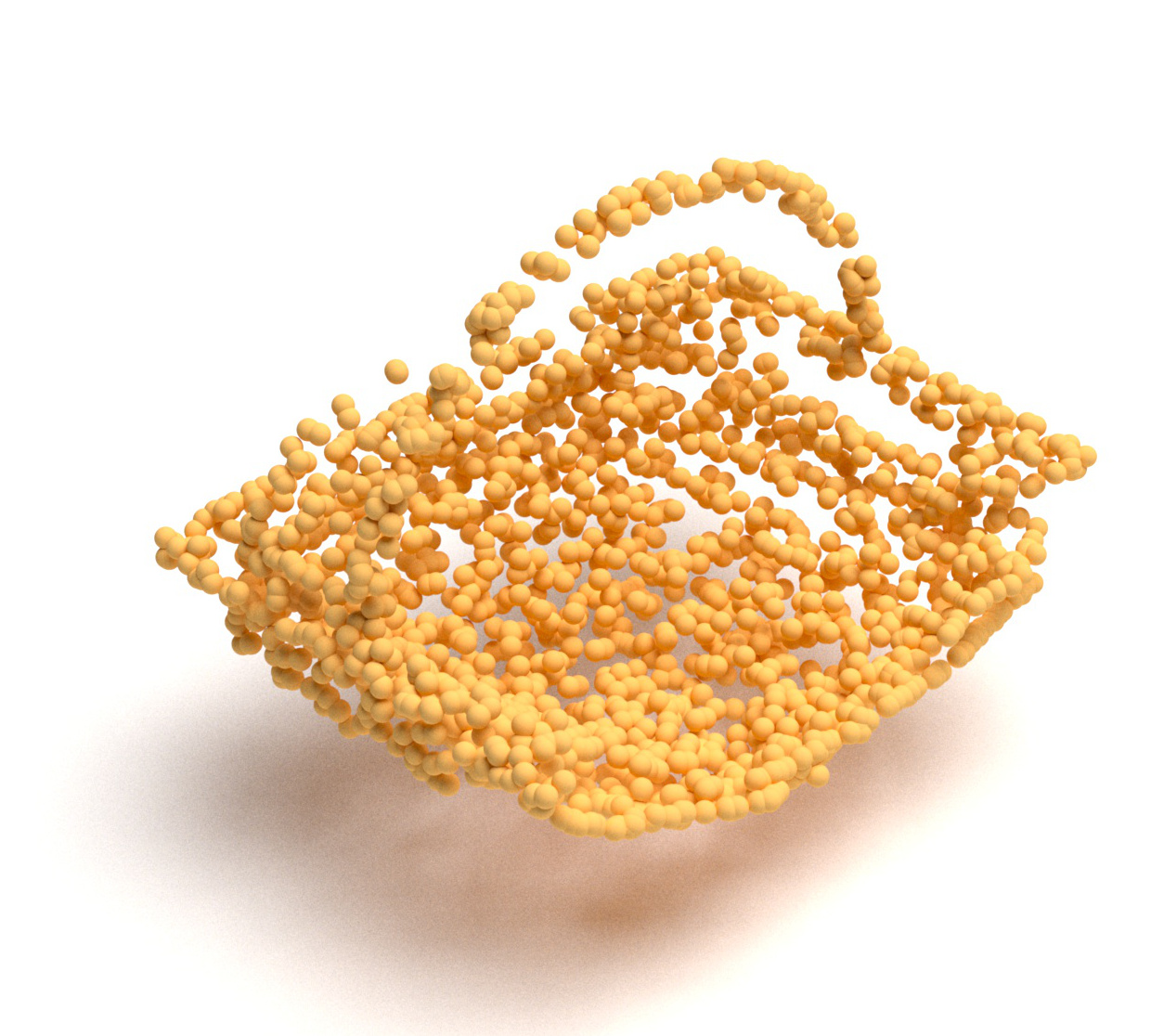}
    & \includegraphics[width=0.14\textwidth]{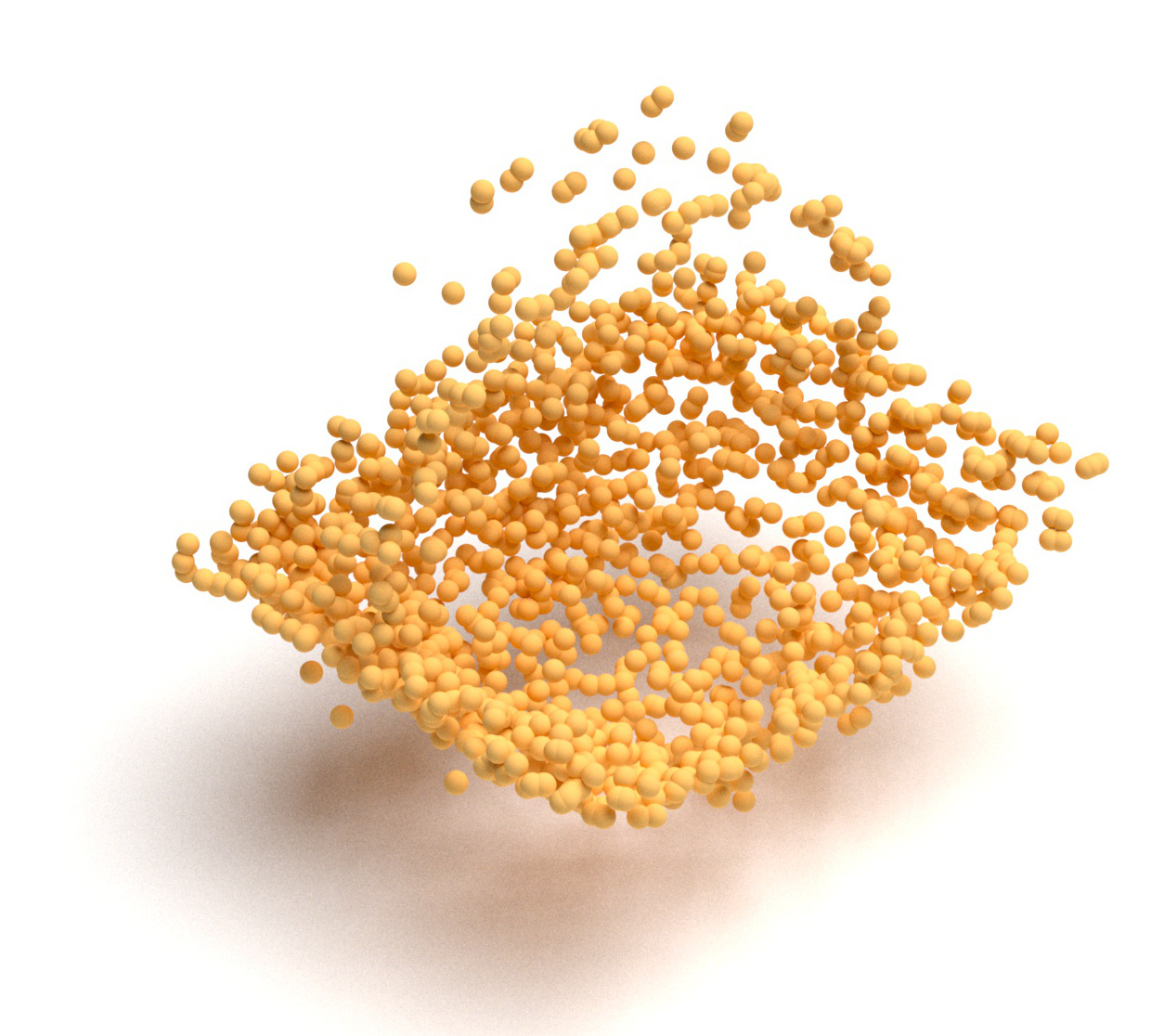}
    & \includegraphics[width=0.14\textwidth]{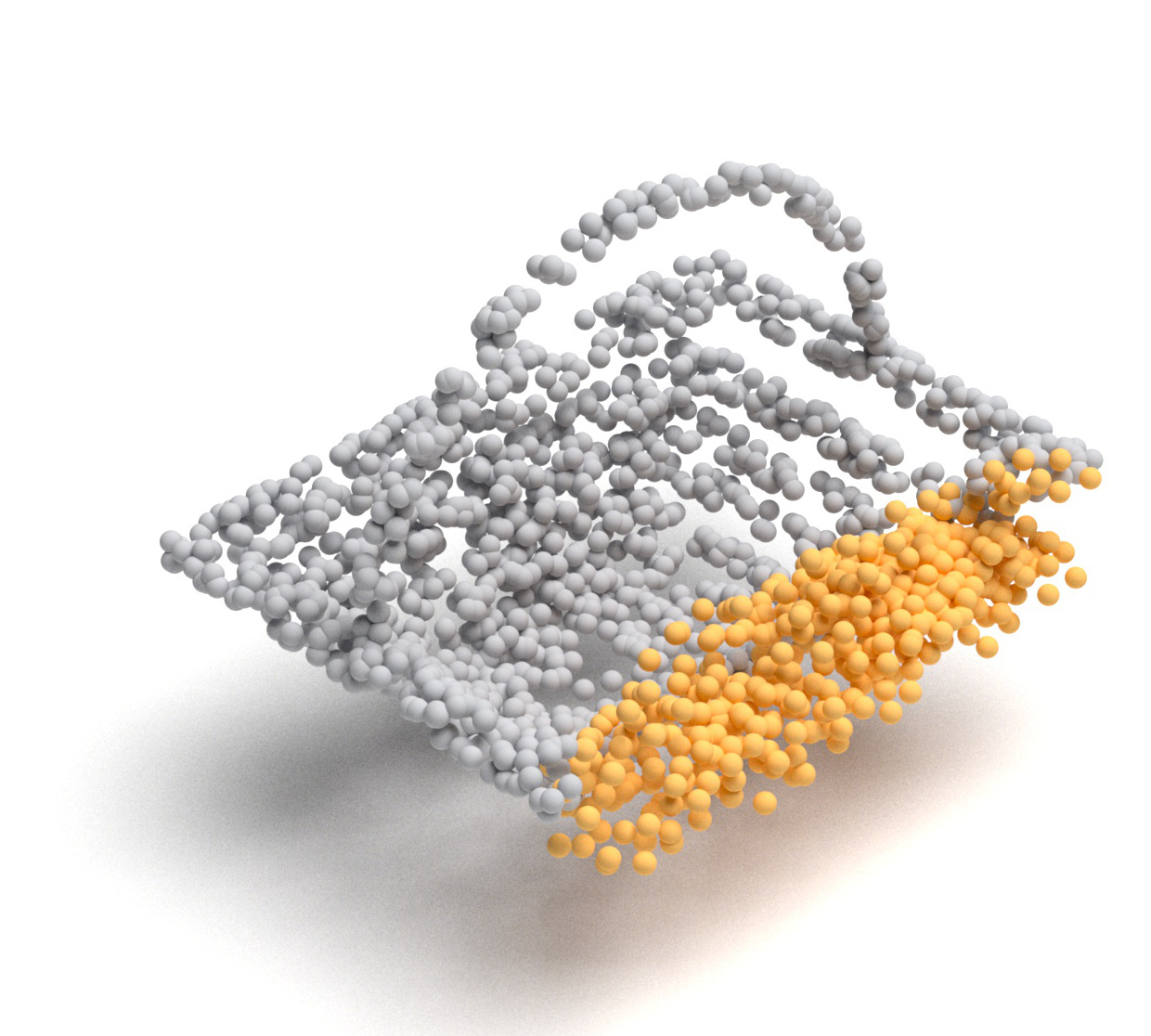}
    & \includegraphics[width=0.14\textwidth]{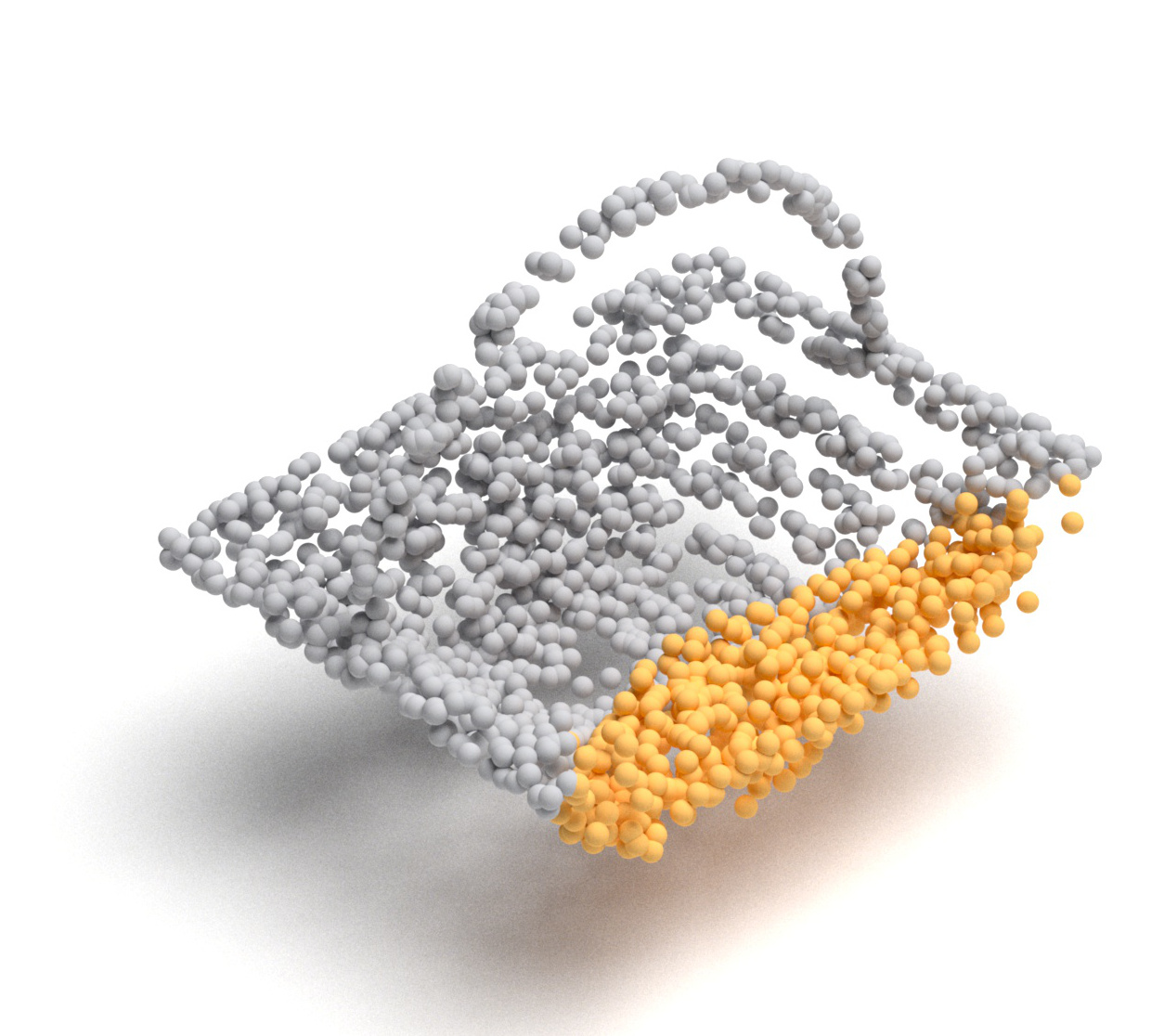}
    & \includegraphics[width=0.14\textwidth]{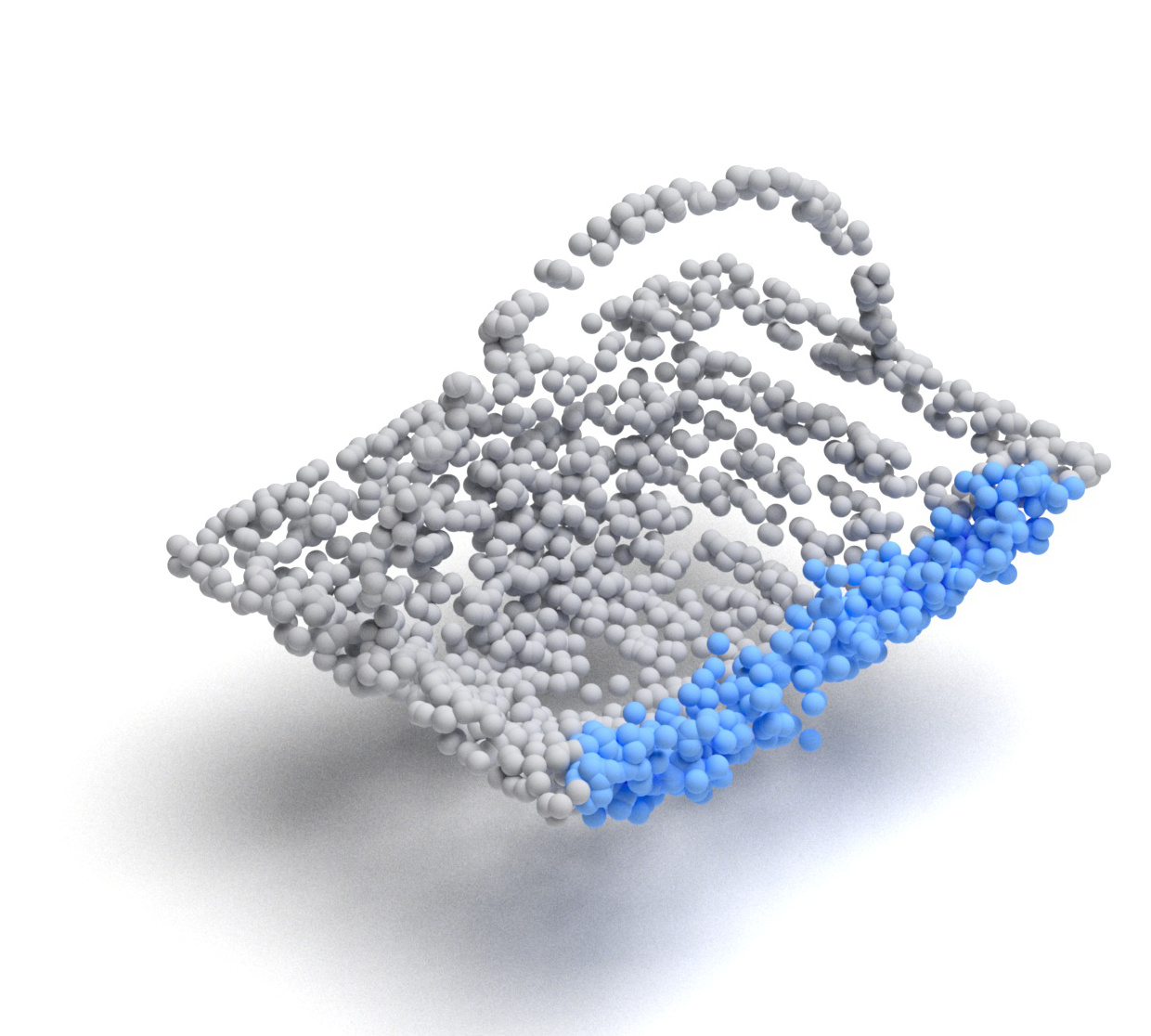}
    & \includegraphics[width=0.14\textwidth]{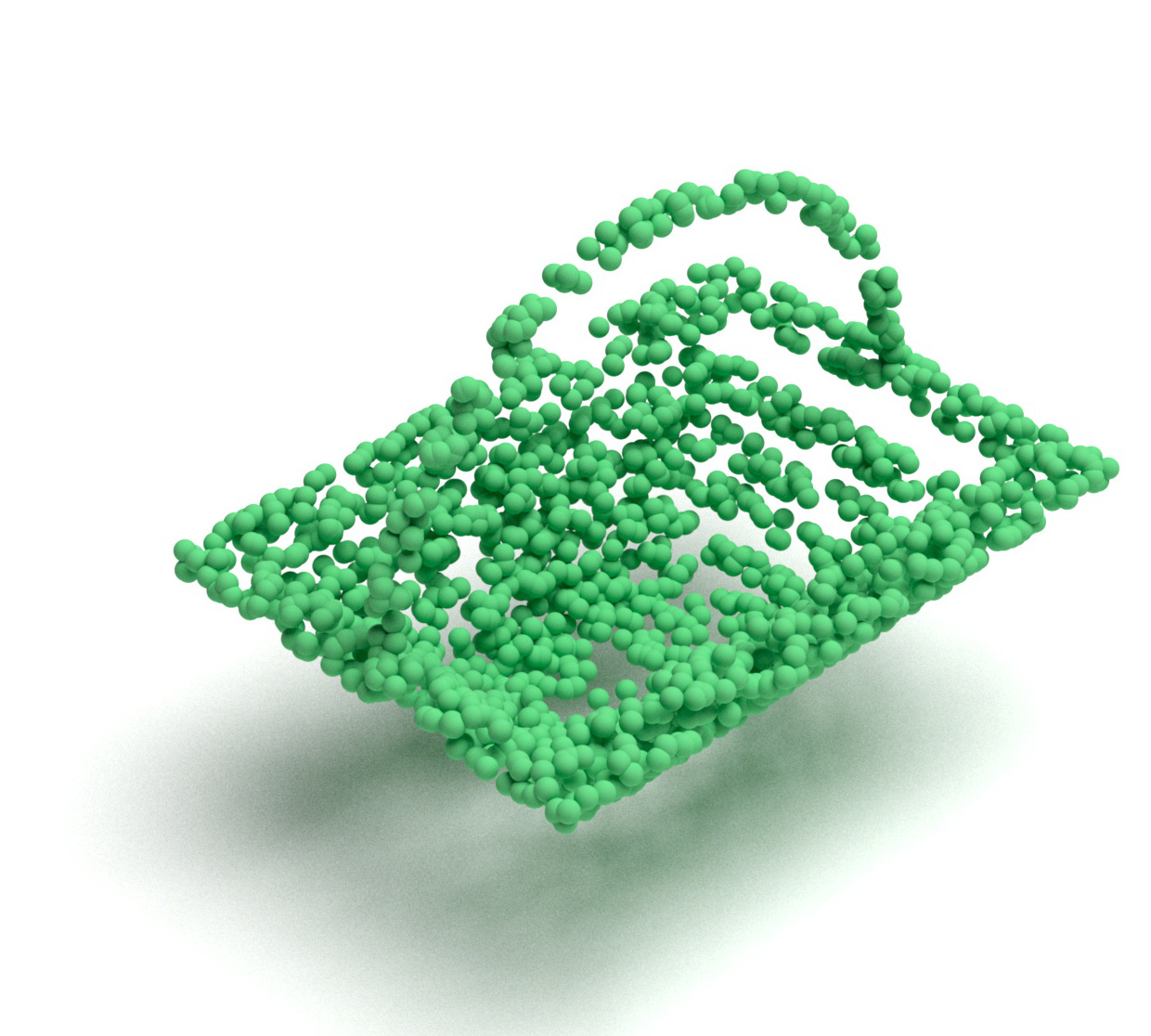}\\
    \includegraphics[width=0.14\textwidth]{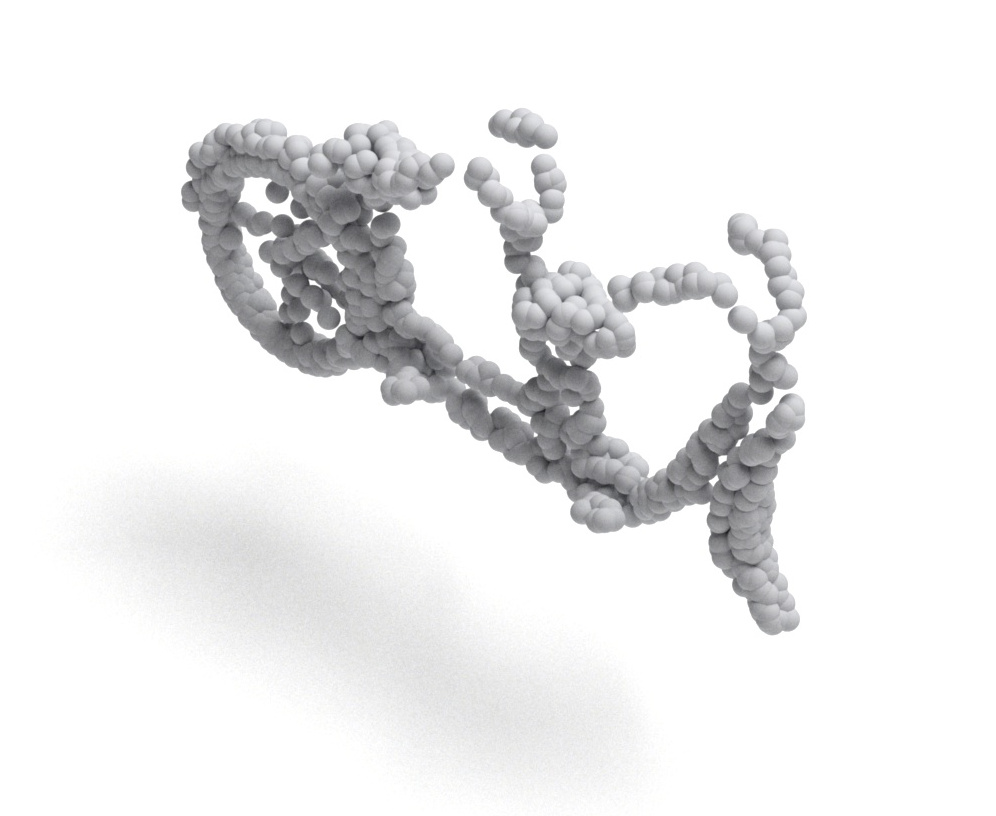}
    & \includegraphics[width=0.14\textwidth]{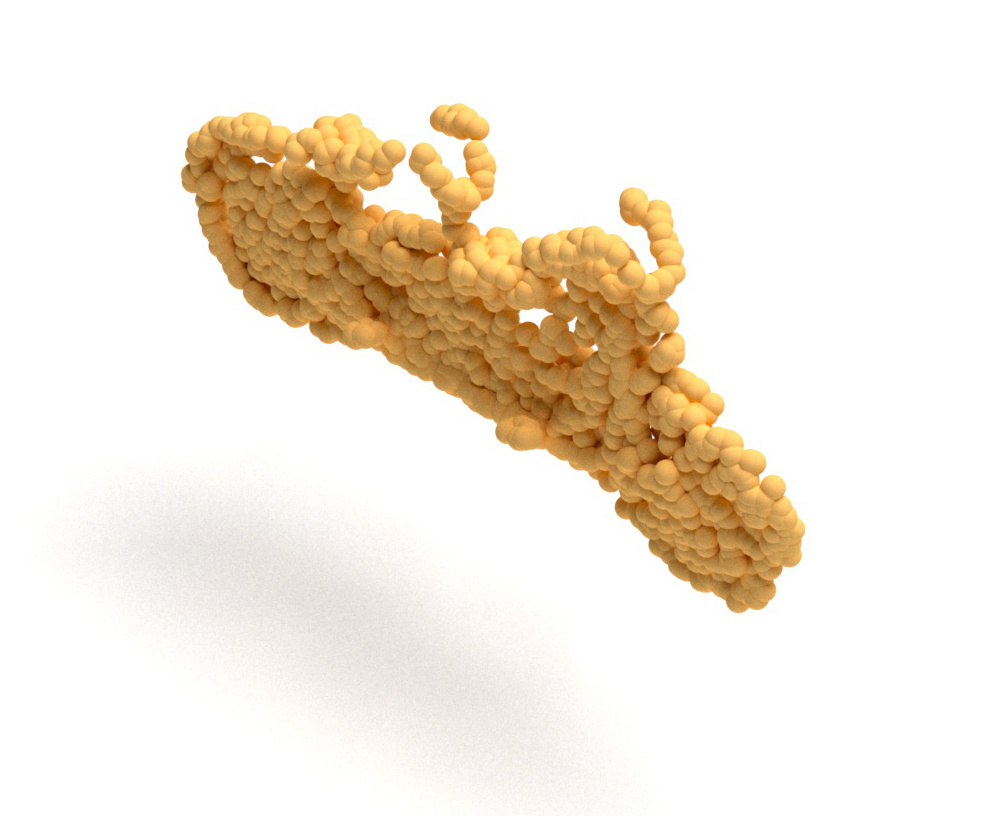}
    & \includegraphics[width=0.14\textwidth]{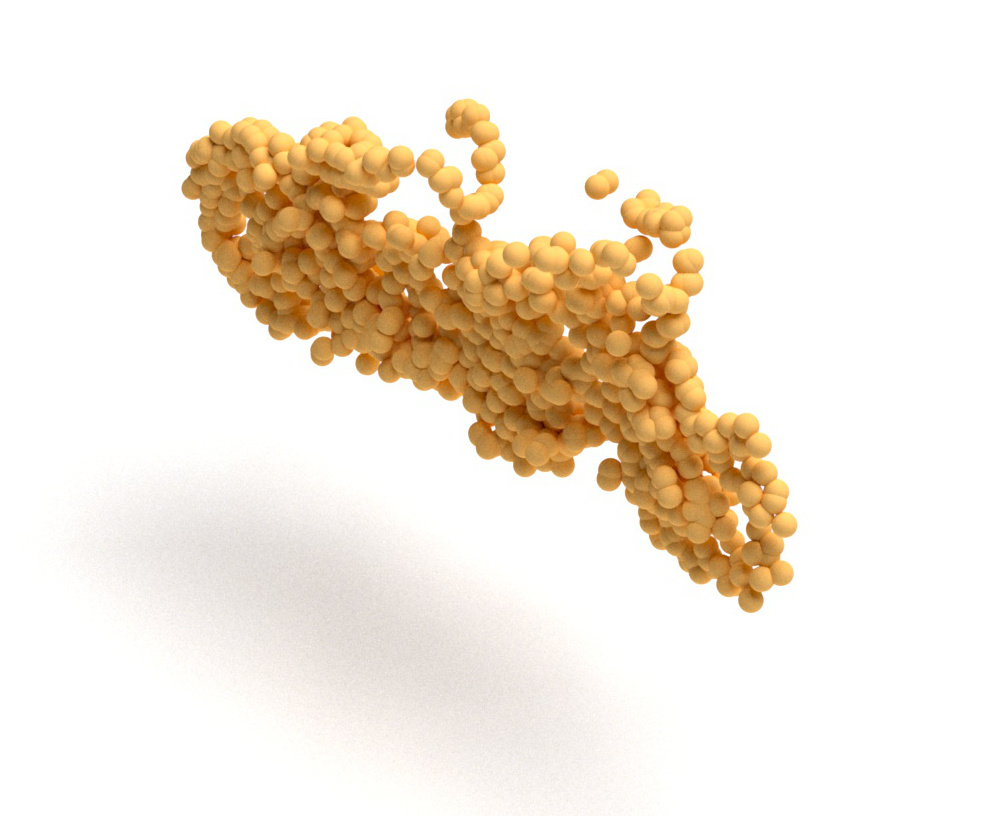}
    & \includegraphics[width=0.14\textwidth]{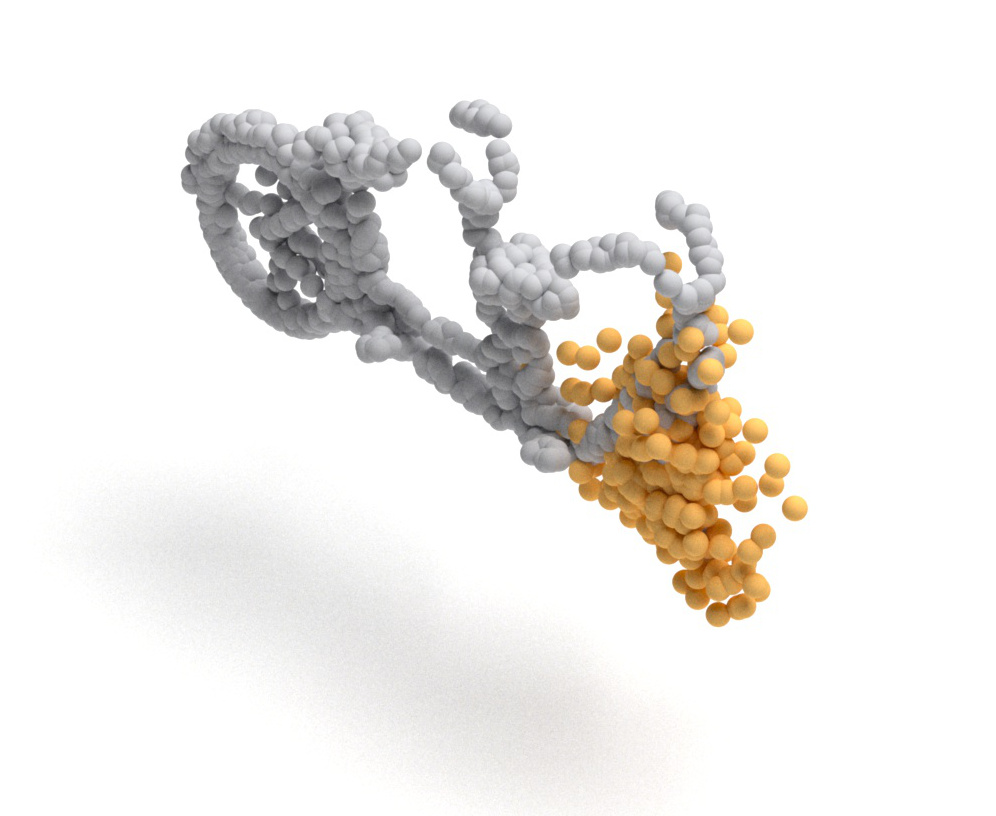}
    & \includegraphics[width=0.14\textwidth]{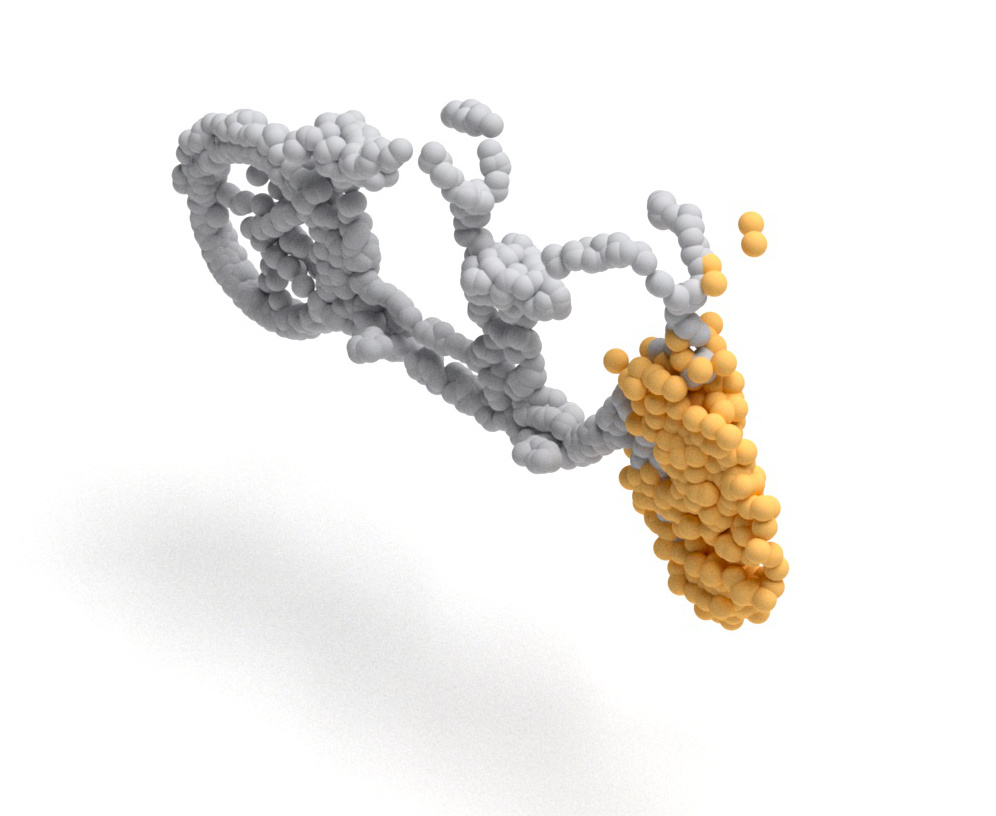}
    & \includegraphics[width=0.14\textwidth]{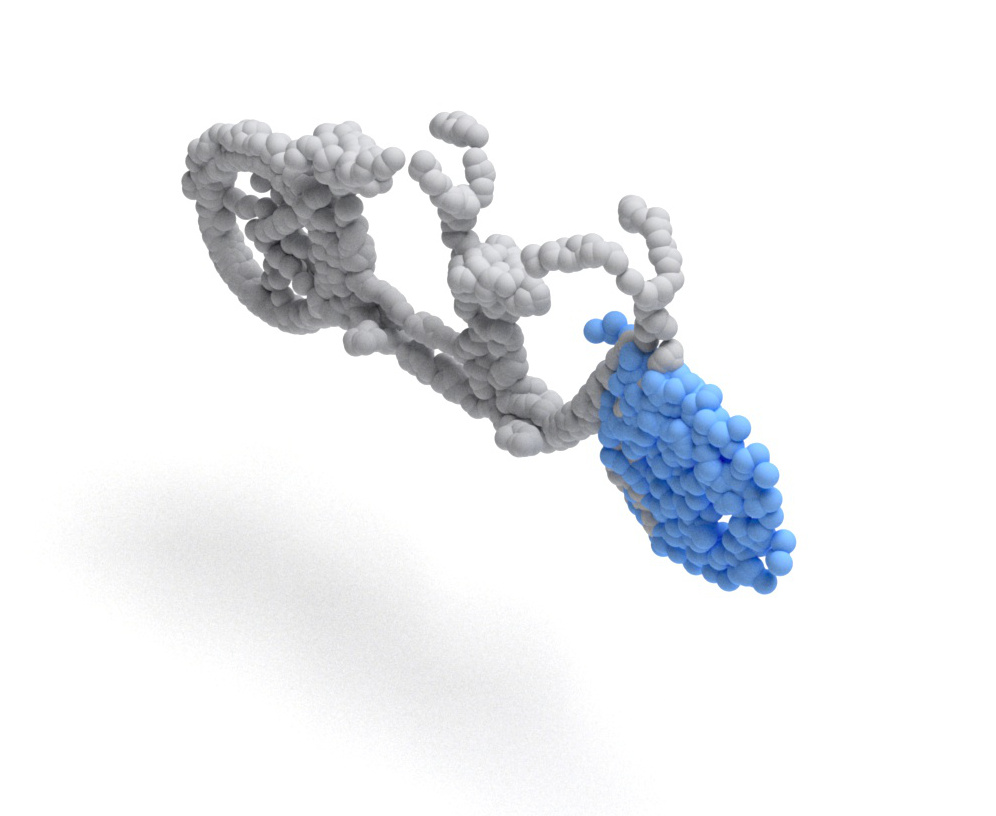}
    & \includegraphics[width=0.14\textwidth]{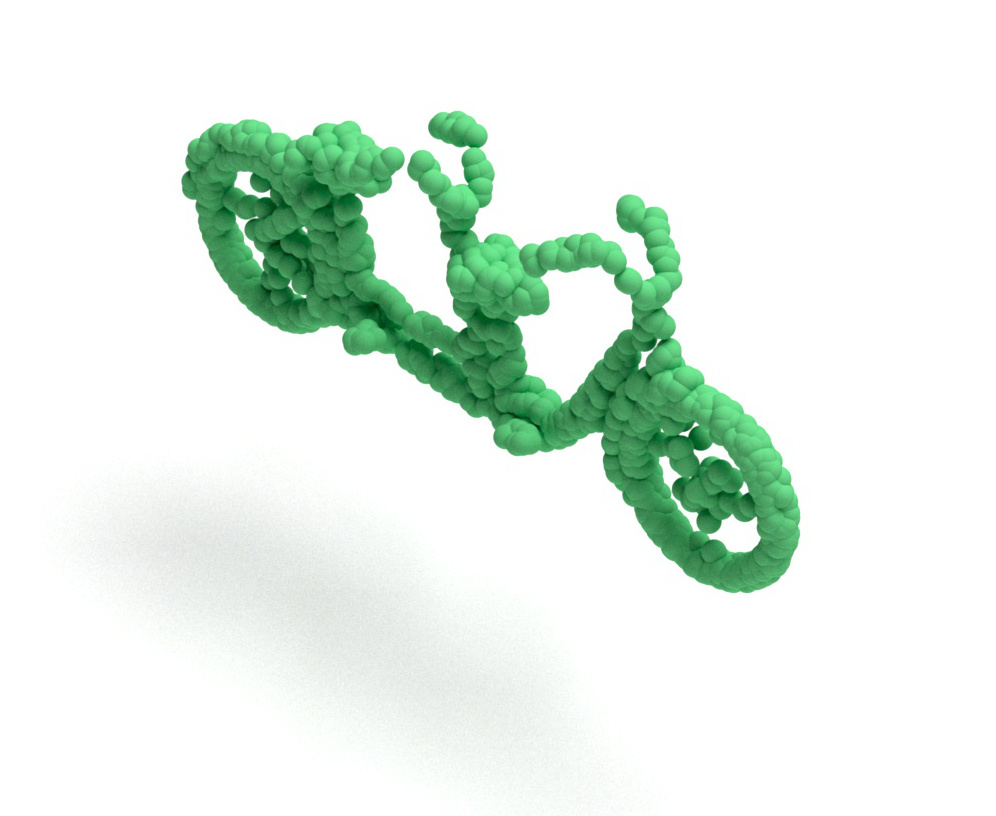}\\
    \includegraphics[width=0.12\textwidth]{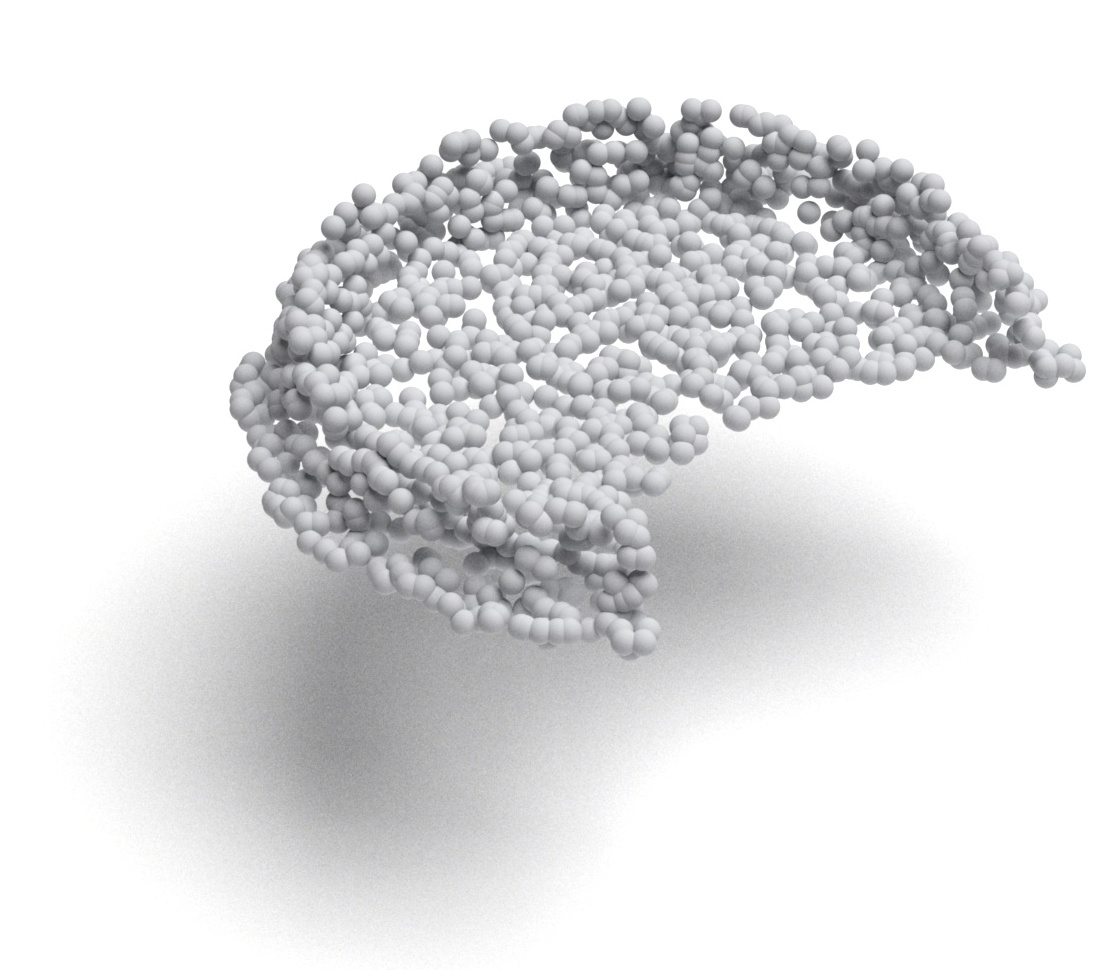} 
    & \includegraphics[width=0.12\textwidth]{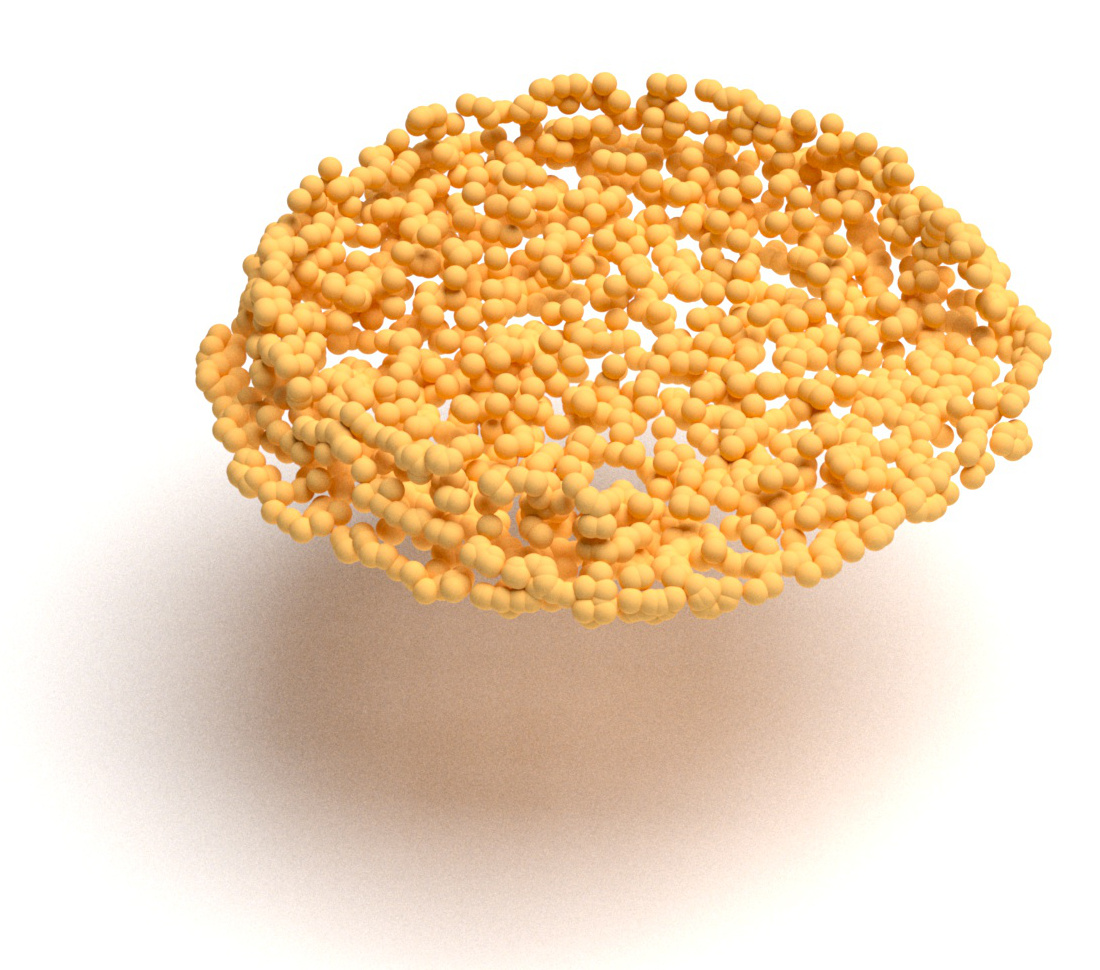}
    & \includegraphics[width=0.12\textwidth]{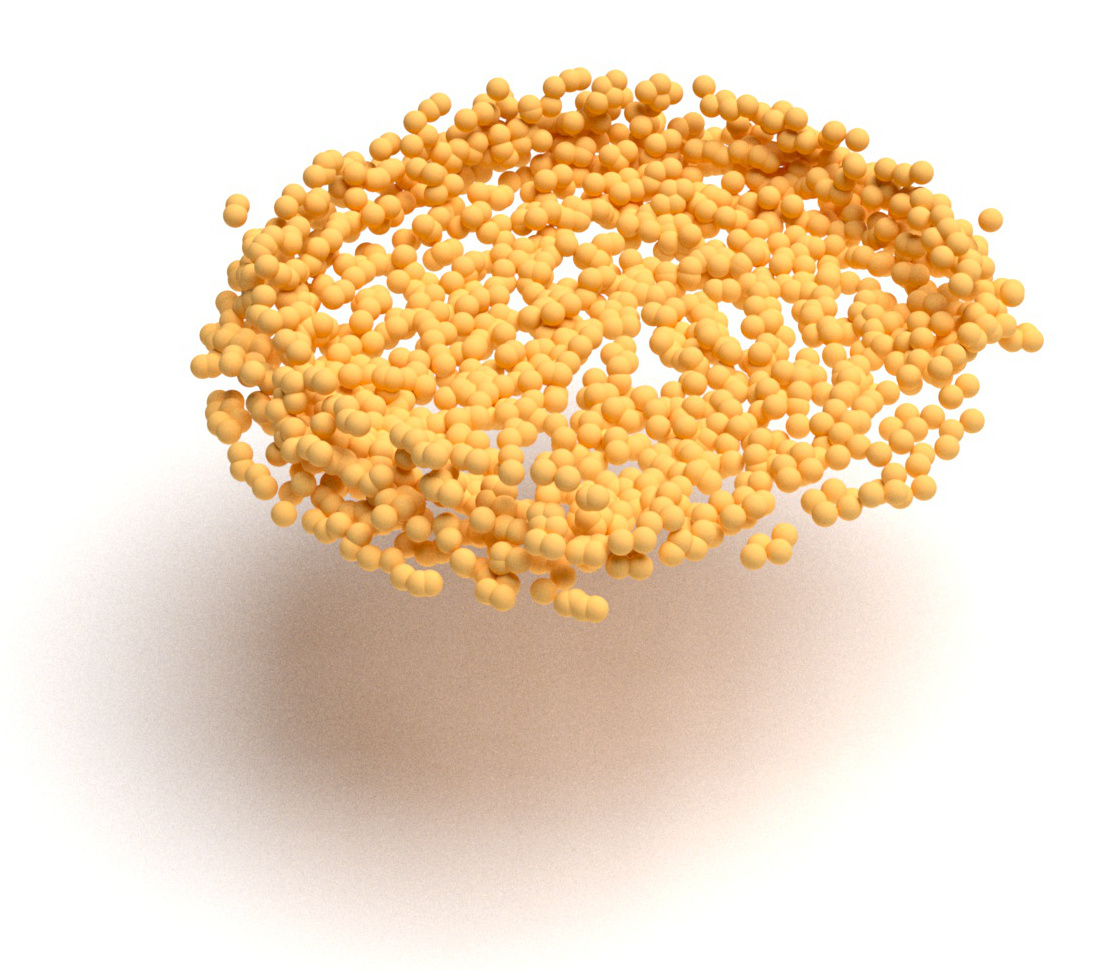}
    & \includegraphics[width=0.12\textwidth]{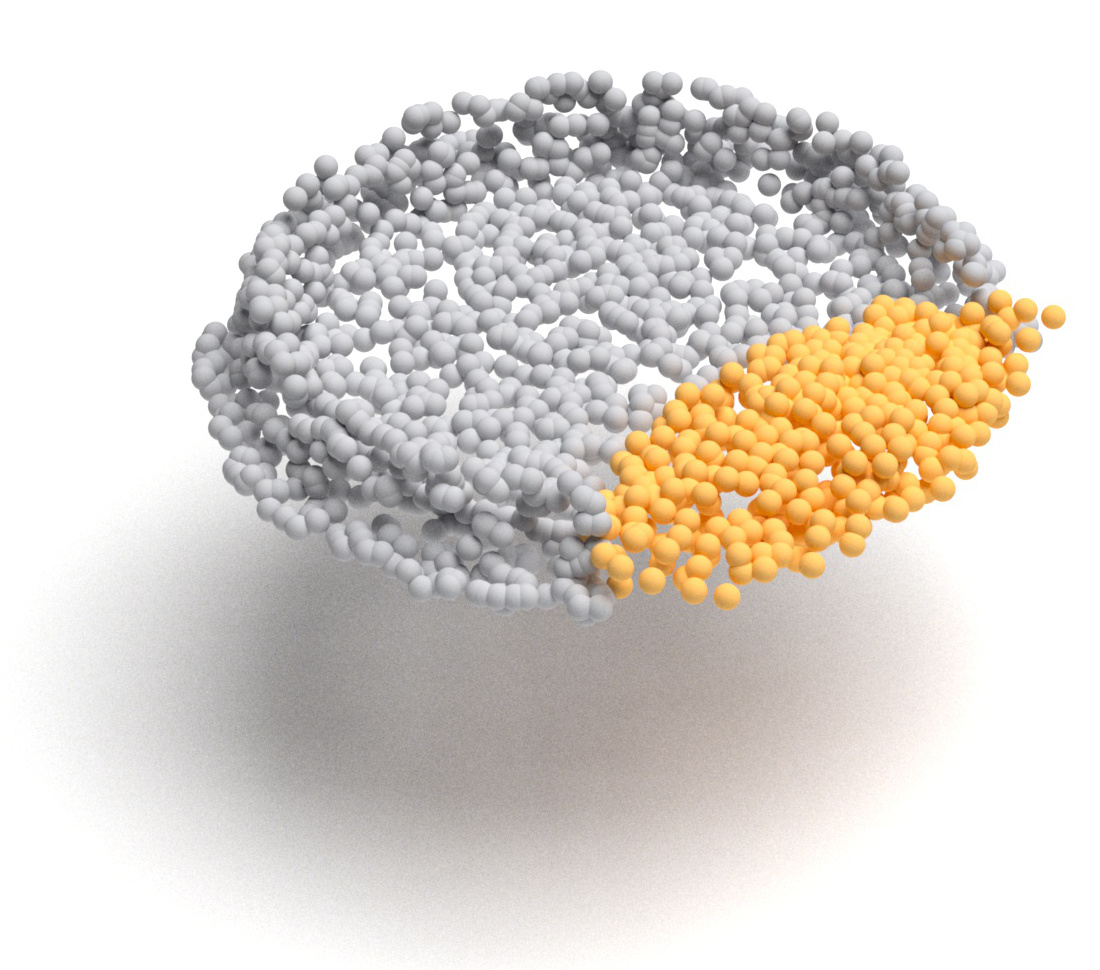}& \includegraphics[width=0.12\textwidth]{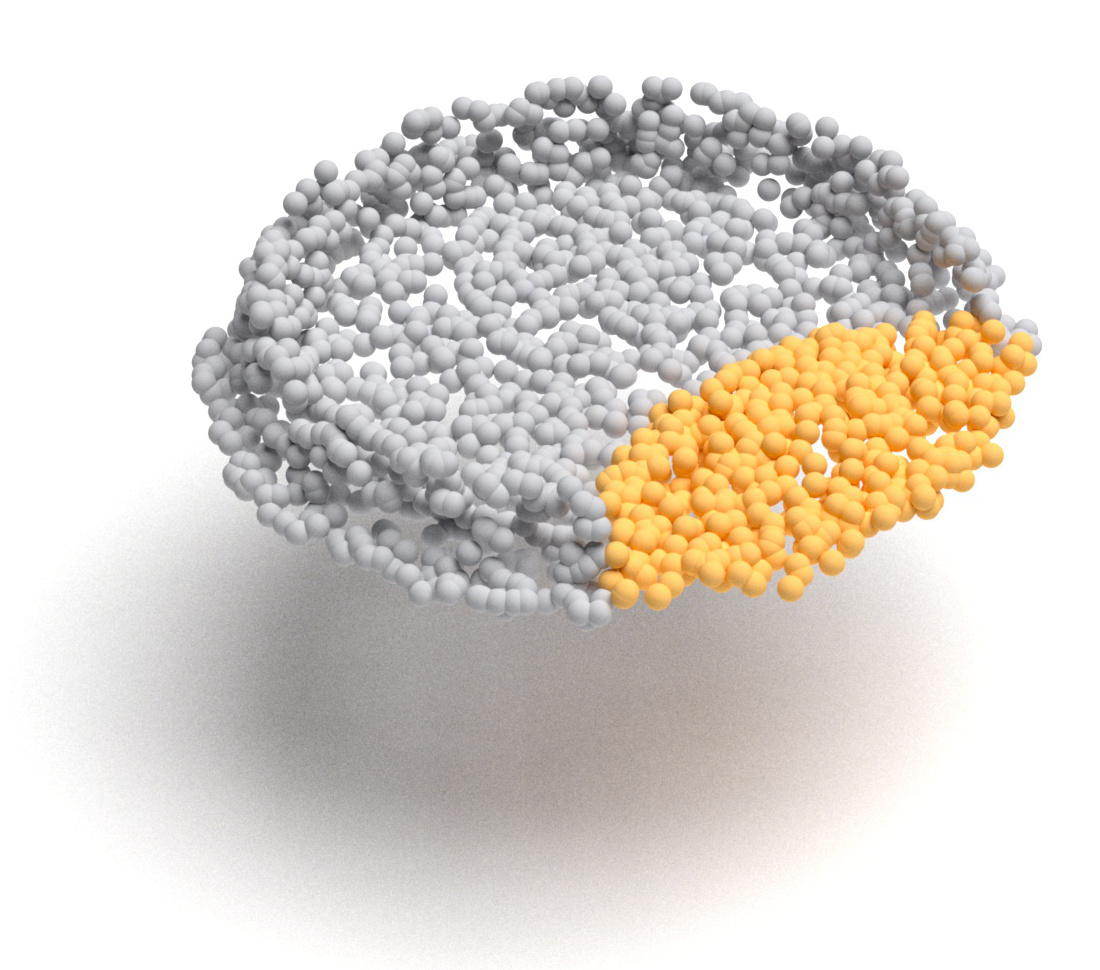} &
    \includegraphics[width=0.12\textwidth]{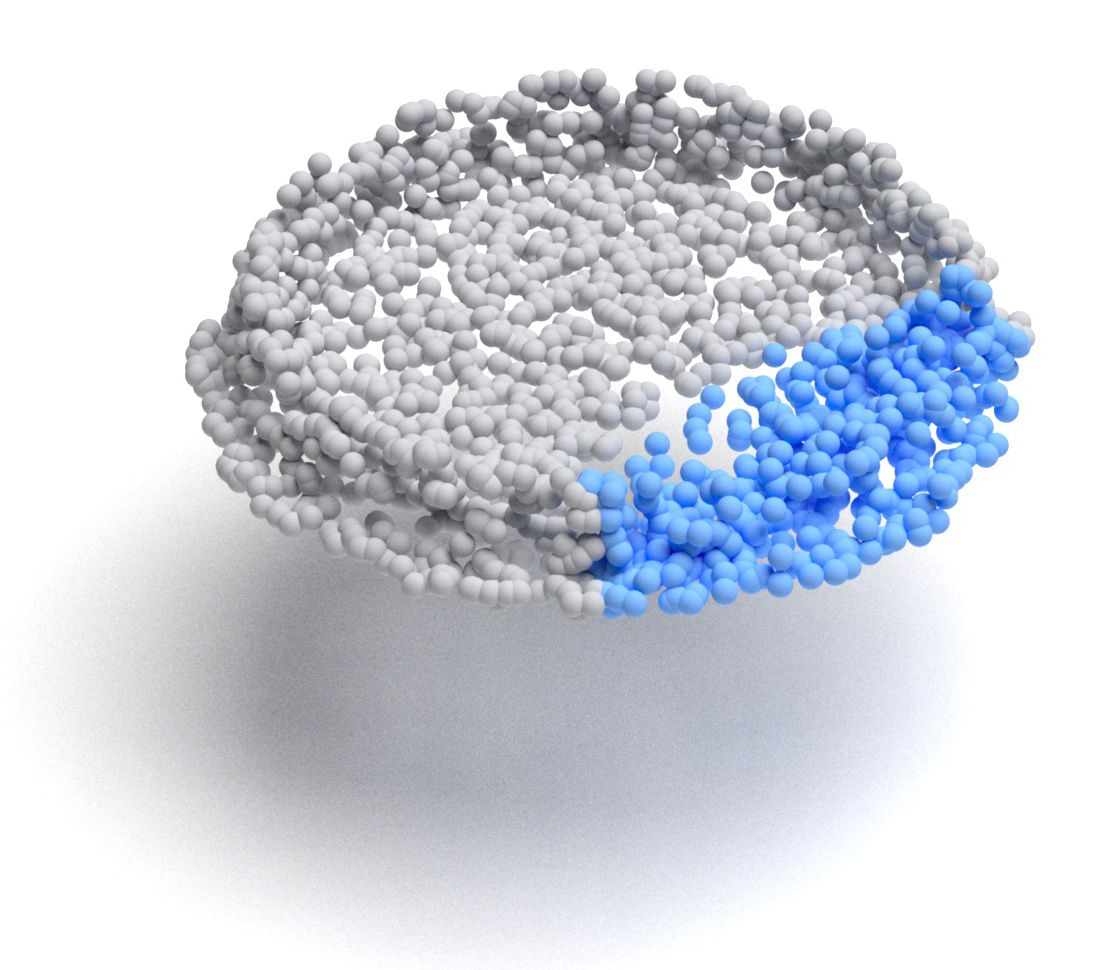} &
    \includegraphics[width=0.12\textwidth]{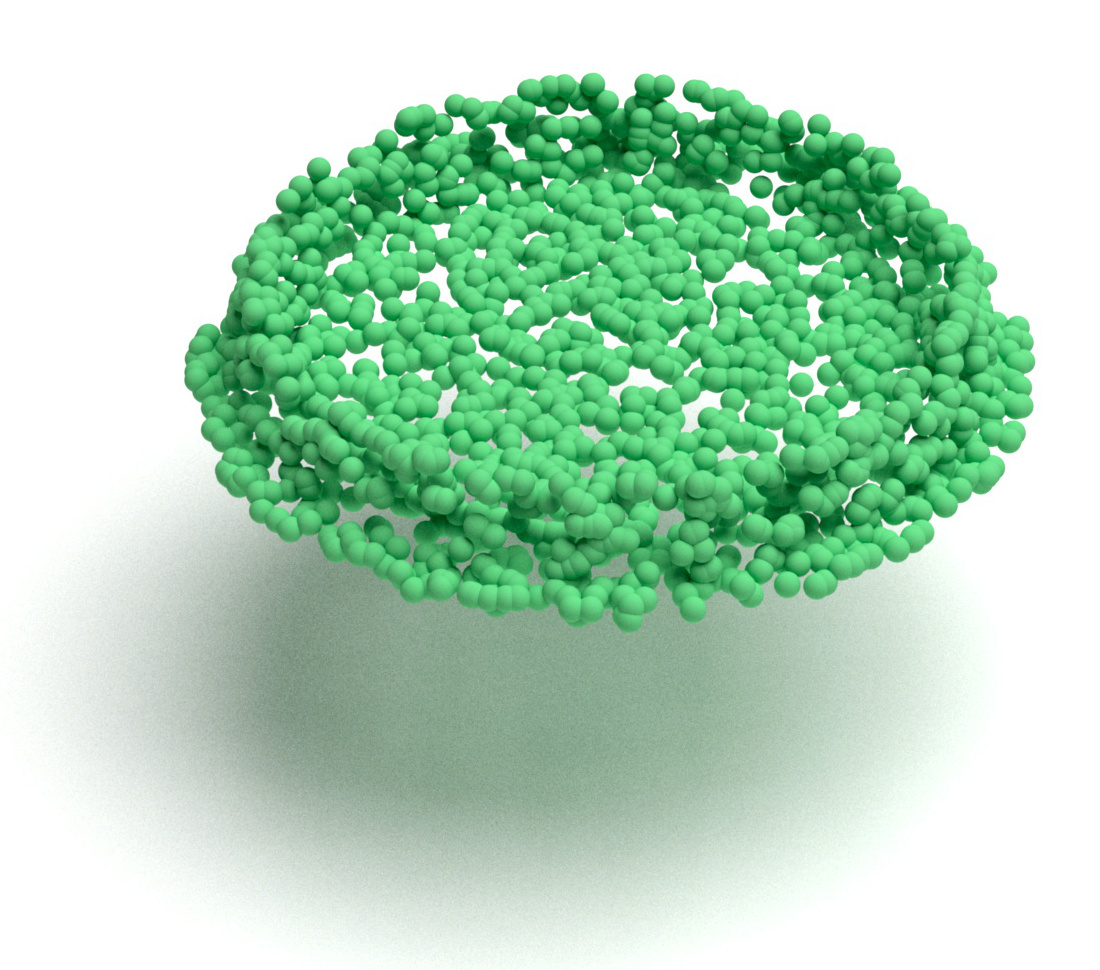}\\     
    \includegraphics[width=0.14\textwidth]{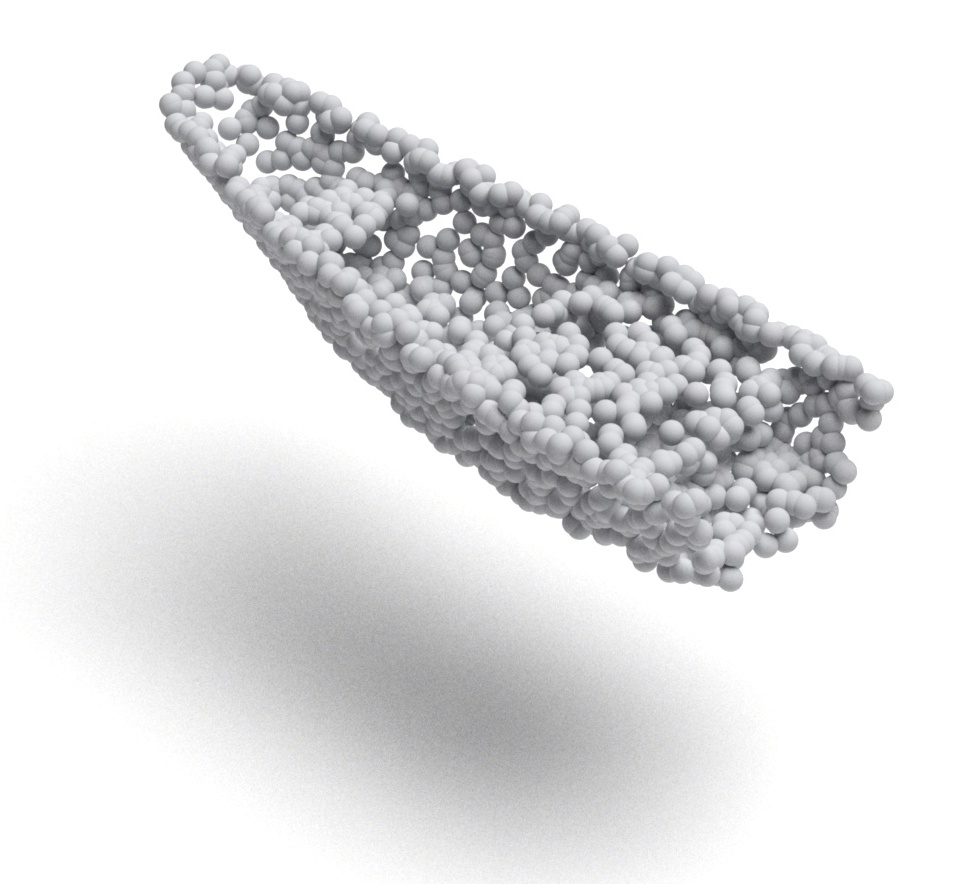} 
    & \includegraphics[width=0.14\textwidth]{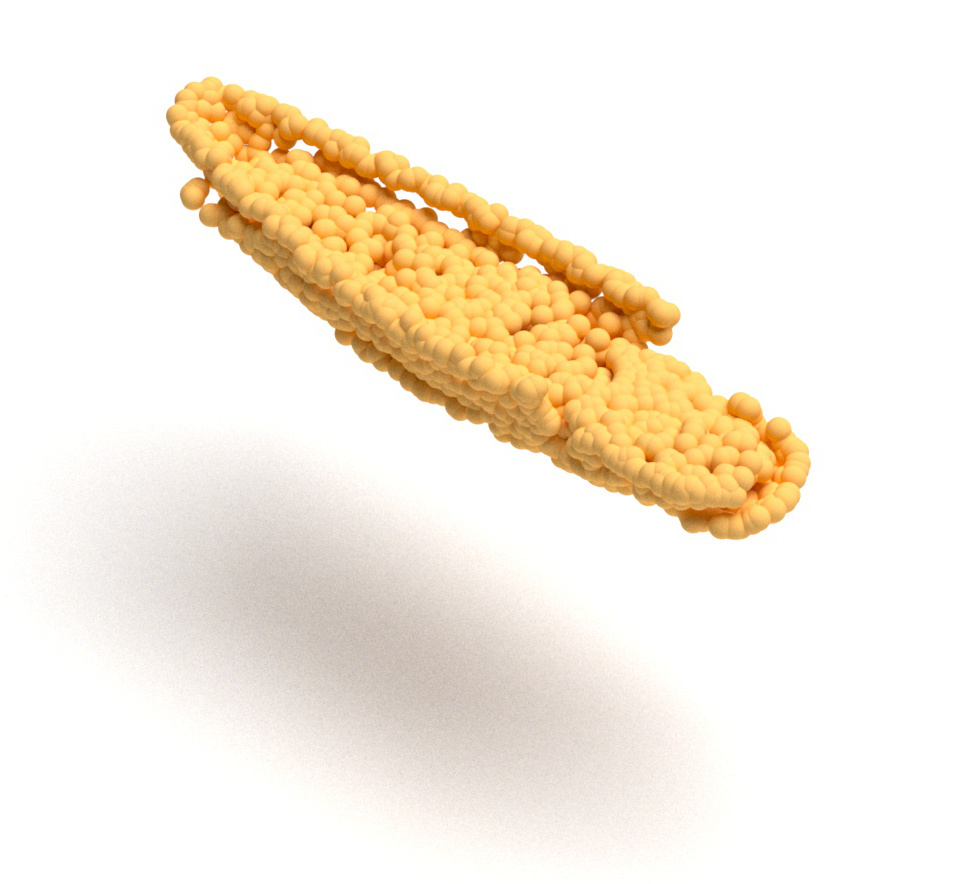}
    & \includegraphics[width=0.14\textwidth]{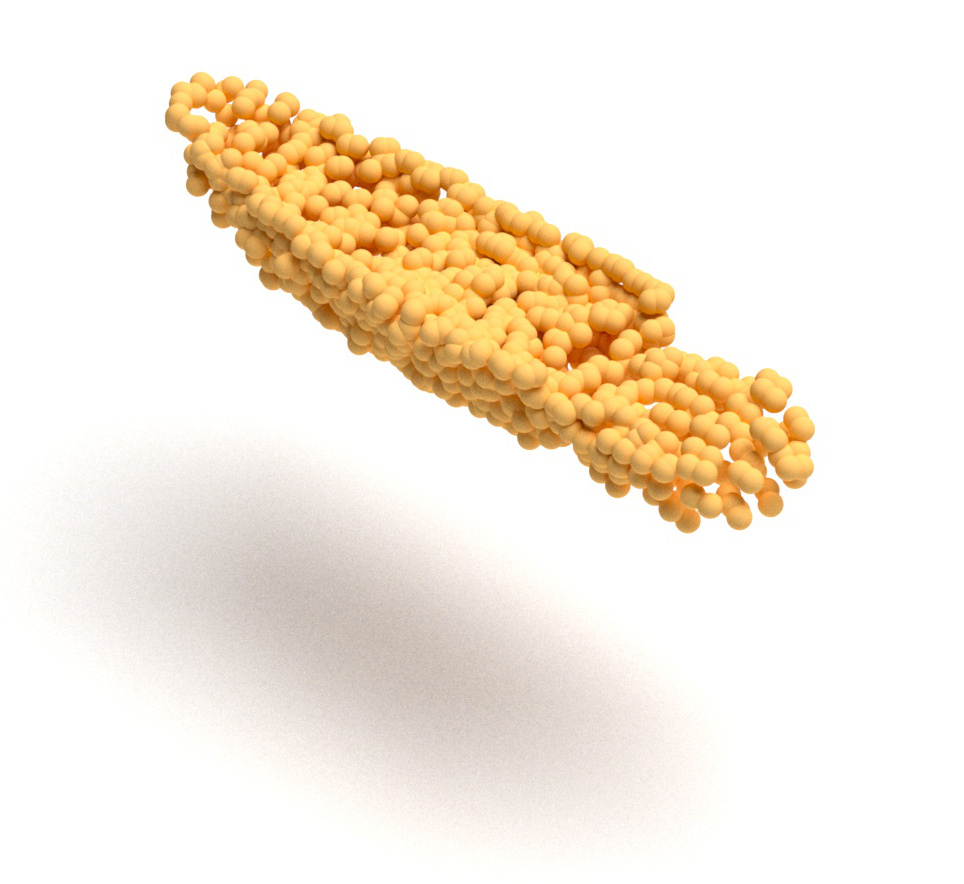}
    & \includegraphics[width=0.14\textwidth]{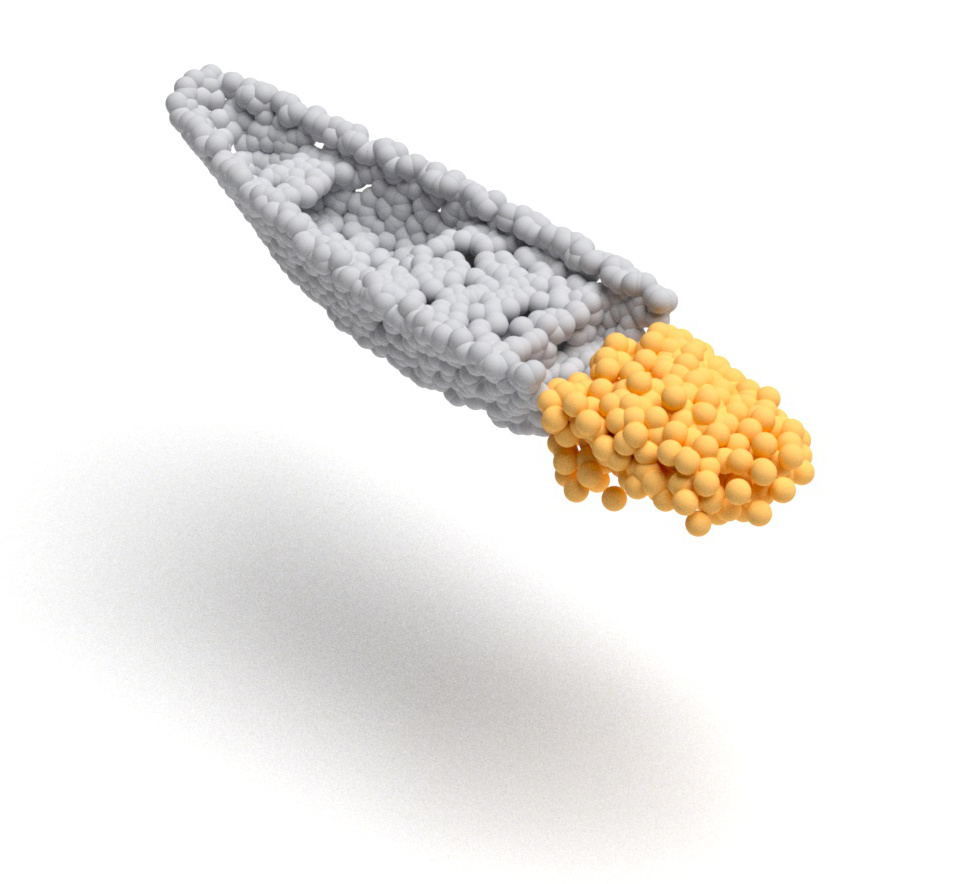}& \includegraphics[width=0.14\textwidth]{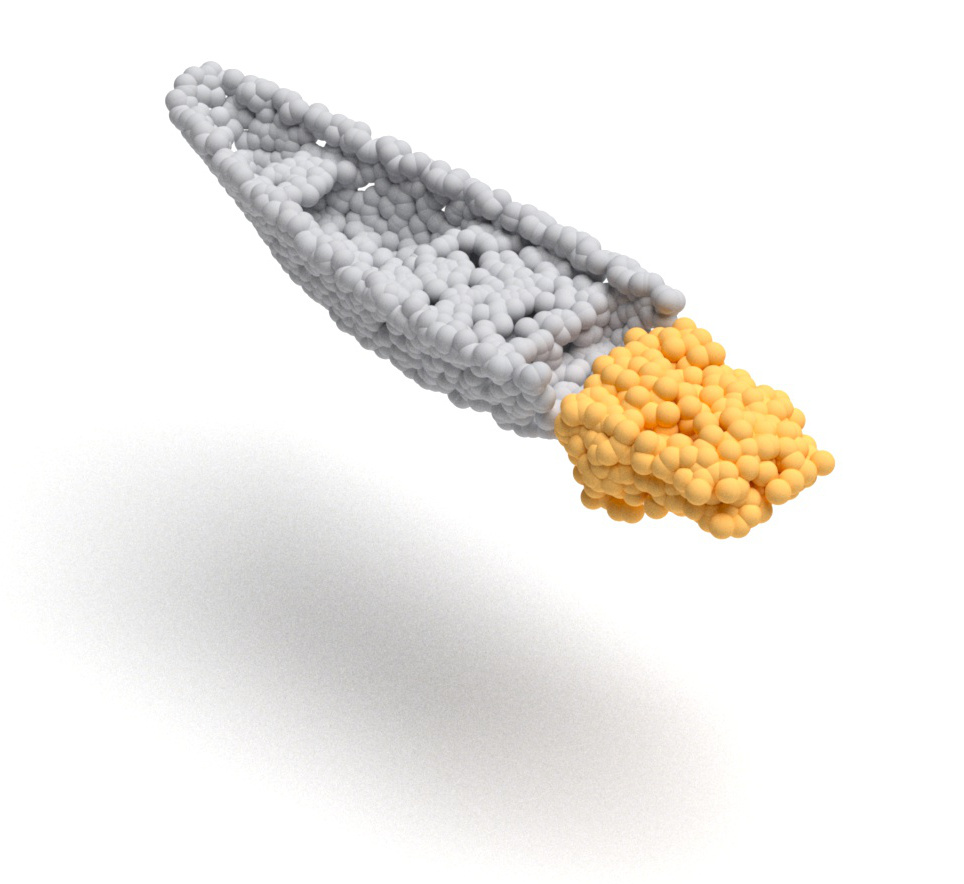} & \includegraphics[width=0.14\textwidth]{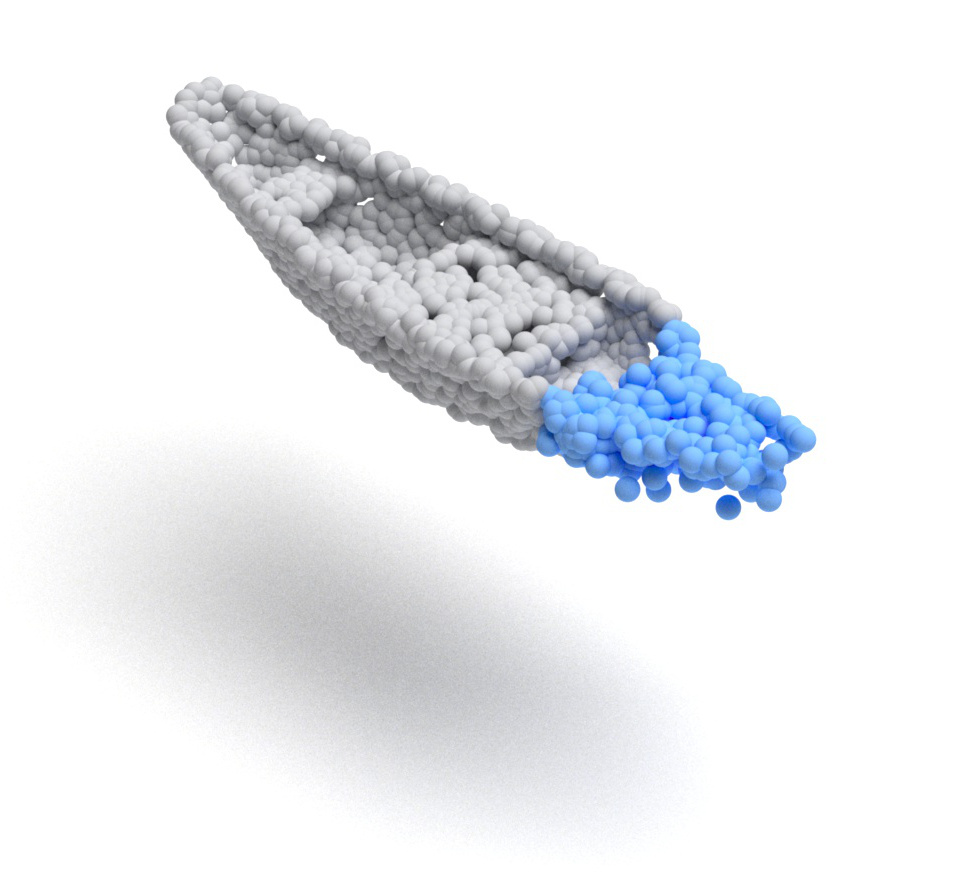} &
    \includegraphics[width=0.14\textwidth]{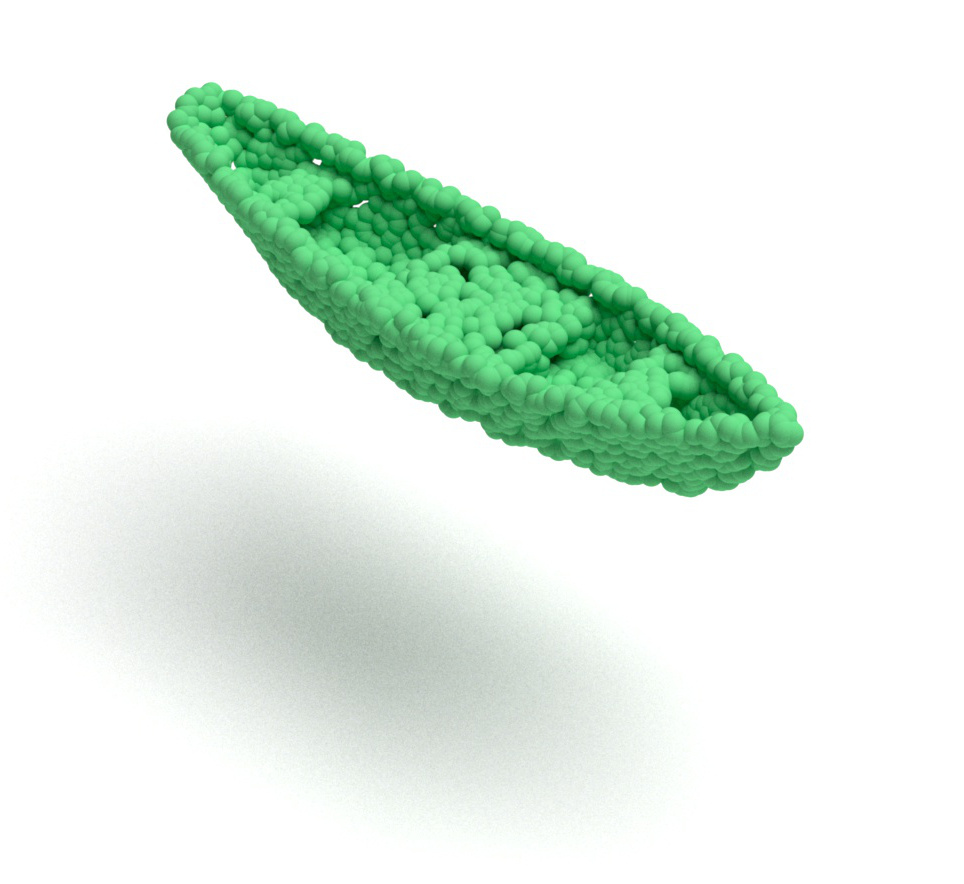}\\   
    \includegraphics[width=0.07\textwidth]{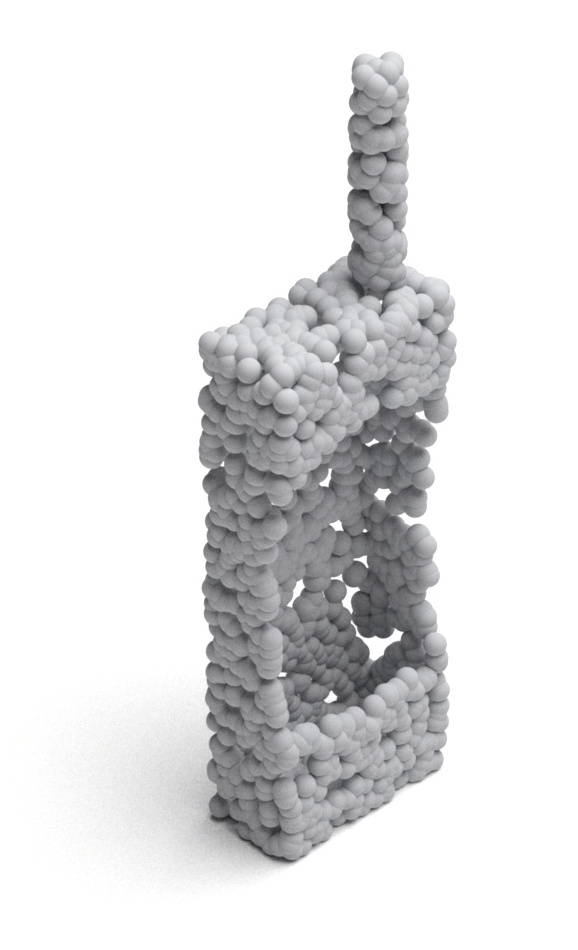} 
    & \includegraphics[width=0.07\textwidth]{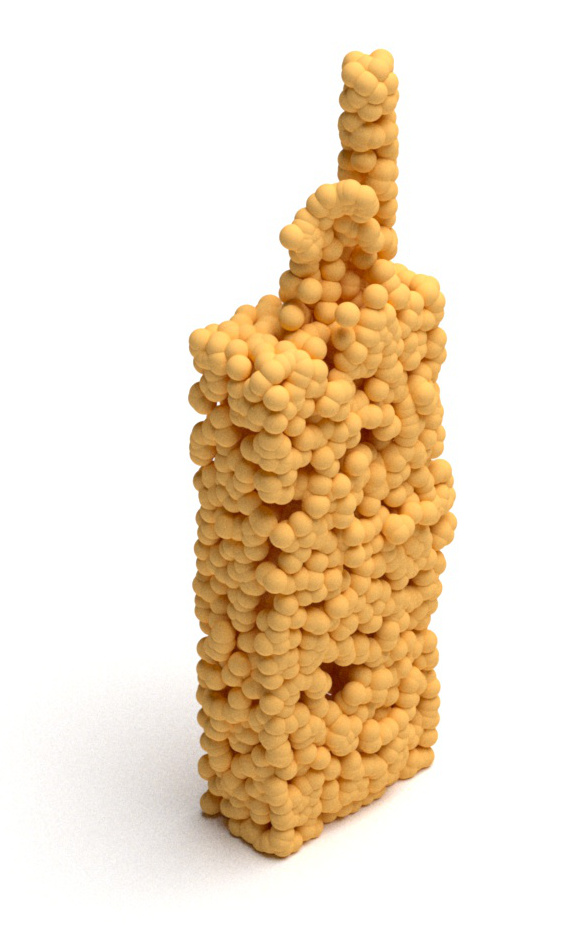}
    & \includegraphics[width=0.07\textwidth]{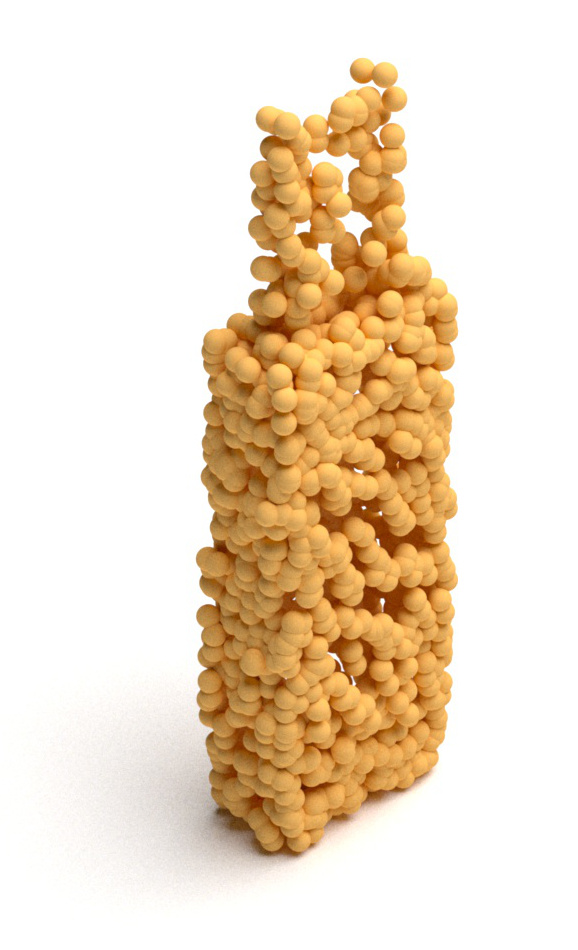}
    & \includegraphics[width=0.07\textwidth]{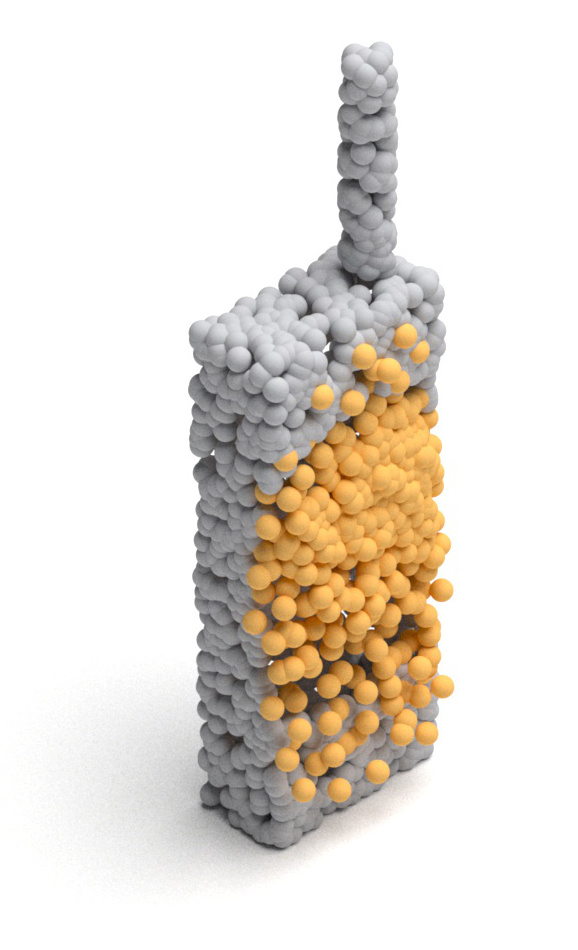}& \includegraphics[width=0.07\textwidth]{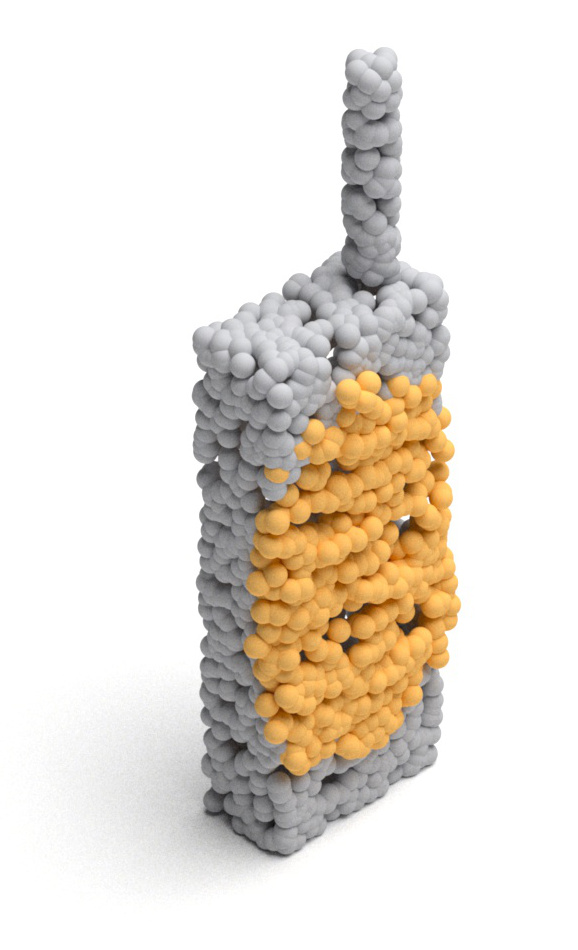} &
    \includegraphics[width=0.07\textwidth]{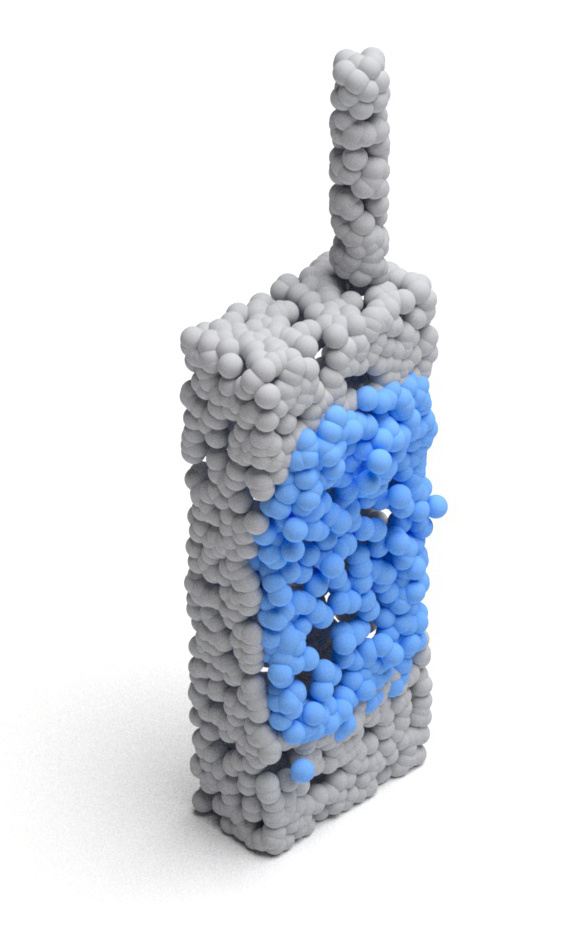} &
    \includegraphics[width=0.07\textwidth]{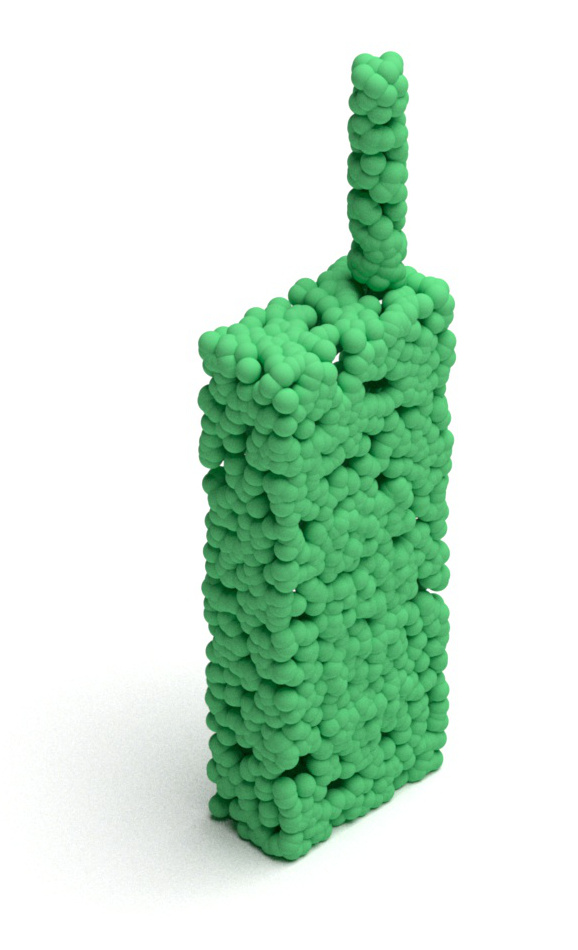}\\ \vspace{-2mm}
    \includegraphics[width=0.11\textwidth]{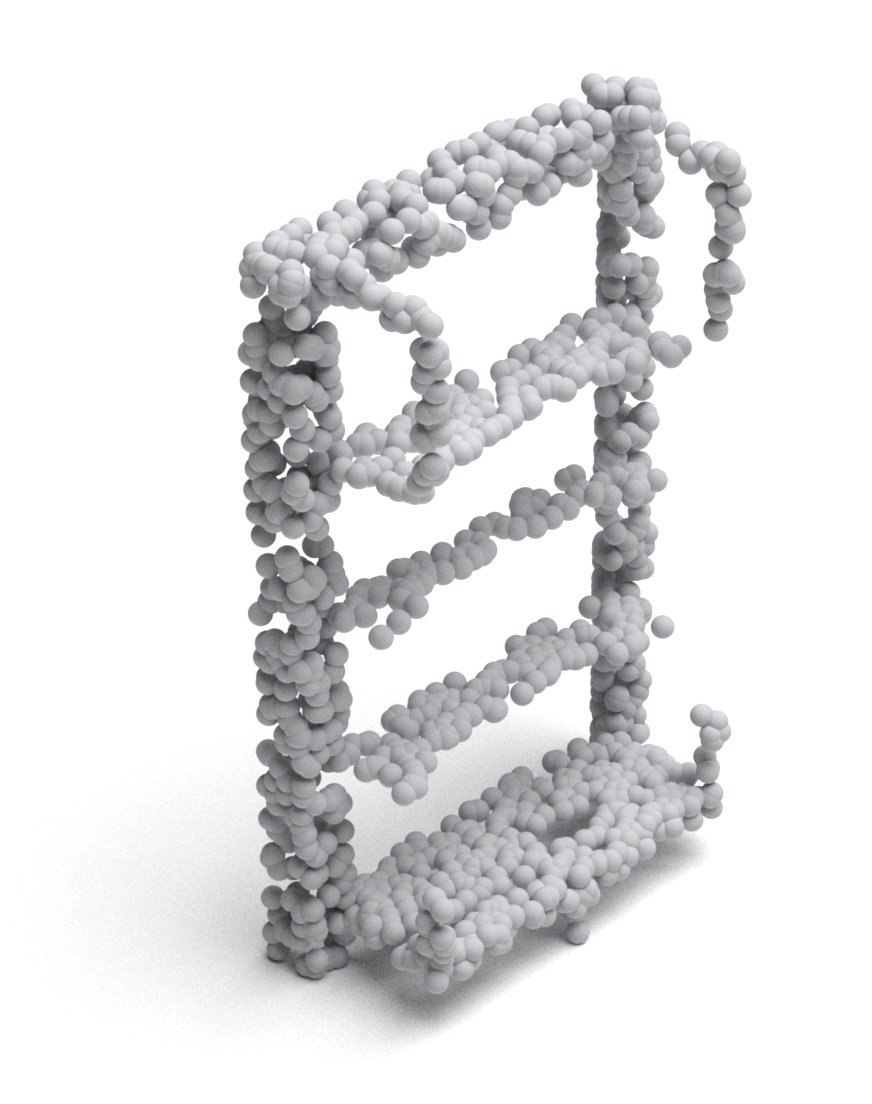} 
    & \includegraphics[width=0.11\textwidth]{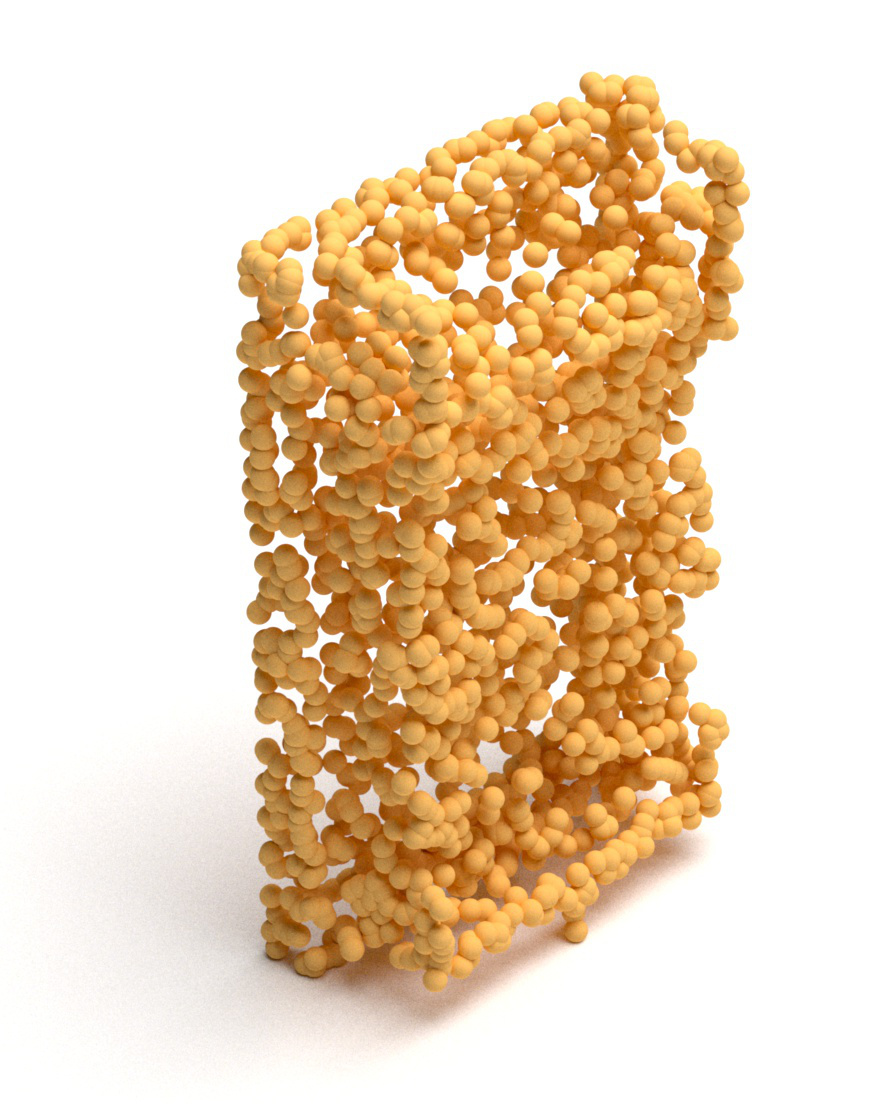}
    & \includegraphics[width=0.11\textwidth]{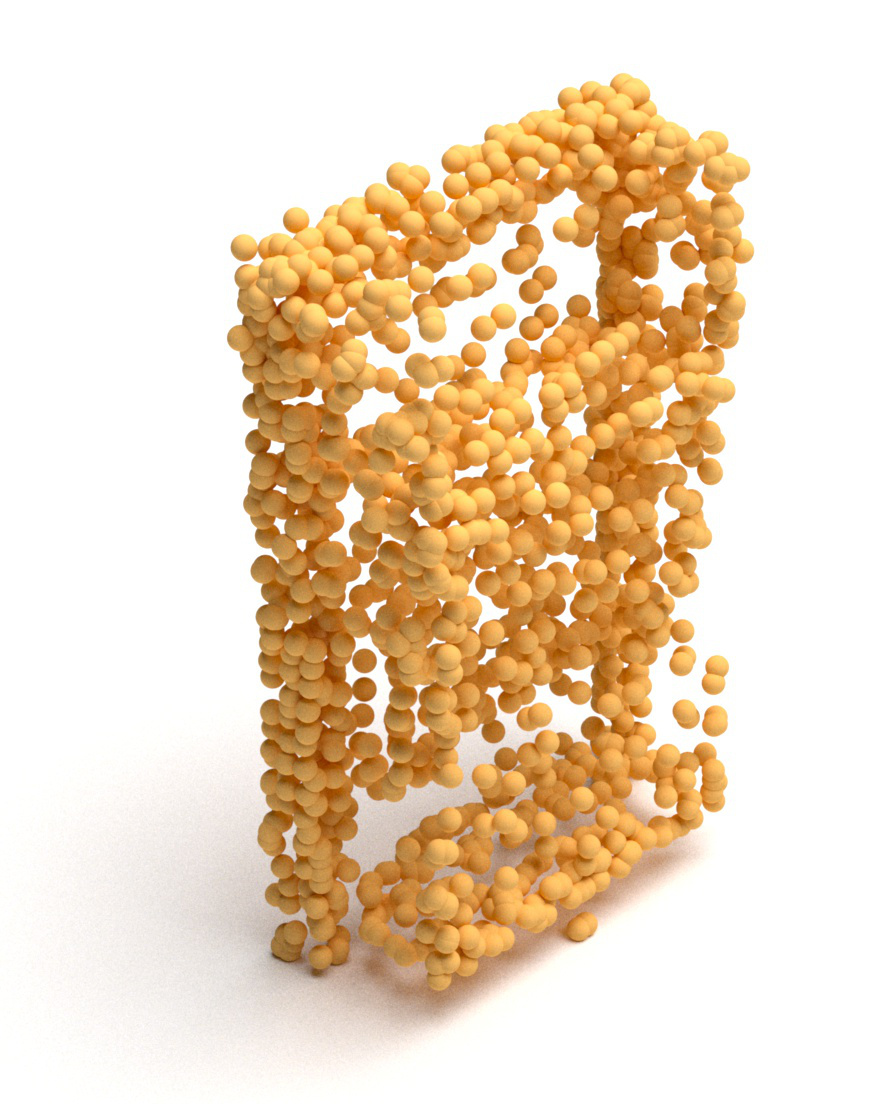}
    & \includegraphics[width=0.11\textwidth]{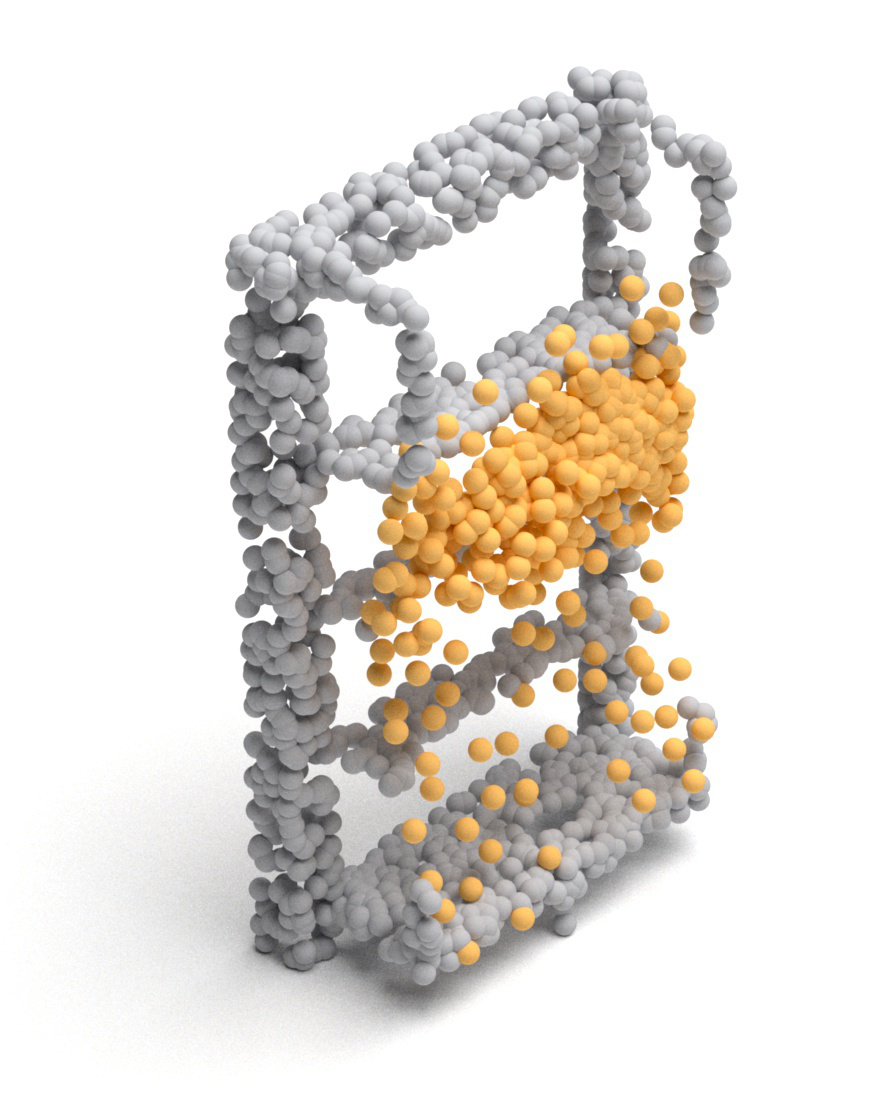}& \includegraphics[width=0.11\textwidth]{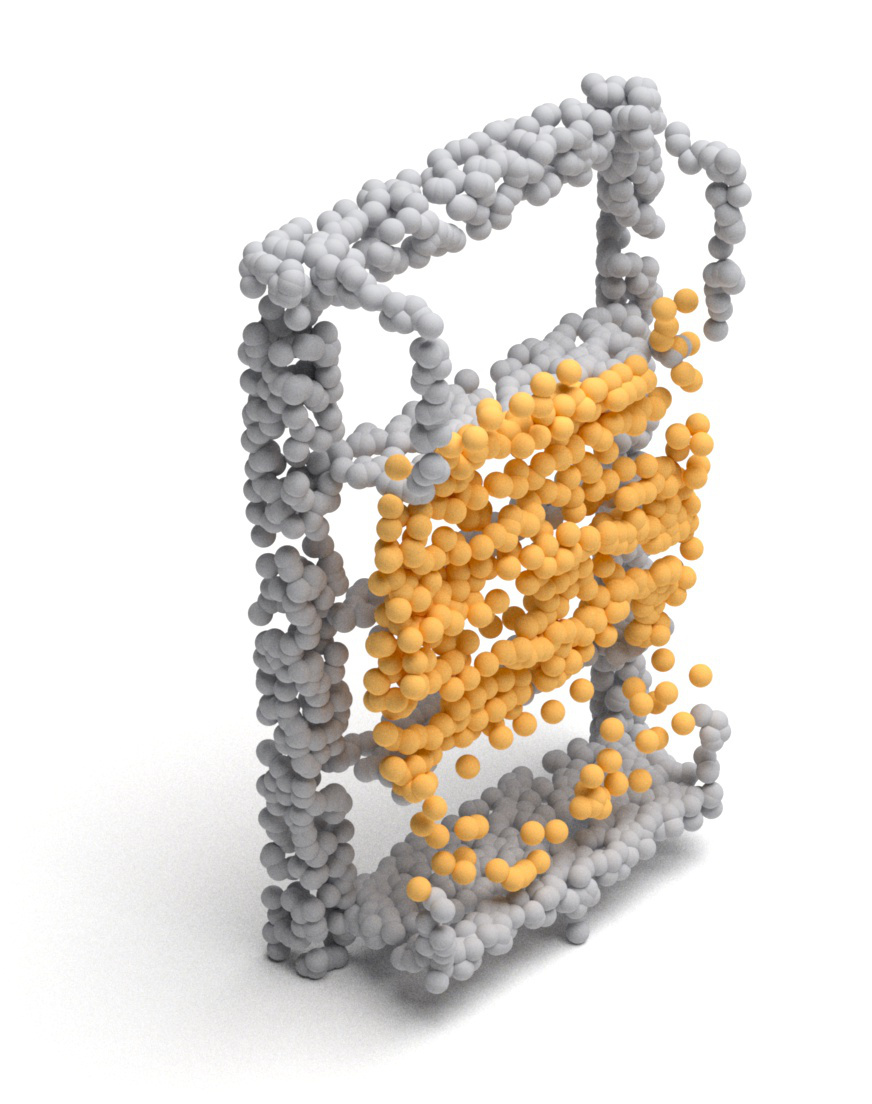} &
    \includegraphics[width=0.11\textwidth]{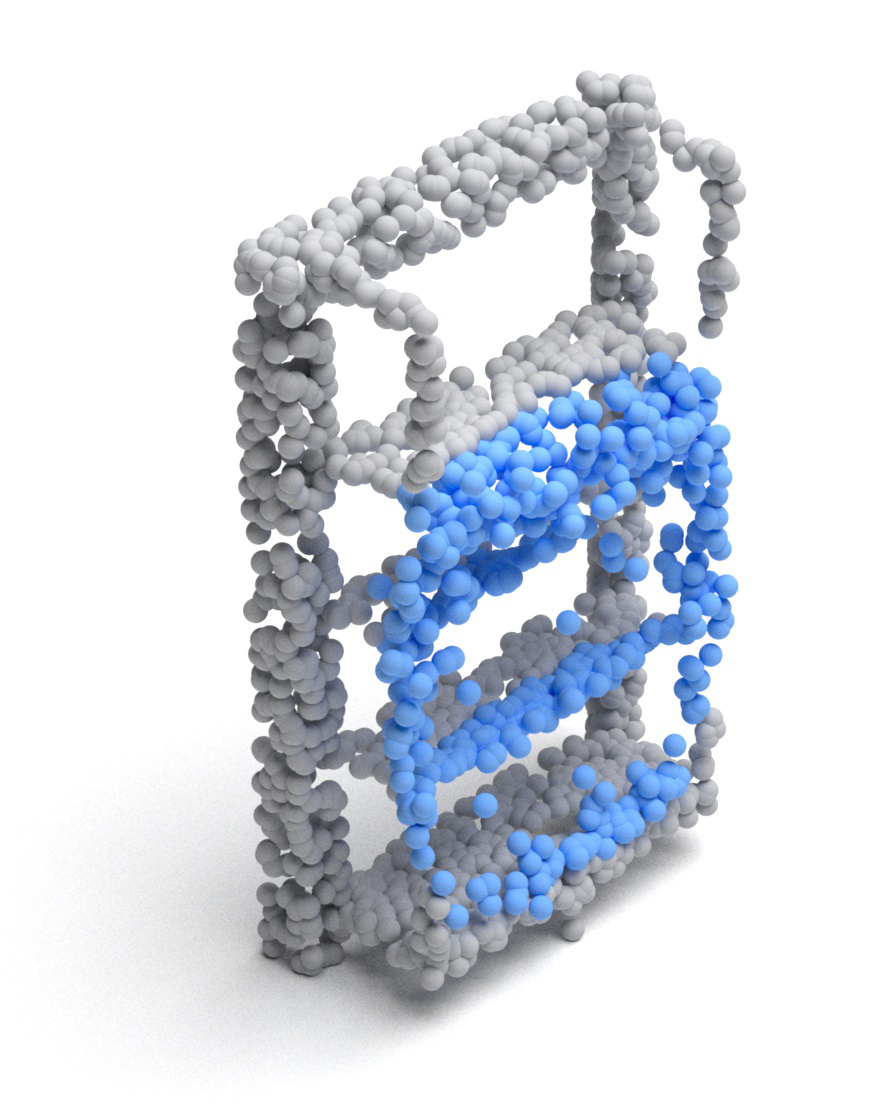} &
    \includegraphics[width=0.11\textwidth]{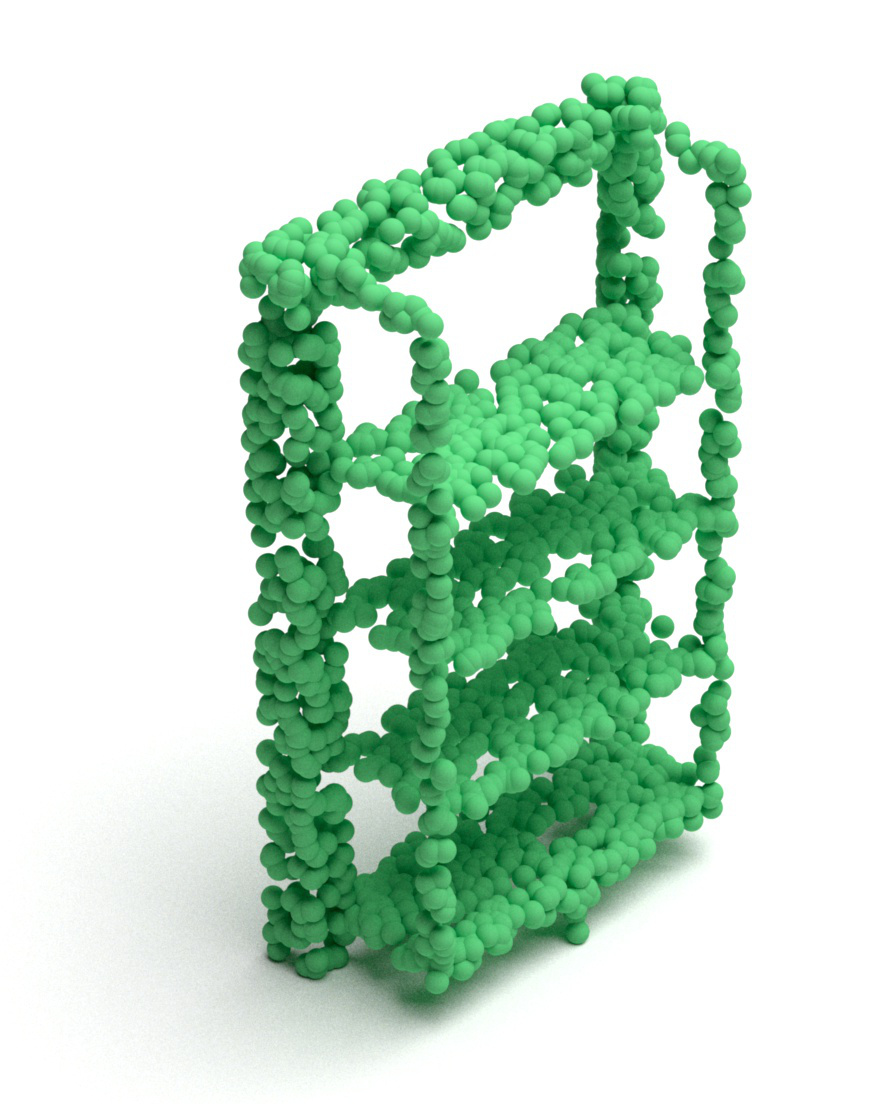}\\ 
    \includegraphics[width=0.14\textwidth]{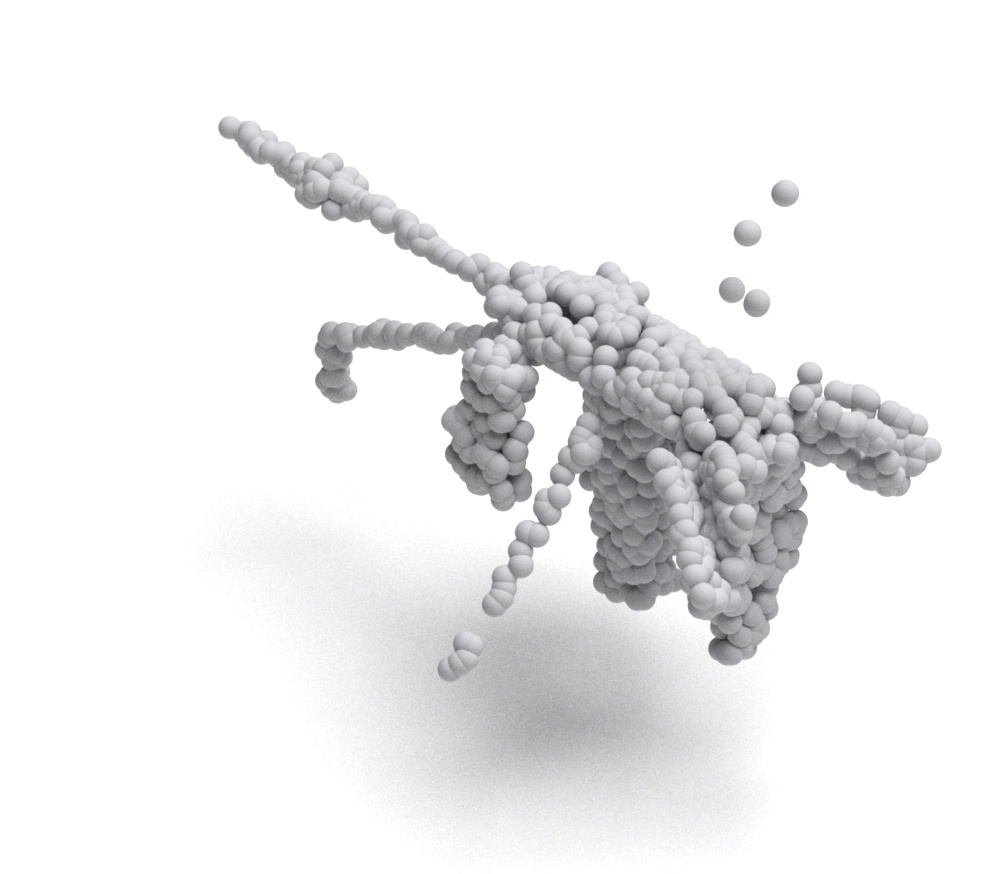} 
    & \includegraphics[width=0.14\textwidth]{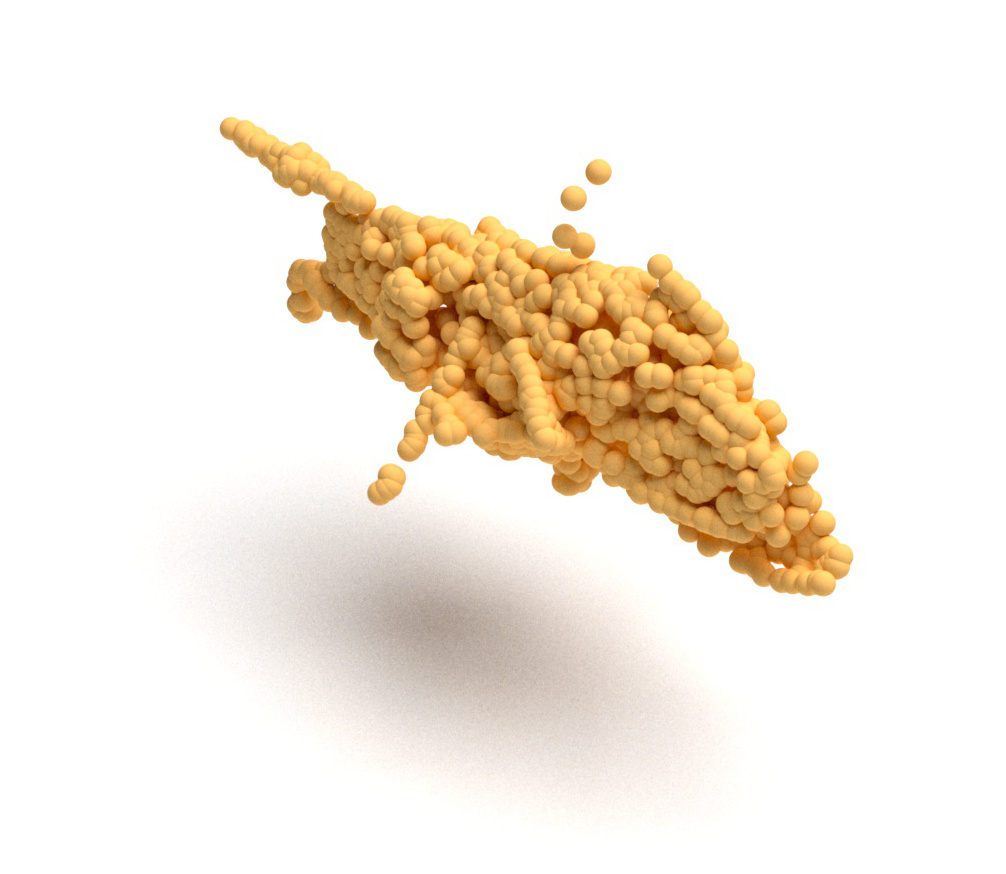}
    & \includegraphics[width=0.14\textwidth]{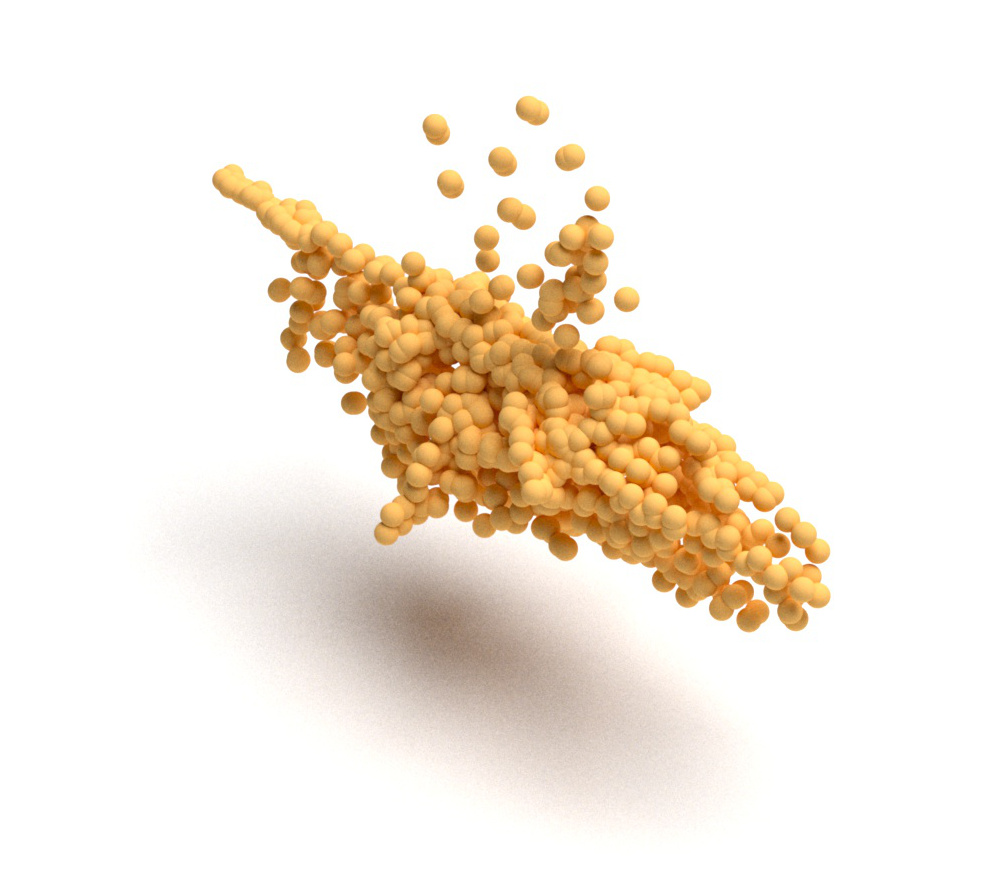}
    & \includegraphics[width=0.14\textwidth]{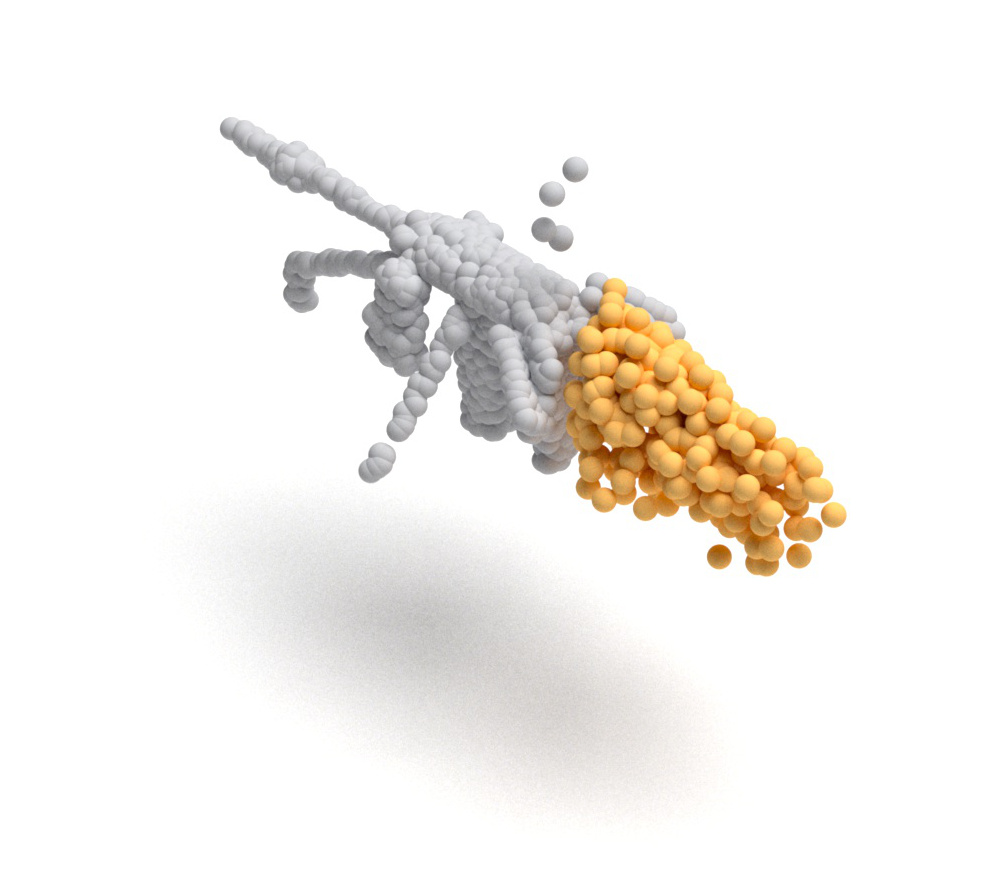}& \includegraphics[width=0.14\textwidth]{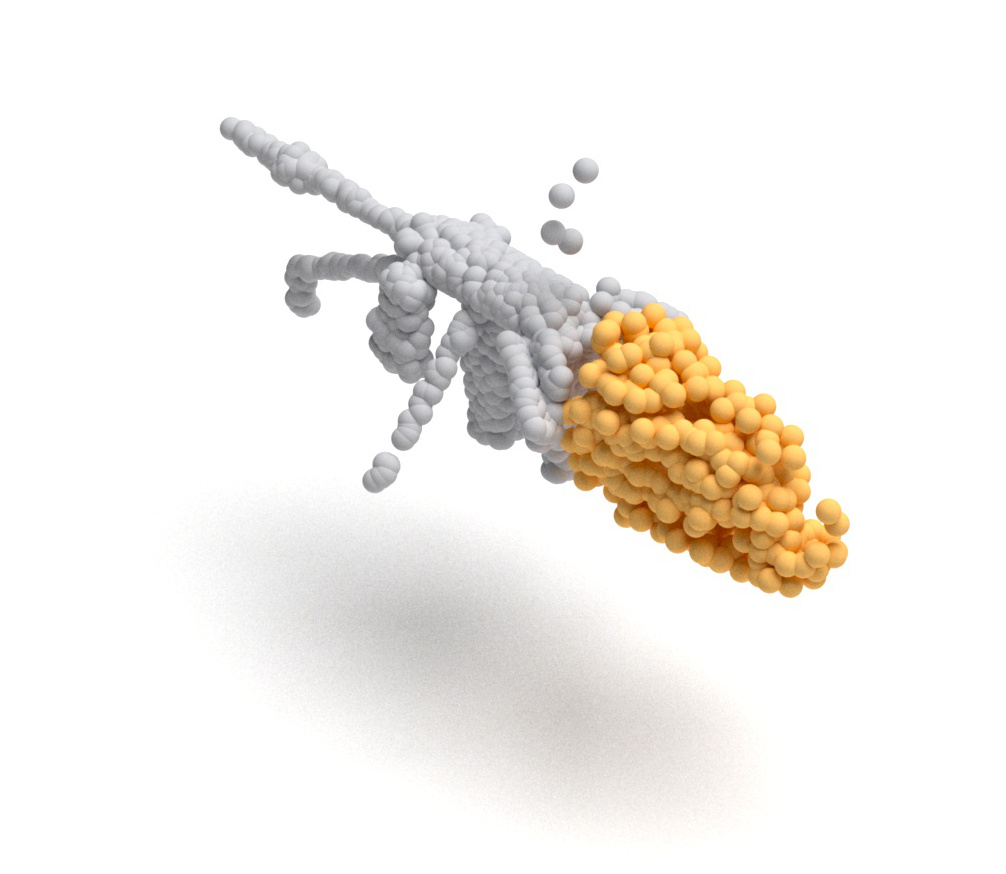} &
    \includegraphics[width=0.14\textwidth]{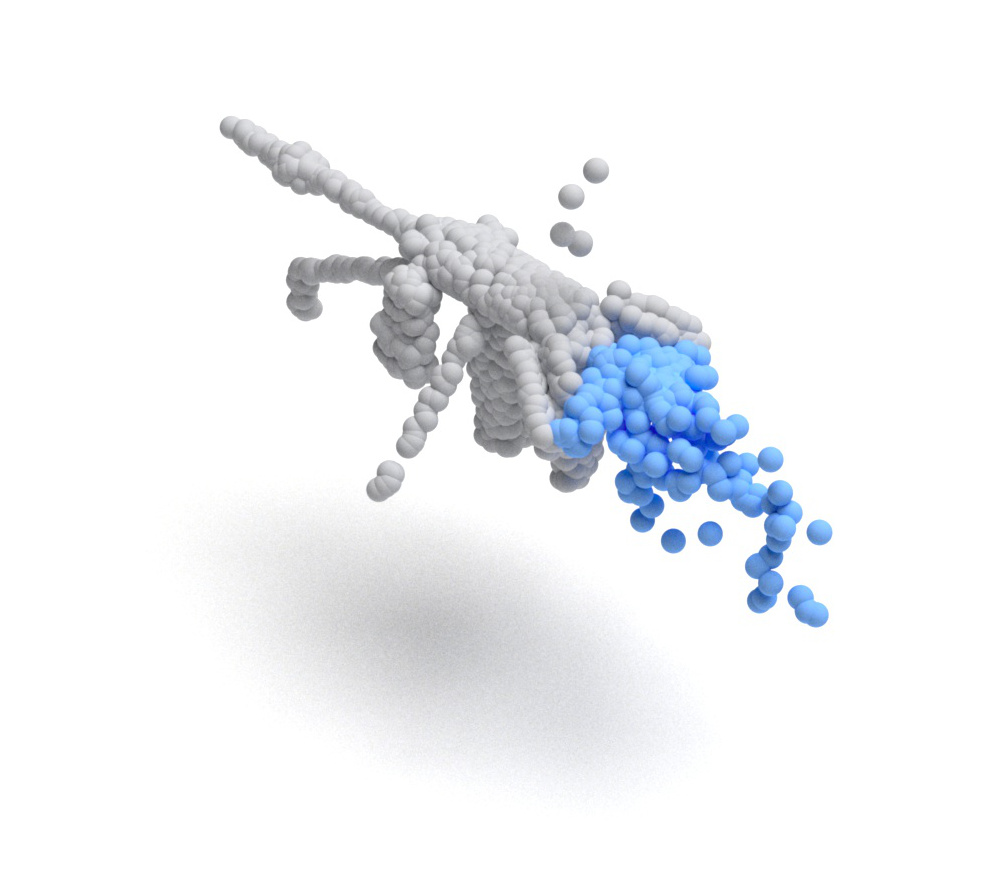} &
    \includegraphics[width=0.14\textwidth]{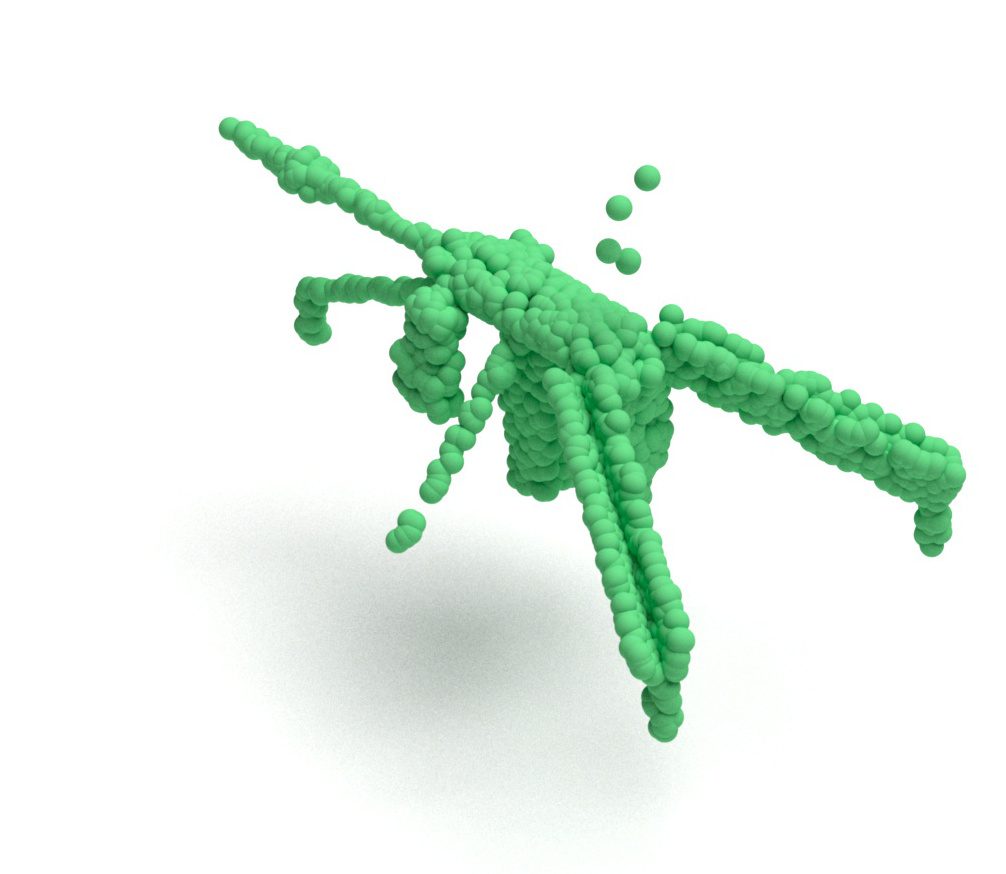}\\ 
    \end{tabular}
    \end{adjustbox}
\caption{\emph{Novel Categories - Qualitative}. Completion results for samples of class basket, bicycle, bowl, vessel, telephone, bookshelf and rifle (top to bottom row).}
\label{table:novelcat-qualitative}
\vspace{-3mm}
\end{figure*}

\subsection{Novel Categories}
\noindent\textbf{Quantitative Analysis} We extend our experimental analysis to novel categories, unseen at training time. Given its low reconstruction accuracy, we disregard PCN here while keeping all the other baselines. The quantitative results in Table \ref{table:novelcat-quantitative} show how \our outperforms all the considered competitors by a large margin regardless of the semantic relatedness between the known and new classes. 

\noindent\textbf{Qualitative Analysis} Figure \ref{table:novelcat-qualitative} collects the point clouds completed by the different considered methods. In general \our is the only approach that, besides not loosing information on the partial input, is able to fill the hole maintaining a smooth transition to the missing part as well as shape continuity  (\eg the boarder of the bowl, the pointy end of the vessel). 
The results of MSN and CRN are often noisy (\eg basket and bicycle) or present artifacts (\eg telephone), while the PF-Net variants are less precise than \our. The bookshelf can be considered a mild failure case: none of the approaches is able to complete correctly the second and third partially missing shelves. \our has the best overall appearance, also considering the details of the vertical shelf connections, but the second and first shelf get merged together. The failure is even more evident in the case of rifle in the last row.

\begin{figure*}[t!]
    \begin{center}
    \includegraphics[width=0.9\linewidth]{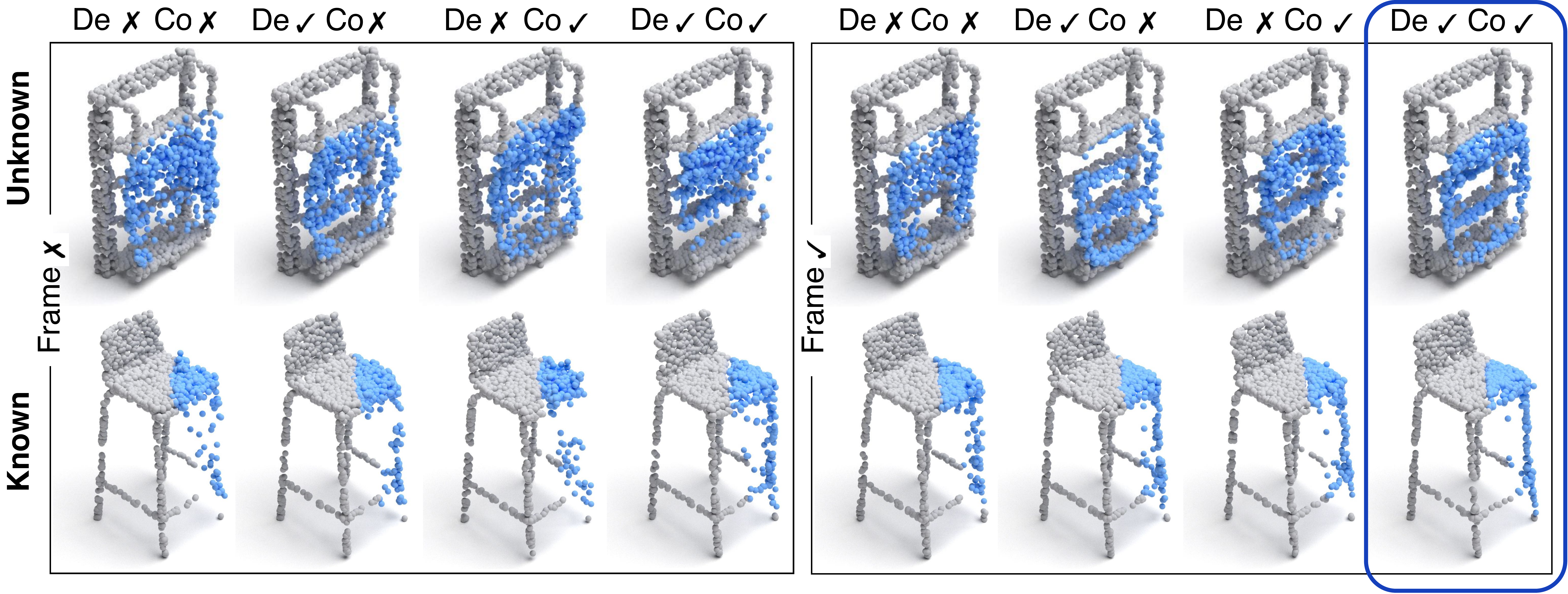}
    \end{center}
    \caption{\emph{Visual Ablation}. Qualitative results showing the impact of each single component of our model. First row: Unknown object Bookshelf. Second row: Known object Chair. Given the partial input points (grey), \our predicts the missing part points (blue). }
    \label{fig:visual_abl_newcat}
    \vspace{-3mm}
\end{figure*}

\subsection{Ablation Study}
We can identify three main components in \our: the local encoder, the global encoder and the auxiliary condition of reconstructing the frame region around the missing part. 
To carefully study the effect of each of them we perform extensive ablation experiments and organize Table \ref{table:ablation-single-hole} into three groups. The first and second groups analyze the benefits induced by the pretext tasks (denoising and contrastive learning) when respectively the frame spatial constraint is turned off and on; in the third group we consider at the global encoder a supervised classification pretext for comparison with our contrastive formulation. 

If we focus on the known classes, the first row in Table \ref{table:ablation-single-hole} indicates that, by turning off all the pre-trainings and the frame constraint, we get a CD (23.865) similar to MSN (22.410) but higher than what obtained by PF-Net (20.445). This indicates at the same time a good backbone design for \our, as well as clear room for improvement in the training procedure. 
For the unknown classes our basic architecture outperforms (26.811) the competitors when dealing with novel classes similar to the known ones (best case: PF-Net Vanilla 31.232), but it is significantly worse (40.419) in case of dissimilar ones (best case: CRN 34.625). This confirms that the backbone per-se is not able to generalize.

The following rows in the first part of the table show how both local denoising and global contrastive pre-training lead to lower reconstruction errors. The former appears more effective than the latter: indeed local information is extremely relevant to reconstruct the shape details. Still, their combination always produce a further accuracy gain, obtaining already state of the art results (\our known/unknown sim./unknown diss.: (18.742 / 19.364 / 32.945) vs (PF-Net 20.445 / PF-Net van. 31.232 / CRN 34.625)). 

The second group of results in the table highlights the effect of including the frame regularization: it provides an evident performance uplift, validating our hypothesis of better blending the reconstructed points with the existing ones.

Finally, the bottom part of the table presents the effect of substituting our unsupervised contrastive pretext with a more informed supervised one. Although this choice appears valuable and allows to outperforms the top competitors, the obtained reconstruction error results are worse than what obtained by \our. The difference is particularly evident on the unknown classes where the information coming from the closed-world classification task is clearly unable to support generalization. Our unsupervised local and global pretext tasks provide less distortions in the generated missing part besides not requiring costly labeled data. The effect of each component of our model is also shown by the visual ablation in Figure \ref{fig:visual_abl_newcat}.

\begin{table}[t!]
    \centering
    \begin{adjustbox}{width=0.48\textwidth}
    \begin{tabular}{@{}c@{~~} |c@{~~}c@{~~}|c@{~~}|c@{~~}|c@{~~}|c@{~~}}
    \hline
    \textbf{Local} & \multicolumn{2}{c|}{\textbf{Global}} & \multirow{2}{*}{\textbf{Frame}} & \multicolumn{3}{c}{\textbf{Overall}}\\
    {Denoise} & {Cla.} & {Contr.} &  & Known & Unknown Sim. & Unknown Diss.\\
    \hline
    \xmark & \xmark & \xmark & \xmark & 23.865 & 26.811 & 40.419 \\
    \cmark & \xmark & \xmark & \xmark & 21.022 & 22.213 & 33.602 \\ 
    \xmark & \xmark & \cmark & \xmark & 23.067 & 25.723  & 36.567 \\
    \cmark & \xmark & \cmark & \xmark & 18.742 & 19.364 & 32.945 \\
    \hline
    \xmark & \xmark & \xmark & \cmark & 20.586 & 22.538 & 32.661 \\
    \cmark & \xmark & \xmark & \cmark & 18.131 & 20.064 & 31.824 \\
    \xmark & \xmark & \cmark & \cmark & 18.995 & 20.989 & 32.070 \\
    \cmark & \xmark & \cmark & \cmark & \textbf{16.517} & \textbf{17.680} & \textbf{28.403} \\
    \hline
    \xmark & \cmark & \xmark & \xmark & 24.364 & 29.470 & 40.269 \\
    \cmark & \cmark & \xmark & \xmark & 20.272 & 21.187 & 34.592\\
    \xmark & \cmark & \xmark & \cmark & 19.378 & 21.942 & 32.456 \\
    \cmark & \cmark & \xmark & \cmark & 18.699 & 21.712 & 31.645 \\
    \hline
    \end{tabular}
    \end{adjustbox}
\caption{\emph{Ablation analysis}. Chamfer Distance scaled by $10^4$. The results show the gain provided by each single component of our model.
At the global encoder, we also compare the Contrastive pretext (\emph{Contr.}) with a supervised Classification pretext (\emph{Cla.}). 
}
\label{table:ablation-single-hole}
\vspace{-3mm}
\end{table}

\section{Conclusions}
In this work we introduced a new point cloud completion method that encodes shape knowledge via two self-supervised pretext task: denoising to gather local cues and contrastive learning for global information.  Our \our focuses on reconstructing the missing part of the point cloud by also exploiting a context frame region as anchor reference: it avoids to re-generate the whole shape while keeping strong spatial continuity with the observed partial input. The obtained completion results as well as the conducted ablation and robustness studies indicate that \our outperforms existing competitors defining the new state of the art on the standard closed-class setting.  Moreover we extensively evaluated \our on novel categories, further showing the effectiveness of our approach.

How to deal with fine-grained structured objects as rifles or modern design lamp and bookshelves remains a challenging open question for all point cloud completion methods. For the future we plan to extend \our in this direction as well as on real-world scans where the missing part issue comes along with extremely sparse partial inputs. 

\noindent\textbf{Acknowledgements} This work was partially supported by the CHIST-ERA BURG Project (TT).
We also acknowledge that the research activity herein was carried out using the IIT HPC infrastructure.
{\small
\bibliographystyle{ieee_fullname}
\bibliography{egbib}

\begin{thebibliography}{10}\itemsep=-1pt

\bibitem{Chang_shapenet_2015}
Angel~X. Chang, Thomas Funkhouser, Leonidas Guibas, Pat Hanrahan, Qixing Huang,
  Zimo Li, Silvio Savarese, Manolis Savva, Shuran Song, Hao Su, Jianxiong Xiao,
  Li Yi, and Fisher Yu.
\newblock {ShapeNet: An Information-Rich 3D Model Repository}.
\newblock Technical Report arXiv:1512.03012 [cs.GR], Stanford University ---
  Princeton University --- Toyota Technological Institute at Chicago, 2015.

\bibitem{Chen_2020_IEEEspm_autonomousdrive}
Siheng Chen, Baoan Liu, Chen Feng, Carlos Vallespi-Gonzalez, and Carl
  Wellington.
\newblock 3d point cloud processing and learning for autonomous driving.
\newblock {\em IEEE Signal Processing Magazine}, May 2020.

\bibitem{chen_2020_icml_simclr}
Ting Chen, Simon Kornblith, Mohammad Norouzi, and Geoffrey Hinton.
\newblock A simple framework for contrastive learning of visual
  representations.
\newblock In {\em ICML}, 2020.

\bibitem{dai_2017_cvpr_voxel}
Angela Dai, Charles~Ruizhongtai Qi, and Matthias Nie{\ss}ner.
\newblock Shape completion using 3d-encoder-predictor cnns and shape synthesis.
\newblock In {\em CVPR}, 2017.

\bibitem{DavisMGL_2002_geometry}
James Davis, Stephen~R. Marschner, Matt Garr, and Marc Levoy.
\newblock Filling holes in complex surfaces using volumetric diffusion.
\newblock In {\em 3DPVT}, 2002.

\bibitem{Dosovitskiy_nips_2014_instanceContrast}
Alexey Dosovitskiy, Jost~Tobias Springenberg, Martin Riedmiller, and Thomas
  Brox.
\newblock Discriminative unsupervised feature learning with convolutional
  neural networks.
\newblock In {\em NIPS}, 2014.

\bibitem{groueix_2018_CVPR_mesh}
Thibault Groueix, Matthew Fisher, Vladimir~G. Kim, Bryan Russell, and Mathieu
  Aubry.
\newblock {AtlasNet: A Papier-M\^ach\'e Approach to Learning 3D Surface
  Generation}.
\newblock In {\em CVPR}, 2018.

\bibitem{Han_2017_iccv_voxel}
X. Han, Z. Li, H. Huang, E. Kalogerakis, and Y. Yu.
\newblock High-resolution shape completion using deep neural networks for
  global structure and local geometry inference.
\newblock In {\em ICCV}, 2017.

\bibitem{He_2020_CVPR_moco}
Kaiming He, Haoqi Fan, Yuxin Wu, Saining Xie, and Ross Girshick.
\newblock Momentum contrast for unsupervised visual representation learning.
\newblock In {\em CVPR}, 2020.

\bibitem{hou_cvpr_2019_sceneunderstanding}
Ji Hou, Angela Dai, and Matthias Nie{\ss}ner.
\newblock 3d-sis: 3d semantic instance segmentation of rgb-d scans.
\newblock In {\em CVPR}, 2019.

\bibitem{Huang_2020_CVPR_pfnet}
Zitian Huang, Yikuan Yu, Jiawen Xu, Feng Ni, and Xinyi Le.
\newblock Pf-net: Point fractal network for 3d point cloud completion.
\newblock In {\em CVPR}, 2020.

\bibitem{Kazhdan_2013_TOG_geometry}
Michael Kazhdan and Hugues Hoppe.
\newblock Screened poisson surface reconstruction.
\newblock {\em ACM Transactions on Graphics (TOG)}, 32(3), 2013.

\bibitem{adam}
Diederik~P. Kingma and Jimmy Ba.
\newblock Adam: {A} method for stochastic optimization.
\newblock In Yoshua Bengio and Yann LeCun, editors, {\em ICLR}, 2015.

\bibitem{Lee_2019_icml_sagpool}
Junhyun Lee, Inyeop Lee, and Jaewoo Kang.
\newblock Self-attention graph pooling.
\newblock In {\em ICML}, 2019.

\bibitem{li_2015_datadriven}
Yangyan Li, Angela Dai, Leonidas Guibas, and Matthias Nie{\ss}ner.
\newblock Database-assisted object retrieval for real-time 3d reconstruction.
\newblock In {\em Computer Graphics Forum}. Wiley Online Library, 2015.

\bibitem{Liu_2020_AAAI_morphing}
Minghua Liu, Lu Sheng, Sheng Yang, Jing Shao, and Shi-Min Hu.
\newblock Morphing and sampling network for dense point cloud completion.
\newblock In {\em AAAI}, 2020.

\bibitem{Liu_2020_CVR_augmentedreal}
Weiquan Liu, Baiqi Lai, Cheng Wang, Xuesheng Bian, Wentao Yang, Yan Xia,
  Xiuhong Lin, Shang{-}Hong Lai, Dongdong Weng, and Jonathan Li.
\newblock Learning to match 2d images and 3d lidar point clouds for outdoor
  augmented reality.
\newblock In {\em IEEE VR}, 2020.

\bibitem{Mitra_2006_TOG_symmetry}
Niloy~J. Mitra, Leonidas~J. Guibas, and Mark Pauly.
\newblock Partial and approximate symmetry detection for 3d geometry.
\newblock {\em ACM Transactions on Graphics (TOG)}, 25(3):560–568, 2006.

\bibitem{nie_2020_NIPS_skeletonbridged}
Yinyu Nie, Yiqun Lin, Xiaoguang Han, Shihui Guo, Jian Chang, Shuguang Cui, and
  Jian~Jun Zhang.
\newblock Skeleton-bridged point completion: From global inference to local
  adjustment.
\newblock In {\em NIPS}, 2020.

\bibitem{Pistilli_2020_ECCV_denoise}
Francesca Pistilli, Giulia Fracastoro, Diego Valsesia, and Enrico Magli.
\newblock Learning graph-convolutional representations for point cloud
  denoising.
\newblock In {\em ECCV}, 2020.

\bibitem{Sarmad_2019_CVPR_rlgan}
Muhammad Sarmad, Hyunjoo~Jenny Lee, and Young~Min Kim.
\newblock Rl-gan-net: A reinforcement learning agent controlled gan network for
  real-time point cloud shape completion.
\newblock In {\em CVPR}, 2019.

\bibitem{Simonovsky_2017_cvpr_ecc}
Martin Simonovsky and Nikos Komodakis.
\newblock Dynamic edge-conditioned filters in convolutional neural networks on
  graphs.
\newblock In {\em CVPR}, 2017.

\bibitem{song_CVPR16_metric}
Hyun~Oh Song, Yu Xiang, Stefanie Jegelka, and Silvio Savarese.
\newblock Deep metric learning via lifted structured feature embedding.
\newblock In {\em CVPR}, 2016.

\bibitem{Stutz_2018_IJCV_voxel}
David Stutz and Andreas Geiger.
\newblock Learning 3d shape completion under weak supervision.
\newblock {\em IJCV}, 128(5):1162--1181, 2018.

\bibitem{Sung_2015_TOG_datadriven}
Minhyuk Sung, Vladimir~G. Kim, Roland Angst, and Leonidas Guibas.
\newblock Data-driven structural priors for shape completion.
\newblock {\em ACM Transactions on Graphics (TOG)}, 34(6), 2015.

\bibitem{Tchapmi_2019_cvpr_topnet}
Lyne~P Tchapmi, Vineet Kosaraju, S.~Hamid Rezatofighi, Ian Reid, and Silvio
  Savarese.
\newblock Topnet: Structural point cloud decoder.
\newblock In {\em CVPR}, 2019.

\bibitem{ThrunW_2005_ICCV_symmetry}
Sebastian Thrun and Ben Wegbreit.
\newblock Shape from symmetry.
\newblock In {\em ICCV}, 2005.

\bibitem{Varley_2017_IROS_robotgrasp}
Jacob Varley, Chad DeChant, Adam Richardson, Joaqu{\'{\i}}n Ruales, and
  Peter~K. Allen.
\newblock Shape completion enabled robotic grasping.
\newblock In {\em IROS}, 2017.

\bibitem{wang_2018_ECCV_pixel2mesh}
Nanyang Wang, Yinda Zhang, Zhuwen Li, Yanwei Fu, Wei Liu, and Yu-Gang Jiang.
\newblock Pixel2mesh: Generating 3d mesh models from single rgb images.
\newblock In {\em ECCV}, 2018.

\bibitem{Wang_2020_CVPR_cascaded}
Xiaogang Wang, Marcelo H. Ang~Jr.  , and Gim~Hee Lee.
\newblock Cascaded refinement network for point cloud completion.
\newblock In {\em CVPR}, 2020.

\bibitem{wang_2019_dgcnn}
Yue Wang, Yongbin Sun, Ziwei Liu, Sanjay~E. Sarma, Michael~M. Bronstein, and
  Justin~M. Solomon.
\newblock Dynamic graph cnn for learning on point clouds.
\newblock {\em ACM Transactions on Graphics (TOG)}, 2019.

\bibitem{Wang_2020_ECCV_softpoolnet}
Yida Wang, David~Joseph Tan, Nassir Navab, and Federico Tombari.
\newblock Softpoolnet: Shape descriptor for point cloud completion and
  classification.
\newblock In {\em ECCV}, 2020.

\bibitem{Weinberger_jmlr_2009_metric}
Kilian~Q. Weinberger and Lawrence~K. Saul.
\newblock Distance metric learning for large margin nearest neighbor
  classification.
\newblock {\em J. Mach. Learn. Res.}, page 207–244, 2009.

\bibitem{Wen_2020_CVPR_skipattention}
Xin Wen, Tianyang Li, Zhizhong Han, and Yu-Shen Liu.
\newblock Point cloud completion by skip-attention network with hierarchical
  folding.
\newblock In {\em CVPR}, 2020.

\bibitem{wu_2018_cvpr_instance}
Zhirong Wu, Yuanjun Xiong, X~Yu Stella, and Dahua Lin.
\newblock Unsupervised feature learning via non-parametric instance
  discrimination.
\newblock In {\em CVPR}, 2018.

\bibitem{Yang_CVPR_2018_foldingnet-genuswise}
Yaoqing Yang, Chen Feng, Yiru Shen, and Dong Tian.
\newblock Foldingnet: Point cloud auto-encoder via deep grid deformation.
\newblock In {\em CVPR}, 2018.

\bibitem{Yi_ACM_2016_shapenetpart}
Li Yi, Vladimir~G. Kim, Duygu Ceylan, I-Chao Shen, Mengyan Yan, Hao Su, Cewu
  Lu, Qixing Huang, Alla Sheffer, and Leonidas Guibas.
\newblock A scalable active framework for region annotation in 3d shape
  collections.
\newblock {\em ACM Trans. Graph.}, 35(6), Nov. 2016.

\bibitem{yu_2018_CVPR_upsampling}
Lequan Yu, Xianzhi Li, Chi-Wing Fu, Daniel Cohen-Or, and Pheng-Ann Heng.
\newblock Pu-net: Point cloud upsampling network.
\newblock In {\em CVPR}, 2018.

\bibitem{yuan_2018_3dv_pcn}
Wentao Yuan, Tejas Khot, David Held, Christoph Mertz, and Martial Hebert.
\newblock Pcn: Point completion network.
\newblock In {\em 3DV}, 2018.

\bibitem{Zhang_2020_ECCV_detail}
Wenxiao Zhang, Qingan Yan, and Chunxia Xiao.
\newblock Detail preserved point cloud completion via separated feature
  aggregation.
\newblock In {\em ECCV}, 2020.

\end{thebibliography}
}

\null\newpage
\clearpage
\appendix

\section{Supplementary Material}
\subsection{Additional Benchmark Analysis}
\begin{figure}[t!]
    \centering
    \begin{adjustbox}{width=0.5\textwidth}
    \begin{tabular}{@{~}c@{~~}c@{~~}c@{~~}c@{~~}c@{~}}
    Input & RFA & PF-Net & \our & GT \\
    \includegraphics[width=0.1\textwidth]{figures/known_chair/partial_trimmed.jpg}& 
    \includegraphics[width=0.1\textwidth]{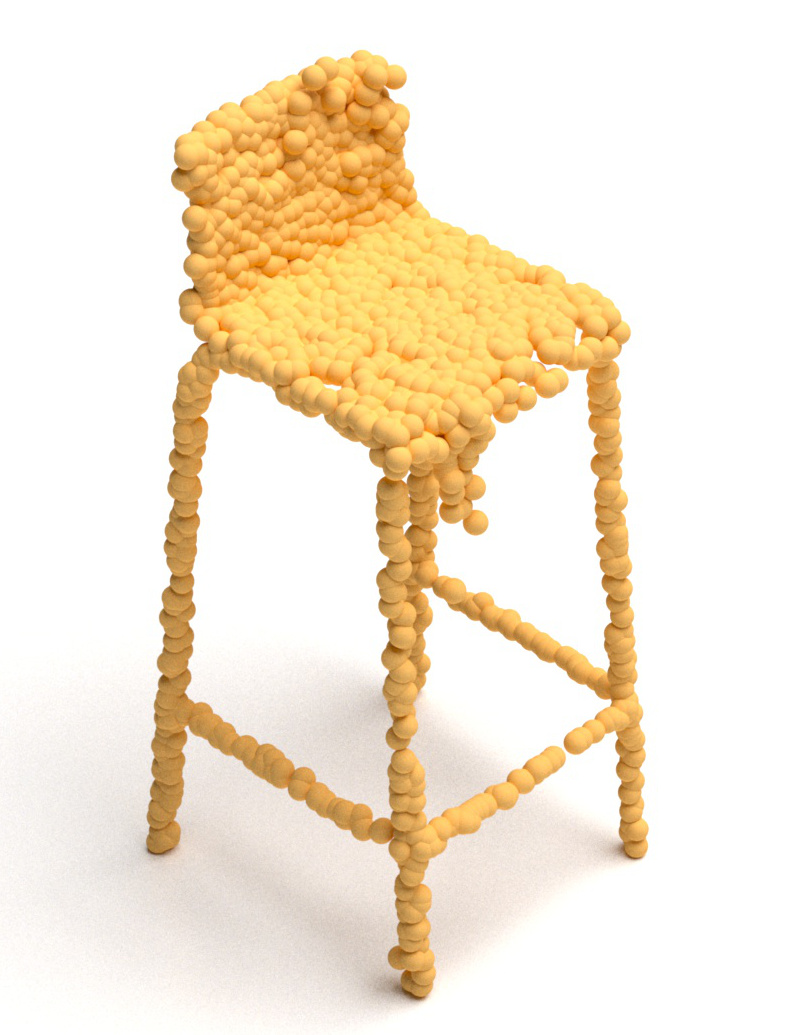}&
    \includegraphics[width=0.1\textwidth]{figures/known_chair/PFNet_discr_trimmed.jpg}&
    \includegraphics[width=0.1\textwidth]{figures/known_chair/deco_trimmed.jpg}&
    \includegraphics[width=0.1\textwidth]{figures/known_chair/gt_trimmed.jpg}\\
    \includegraphics[width=0.1\textwidth]{figures/known_guitar/partial_trimmed.jpg}& 
    \includegraphics[width=0.1\textwidth]{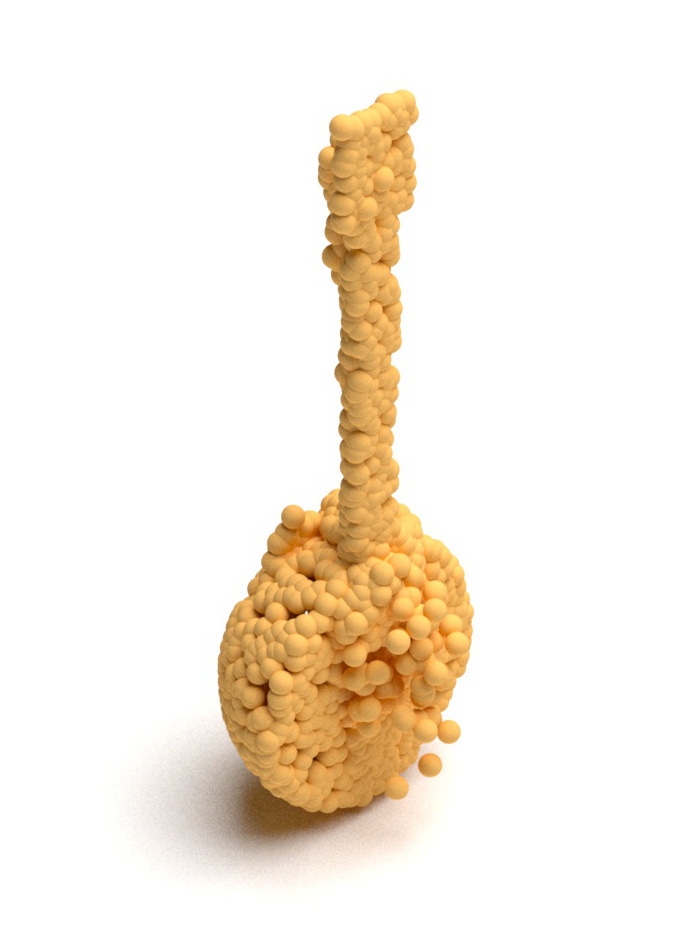}&
    \includegraphics[width=0.1\textwidth]{figures/known_guitar/PFNet_discr_trimmed.jpg}&
    \includegraphics[width=0.1\textwidth]{figures/known_guitar/deco_trimmed.jpg}&
    \includegraphics[width=0.1\textwidth]{figures/known_guitar/gt_trimmed.jpg}\\
    \includegraphics[width=0.1\textwidth]{figures/known_table/partial_trimmed.jpg}& 
    \includegraphics[width=0.1\textwidth]{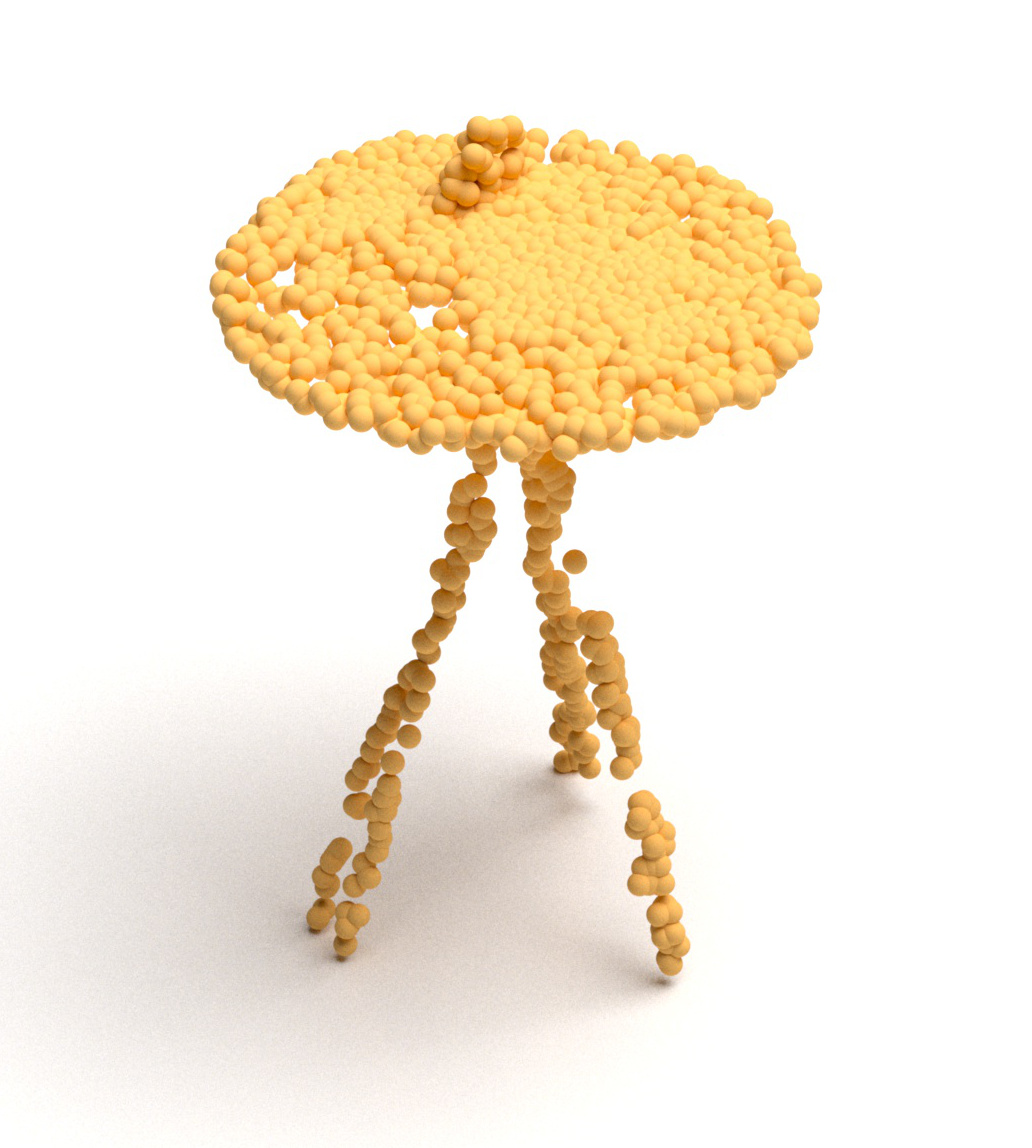}&
    \includegraphics[width=0.1\textwidth]{figures/known_table/PFNet_discr_trimmed.jpg}&
    \includegraphics[width=0.1\textwidth]{figures/known_table/deco_trimmed.jpg}&
    \includegraphics[width=0.1\textwidth]{figures/known_table/gt_trimmed.jpg}\\
    \end{tabular}
\end{adjustbox}
\caption{\emph{Known Categories - Qualitative}. In order from top to bottom: chair, guitar, table. RFA shows artifacts and less precise reconstruction than PF-Net and \our.}
\vspace{-3mm}
\label{fig:additional_qualitative}
\end{figure}

In the main paper we mentioned a recent work on point-cloud shape completion based on the idea of separated feature aggregation  \cite{Zhang_2020_ECCV_detail}. It uses local features to represent the known part and keep the original details, while global features are exploited for the missing part to describe the latent underlying surface.
Since the proposed network is designed to reconstruct the complete shape with ground truth clouds containing 16384 points, we operated some minimal changing on the architecture to get a fair comparison on 2048-points ground truth without corrupting its original nature.
Specifically, we started from the Residual Feature Aggregation (RFA) method, in which the missing part is represented with residual features between the global shape and the known part. 
We considered two variants: in the first one we generated a coarse output of 1024 points, then refined to 2048 points by the folding module inherited from PCN \cite{yuan_2018_3dv_pcn}.
In the second one we dropped the folding module and we selected the top scored 2048 points at the final attention module as prediction. 
We also experimentally verified that the repulsion loss of the method becomes detrimental when dealing with a low-resolution ground truth, so we did not include it in the learning process. 
This second variant obtained better results than the first and the corresponding CD are collected in Table \ref{table:additional}. The comparison indicates that both PF-Net and \our largely outperform RFA.  
The renderings in Figure \ref{fig:additional_qualitative} confirm that RFA produces a reasonable overall object shape, but the missing part is often noisy and reconstructed with artifacts.

\begin{table}[t!]
    \centering
    \begin{tabular}{@{~~}c@{~~}|@{~~}c@{~~~~}c@{~~~~}c@{~~}}
\hline
\textbf{Category} & \textbf{RFA} \cite{Zhang_2020_ECCV_detail} & \textbf{PF-Net} \cite{Huang_2020_CVPR_pfnet} & \textbf{\our} \\
\hline
Airplane & 26.747 &10.805 & \textbf{10.003} \\
Bag & 40.153 & 38.485  &\textbf{28.508} \\
Cap & 47.150 & 50.450 & \textbf{36.436}\\
Car & 59.167 & \textbf{21.640} &22.963\\
Chair & 29.227 & 19.490 & \textbf{16.428}\\
Lamp & 64.243 & 42.910 & \textbf{24.150} \\
Laptop & 27.880 &\textbf{11.220}  &12.706 \\
Motorbike & 71.623 & 19.905 & \textbf{19.136}\\
Mug & 80.200 &\textbf{31.880} & 34.239 \\
Pistol & 23.783 &\textbf{10.885}  &12.266\\
Skateboard & 127.413 & 12.365 & \textbf{9.861} \\
Table & 31.903 & 20.845 & \textbf{17.120}\\
Guitar & 13.357 & \textbf{4.425}  &4.482\\
\hline
Overall & 36.773 & 20.445 & \textbf{16.517} \\
\hline
\end{tabular}
\caption{\emph{Known Categories - Quantitative}. Chamfer Distance on the point cloud missing region scaled by $10^4$. The lower, the better.}
\label{table:additional}
\end{table}

\subsection{Decoder Output \& Frame Dimension}
Our decoder includes two SAG Pool layers \cite{Lee_2019_icml_sagpool}, whose purpose is to reduce the number of input points down to the number of points of the missing part.
We exploit a hierarchical pooling logic in order to predict at different decoder depth both the \emph{frame + missing region} $\bY_{fm}$, and the \emph{missing region} $\bY_{m}$.
The total number of output points at the two prediction heads depends on the number of pooled feature-space nodes, which are then decoded from the feature to the 3D space. 
While the choice of the $N_2$ parameter is constrained by the \emph{missing part} ground truth size ($N_2=M$), we are free to tune $N_1$, as long as it holds $N_1 \geq M+F$.
As specified in the main paper, in case of the standard single hole analysis, we set $M=F=512$, so $\bX_m$ has dimension $(512,3)$, while $\bX_{fm}$ has dimension $(1024,3)$. For the decoder we had $N_1=1280$ and $N_2=512$, thus resulting in $\bY_{fm}$ and $\bY_{m}$ respectively with size $(1280, 3)$ and $(512, 3)$. 

In Table \ref{tab:frame} we show the results obtained by varying $\{N_1,F\}$: the CD are always lower than those of the best competitor PF-Net (20.445). Moreover the obtained results confirm the effectiveness our parameter choice. 

In the main paper we also discussed two robustness tests. In the case of a \emph{single large hole} ($50\%$ of point cloud missing, 1024 points out of a whole shape of 2048 points) we simply dropped the frame condition and removed the two SAG Pool layers from the decoder, thus we did not use the frame auxiliary prediction in training. 
Despite this simplification, \our consistently outperforms its best competitor PFNet, demonstrating the effectiveness of our architecture and training procedure also when half of the complete shape is missing.
In the case of \emph{two holes}, each covering $12.5\%$ of the point cloud, we kept the condition $M=F$, so each crop consists of 256 points with their respective frame of equal cardinality out of a whole shape of 2048 points. The results in the main paper have shown how recovering the complete shape from a multiple-drilled partial input is way harder than recovering from a single-drilled shape, nevertheless \our is still able to outperform all the considered baselines.

\subsection{Contrastive Learning: Quadruplets vs Pairs}
In the main paper we described our strategy to extract global information from the point clouds via contrastive learning. Specifically we adopted a variant based on sample quadruplets, rather than on pairs as in the standard contrastive learning solution \cite{chen_2020_icml_simclr}. We present here a detailed analysis of this choice. More precisely, Table \ref{table:quadrupletsVSpairs} shows how using sample pairs can still provide good results, but passing from pairs to quadruplets allows us to work with a more manageable batch size, while also providing a further improvement in the reconstruction accuracy.

\subsection{DGCNN for Denoising}

In all \our experiments in the main paper we used at the local encoder the powerful Graph-Convolutional Point Denoising network (GPDNet) proposed in \cite{Pistilli_2020_ECCV_denoise}. Here we also present the completion results obtained by replacing it with a more conventional DGCNN \cite{wang_2019_dgcnn} encoder. All the other components of \our remain the ones already described, and we follow the same pre-training procedure adopted in the main paper for the denoising task (Gaussian noise, mean=0, standard deviation=0.02) of the simplified local encoder.
The results in Table \ref{table:denoise} show that the obtained DGCNN-based lighter version of \our still provides state-of-the-art performance, highlighting the effectiveness of our training strategy regardless of the specific adopted graph convolution blocks and backbone. 
As reference we also report the PF-Net baseline results and the number of parameters for all the considered variants which confirms the significant advantage of \our with respect to its best competitor also in terms of parameter cardinality.

\begin{table}[t!]
    \centering
    \begin{tabular}{@{~~}l@{~~~~}|c@{~~~~}c@{~~~~}c@{~~~~}}
    \hline
    \multirow{2}{*}{$\mathbf{N_1}$} &
    \multicolumn{3}{c}{\textbf{M=512}} \\
    & \textbf{F=256} & \textbf{F=512} & \textbf{F=768} \\
     \hline
     1024 & 19.001 &  18.129 & 18.595 \\
     1280 & 17.693 & \textbf{16.517} & 18.068 \\
     \hline
    \end{tabular}
    \caption{\emph{Known Categories - Single Hole.} Chamfer Distance results  scaled by $10^4$, obtained by changing the auxiliary decoder output and frame dimension.} 
    \label{tab:frame}
\end{table}

\begin{table}[t!]
    \centering
    \begin{tabular}{@{~~}l@{~~~~}l@{~~}|c@{~~}}
\hline
\multicolumn{2}{c|}{\textbf{Contrastive Learning Variants}} & {\textbf{Overall CD}} \\
 \hline
Pairs & {Batch Size}~=~$98\times 2$ & 18.030 \\
Quadruplets & {Batch Size}~=~$38 \times 4$ & \textbf{16.517} \\
 \hline
\end{tabular}
\caption{Overall average Chamfer Distance scaled by $10^4$, obtained by changing the Contrastive Learning strategy for the global encoder. }
\label{table:quadrupletsVSpairs}
\end{table}
\begin{table}[t!]
    \centering
    \begin{tabular}{@{~~}l@{~~}|c@{~~}|c@{~~}}
\hline
\multicolumn{1}{c|}{\textbf{Local Denoising Variants}} & {\textbf{Overall CD}} & {\textbf{Parameters}}\\
 \hline
DeCo GPDNet \cite{Pistilli_2020_ECCV_denoise} & \textbf{16.517}  & $1.66 \times 10^6$ \\
DeCo DGCNN \cite{wang_2019_dgcnn} & 19.667  & $1.13 \times 10^6$ \\
\hline
\hline
PF-Net \cite{Huang_2020_CVPR_pfnet} & 20.445 & $76.77 \times 10^6$ \\
\hline
\end{tabular}
\caption{Overall average Chamfer Distance scaled by $10^4$, obtained by changing the Denoising Strategy for the local encoder. }
\label{table:denoise}
\end{table}

\subsection{Further Training Details}
We provide here more details about the global + local feature aggregation logic.
One way to implement the feature combination is by concatenating the global feature vector to each point local feature and feeding them to a 1D conv. layer. In the specific case, the 1D conv. layer has output size 256, which is the dimensionality of the global+local per-point embedding input to the Decoder. 
This would unnecessarily cause the same global features to be processed $N$ times.
We optimized this implementation by separating global and local weight matrices of the 1D conv. layer and combining the obtained representations by summation.
This is equivalent to concatenation \& conv. but more efficient.

\end{document}